\documentclass{article} 
\usepackage{iclr2027_conference,times}

\usepackage{amsmath,amsfonts,bm}

\def\eqref#1{equation~\ref{#1}}

\def\1{\bm{1}}

\DeclareMathAlphabet{\mathsfit}{\encodingdefault}{\sfdefault}{m}{sl}
\SetMathAlphabet{\mathsfit}{bold}{\encodingdefault}{\sfdefault}{bx}{n}

\usepackage{url}            
\usepackage{booktabs}       
\usepackage{amsfonts}       
\usepackage{nicefrac}       
\usepackage{microtype}      
\usepackage{xcolor}         

\usepackage{xspace}
\usepackage{color}
\usepackage{amssymb}
\usepackage{amsmath}
\usepackage{multirow}
\usepackage{colortbl}
\usepackage{makecell,array}
\usepackage{enumitem}
\usepackage{graphicx}
\usepackage{caption}
\usepackage{wrapfig}
\usepackage[most]{tcolorbox}
\usepackage{titletoc}
\usepackage{fontawesome5}

\usepackage{subcaption}
\definecolor{cvprblue}{RGB}{66,93,181}

\definecolor{tablegreen}{RGB}{208,235,244}
\definecolor{tableblue}{RGB}{217,236,255}
\definecolor{tablepurple}{RGB}{218,227,250}
\definecolor{tablegray}{RGB}{242,242,242}

\definecolor{gainblue}{RGB}{60,73,137}

\definecolor{textgreen}{RGB}{70,101,118}
\definecolor{textblue}{RGB}{83,88,125}
\definecolor{textpurple}{RGB}{81,66,122}
\definecolor{textgray}{RGB}{80,80,80}

\definecolor{sementicgreen}{HTML}{25806D}
\definecolor{sementicpurple}{HTML}{5941B5}

\newcommand{\retainsem}[1]{\textit{\textcolor{sementicgreen}{#1}}}
\newcommand{\retractsem}[1]{\textit{\textcolor{sementicpurple}{#1}}}

\usepackage[colorlinks=true,
            citecolor=cvprblue,
            linkcolor=cvprblue,
            urlcolor=cvprblue]{hyperref}

\newcommand{\spira}{{\fontencoding{T1}\fontfamily{ppl}\selectfont SPIRA}}

\newcommand{\pcolor}[2]{%
  \raisebox{0.65ex}{%
    \makebox[0pt][l]{\textcolor{#1}{\scriptsize #2}}%
  }%
}

\newcommand{\stdcolor}[2]{%
  \raisebox{-0.30ex}{%
    \textcolor{#1}{\scriptsize $\pm$#2}%
  }%
}

\newcommand{\statcellgreen}[3]{%
  #1\hspace{0.20em}\pcolor{textgreen}{#2}\stdcolor{textgreen}{#3}%
}

\newcommand{\statcellblue}[3]{%
  #1\hspace{0.20em}\pcolor{textblue}{#2}\stdcolor{textblue}{#3}%
}

\newcommand{\statcellgray}[3]{%
  #1\hspace{0.20em}\pcolor{textgray}{#2}\stdcolor{textgray}{#3}%
}

\newcommand{\ourcellpurple}[2]{%
  #1\hspace{0.20em}\stdcolor{textpurple}{#2}%
}

\newcommand{\roicellgreen}[2]{%
  #1\hspace{0.20em}\stdcolor{textgreen}{#2}%
}

\newcommand{\roicellblue}[2]{%
  #1\hspace{0.20em}\stdcolor{textblue}{#2}%
}

\newcommand{\roicellpurple}[2]{%
  #1\hspace{0.20em}\stdcolor{textpurple}{#2}%
}

\newcommand{\roicellgray}[2]{%
  #1\hspace{0.20em}\stdcolor{textgray}{#2}%
}

\newcommand{\seedcellgreen}[2]{%
  #1\hspace{0.20em}\stdcolor{textgreen}{#2}%
}

\newcommand{\seedcellblue}[2]{%
  #1\hspace{0.20em}\stdcolor{textblue}{#2}%
}

\newcommand{\seedcellpurple}[2]{%
  #1\hspace{0.20em}\stdcolor{textpurple}{#2}%
}

\definecolor{boxblue}{RGB}{221,236,255}
\definecolor{boxcolblue}{RGB}{115,138,191}
\definecolor{boxtitleblue}{RGB}{100,119,180}

\newcommand{\gainblueup}[1]{\textcolor{gainblue}{\textbf{$\uparrow$ #1}}}
\newcommand{\gainbluedown}[1]{\textcolor{gainblue}{\textbf{$\downarrow$ #1}}}

\newcommand{\linkitem}[5]{%
  {\hypersetup{hidelinks}%
   \href{#1}{\mbox{%
     \raisebox{#5}{%
       \resizebox{!}{#4}{\textcolor{black}{#2}}%
     }%
     \hspace{0.40em}%
     {\fontsize{10pt}{11pt}\selectfont
      \textbf{\textcolor{textpurple}{#3}}}%
   }}}%
}

\newtcolorbox{insightbox}[1]{
  enhanced jigsaw,
  breakable,
  colback=boxblue!30!white,
  colframe=boxcolblue,
  boxrule=0.9pt,
  arc=3pt,
  left=7pt,
  right=7pt,
  top=5.5pt,
  bottom=3.0pt,
  before skip=4pt,
  after skip=5pt,
  fontupper=\small,
  title={#1},
  colbacktitle=boxtitleblue,
  coltitle=white,
  fonttitle=\fontsize{8.2}{9.0}\selectfont\bfseries,
  attach boxed title to top left={xshift=14pt,yshift=-2.6mm},
  boxed title style={
    colback=boxtitleblue,
    colframe=boxblue,
    boxrule=0pt,
    arc=3pt,
    left=4pt,
    right=4pt,
    top=0.4pt,
    bottom=0.4pt
  }
}

\newtcolorbox{RQbox}{
  enhanced jigsaw,
  breakable,
  colback=boxblue!30!white,
  colframe=boxcolblue,
  boxrule=0.9pt,
  arc=3pt,
  left=5pt,
  right=7pt,
  top=1.5pt,
  bottom=1.5pt,
  before skip=3pt,
  after skip=-2pt,
  fontupper=\small
}

\setlist[itemize]{topsep=-1pt, itemsep=1pt, parsep=0pt, partopsep=0pt,left=1pt}

\title{Source or Prefix? Retaining What Helps the Target in Autoregressive 3D Medical Image Translation}

\author{%
  Junkai Liu$^1$\;\;
  Anghong Du$^1$\;\;\;
  Xiyuan Wang$^2$\;\;\;
  Le Zhang$^{1}$ \\[3pt]
    $^1$ School of Engineering, University of Birmingham \\
    $^2$ School of Biological Science and Medical Engineering, Southeast University \\[10pt]
    \makebox[\linewidth][c]{%
      \linkitem{https://spira-project.github.io/}
        {\includegraphics{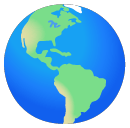}}
        {Project Page}{12pt}{-2.5pt}%
      \hspace{1.0em}\textcolor{gray}{\rule[-1.5pt]{0.8pt}{10.5pt}}\hspace{1.0em}%
      \linkitem{https://github.com/JK-Liu7/SPIRA}
        {\faGithub}{GitHub}{10pt}{-0.15ex}%
    }%
}

\iclrfinalcopy 
\begin{document}

\maketitle

\begin{abstract}

Translating between 3D medical imaging modalities requires preserving patient anatomy while recovering target-specific properties. A source-conditioned blockwise autoregressive model uses the source as a fixed anatomical anchor and earlier blocks as an evolving target prefix, and the difference between predictions with and without the prefix defines the prefix correction. The balance between source and prefix evidence varies across regions, while the magnitude of a correction's prediction change (\emph{effect}) does not necessarily reveal whether it helps the true target (\emph{utility}). In this work, we propose Source-Prefix Influence-aware Residual Autoregression ({\spira}), which learns \emph{how much of each prefix correction to retain across regions}. During training, {\spira} finds the largest removable portion of each correction without increasing target loss at the current prediction
states and trains a lightweight gate to reproduce the resulting predictive distribution, thus learning how to balance source and prefix during generation. Extensive experiments across multiple modalities and diverse 3D medical 
translation tasks demonstrate consistent gains in translation quality.
Mechanism analyses show that similar prefix effects can have different
utilities depending on target alignment, while {\spira} better matches oracle decisions and achieves larger gains when more prefix
correction can be safely retracted. 

\end{abstract}

\section{Introduction}

3D medical image translation synthesizes a target-modality volume from a source-modality volume~\citep{wang2025unisyn, romoli2026cross}, with potential applications such as reducing gadolinium use through contrast-enhanced MRI synthesis and supporting radiotherapy planning through synthetic CT~\citep{10167641, 11316538}. In paired settings, the source already provides rich patient-specific anatomy and spatial structure, making it a strong anchor for target generation~\citep{CHEN2025103605}. However, the same anatomy can appear differently across imaging modalities~\citep{Shin_2025_CVPR, du2026improving}. The source alone may therefore leave target-specific properties such as contrast, density, or functional activity ambiguous~\citep{10444695, 11558466}.

\begin{figure*}[t]
\centering
\includegraphics[width=1.00\linewidth]{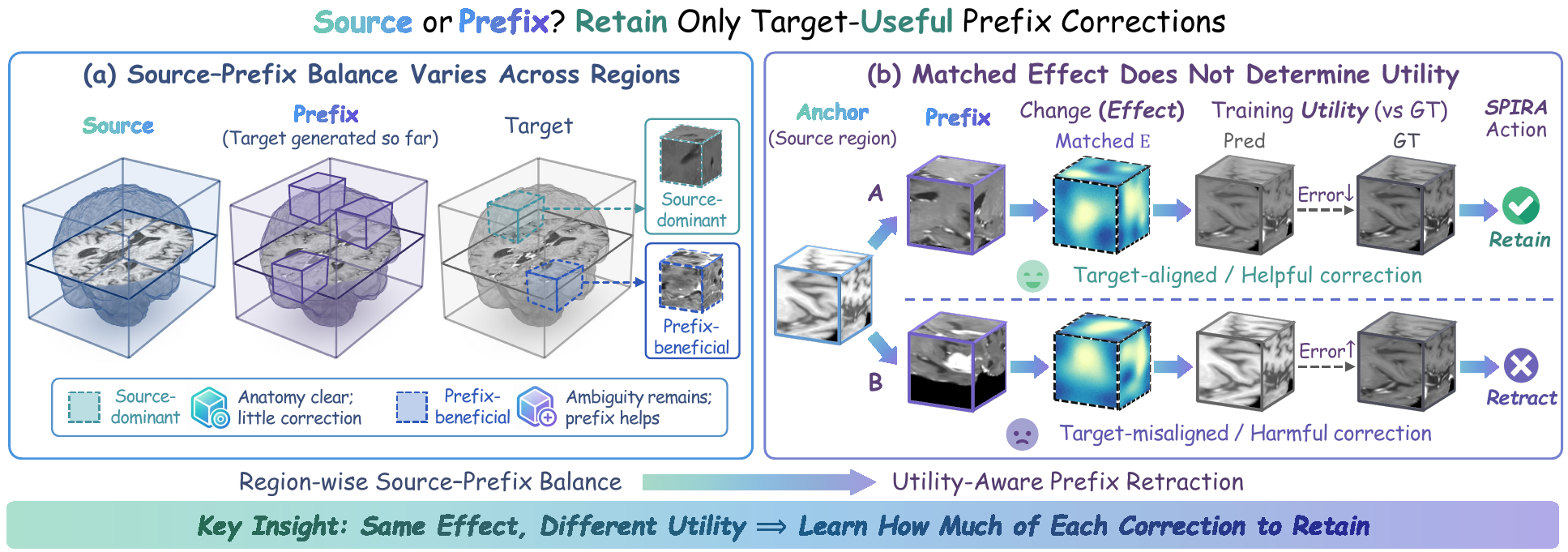}
\vspace{-16pt}
\caption{\textbf{(a) Challenge:} The balance between source evidence and target-side context varies across target regions. \textbf{(b) Insight:} Predictive \emph{effect} measures how much a prefix correction changes the prediction, while correction \emph{utility measures} whether that change favors the true target. Corrections with matched effects can therefore have different utilities, motivating utility-aware prefix retraction.}
\vspace{-16pt}
\label{fig:teaser}
\end{figure*}

An autoregressive (AR) formulation complements this source anchor with earlier target content~\citep{NEURIPS2024_9a24e284, zhu2026causal}, which we call the target \emph{prefix}. Each prediction therefore combines a fixed, spatially aligned source condition with evolving target-side context. During training, the prefix comes from the true target; at inference, it is generated from previous predictions and may carry errors forward~\citep{NEURIPS2025_f4823f83, ICLR2026_8a0501a7}. The relative contributions of the fixed condition and evolving context therefore need not be uniform across predictions~\citep{Jin_2026_CVPR, chen2026context}. In paired medical translation, this variation is spatially explicit because the source is aligned with the target and already captures much of its anatomy~\citep{11197215, 11353005}. The prefix then contributes a \emph{correction} beyond the source-only prediction, whose value can vary across regions. Some regions are already well explained by the source, whereas others still benefit from target-side context~\citep{Liu_2025_ICCV} (Fig.~\ref{fig:teaser}a). This leads to our central question:


\begin{insightbox}{Key Question}
\emph{How much of the current prefix correction should be retained for each target region?}
\end{insightbox}

To answer this question, a natural proxy for prefix need is its predictive \emph{effect}, namely how much the correction changes the predictive distribution~\citep{ICLR2024_edc36117}. We define its \emph{utility} by whether this change favors the true target. Predictive effect alone, however, does not determine utility. In Fig.~\ref{fig:teaser}b, corrections with similar effects may increase the probability of the target, leave it almost unchanged, or decrease it. This pattern is also observed on BraTS 2024~\citep{de20242024}, where among regions with similar prediction changes, 29.3\% reduce the true-target likelihood while 40.0\% increase it (Sec.~\ref{sec:motivation}), further showing that similar effects can have different utilities.

\begin{insightbox}{Key Insight}
Prefix corrections with matched predictive effects can have different utilities, so effect alone cannot determine how much of a correction to retain.
\end{insightbox}

Building upon this insight, we propose
\textbf{\underline{S}ource-\underline{P}refix \underline{I}nfluence-aware
\underline{R}esidual \underline{A}utoregression ({\spira})},
which controls the contribution of generated context
through selective prefix retraction to
\retractsem{reduce harmful or unnecessary prefix corrections}
while \retainsem{preserving target-useful prefix information}. Specifically, we instantiate {\spira} for 3D medical image translation in a source-conditioned blockwise AR model that generates 3D latent blocks sequentially. Within this paradigm, a shared AR model produces source-only and full-prefix predictions, whose difference is the current prefix correction. We define target-aware retraction regret as the change in target loss after retraction relative to full-prefix AR, and show that it decomposes exactly into predictive effect and target alignment. During training, a ground-truth (GT) oracle finds the maximal zero-regret retraction, which is the largest portion of the correction that can be retracted without increasing target loss. A lightweight correction-aware gate learns to reproduce the resulting predictive distribution from the source and current prefix. At inference, the gate applies the predicted retraction before each new target unit is generated and appended to the prefix. Our contributions are summarized as follows:

\begin{itemize}

    \item \textbf{Reframing.}
    We frame source-conditioned AR generation as a balance between a fixed, information-rich condition and evolving target-side context, and formulate this problem region-wise in 3D medical image translation, where predictive effect alone does not determine how much of a prefix correction should be retained.

    

    \item \textbf{Approach.} We derive an exact decomposition of target-aware retraction regret into predictive effect and target alignment, and introduce {\spira} with a maximal zero-regret oracle and a lightweight gate that directly distills the oracle predictive distribution.

    \item \textbf{Experiments.} Across eight translation tasks on three datasets, {\spira} consistently improves translation quality. Further analyses explain why corrections with similar effects can require different retention and how {\spira} learns these distinctions to improve free-running generation.

\end{itemize}
\section{Motivation Analysis}
\label{sec:motivation}

\begin{figure*}[t]
\centering
\includegraphics[width=0.98\linewidth]{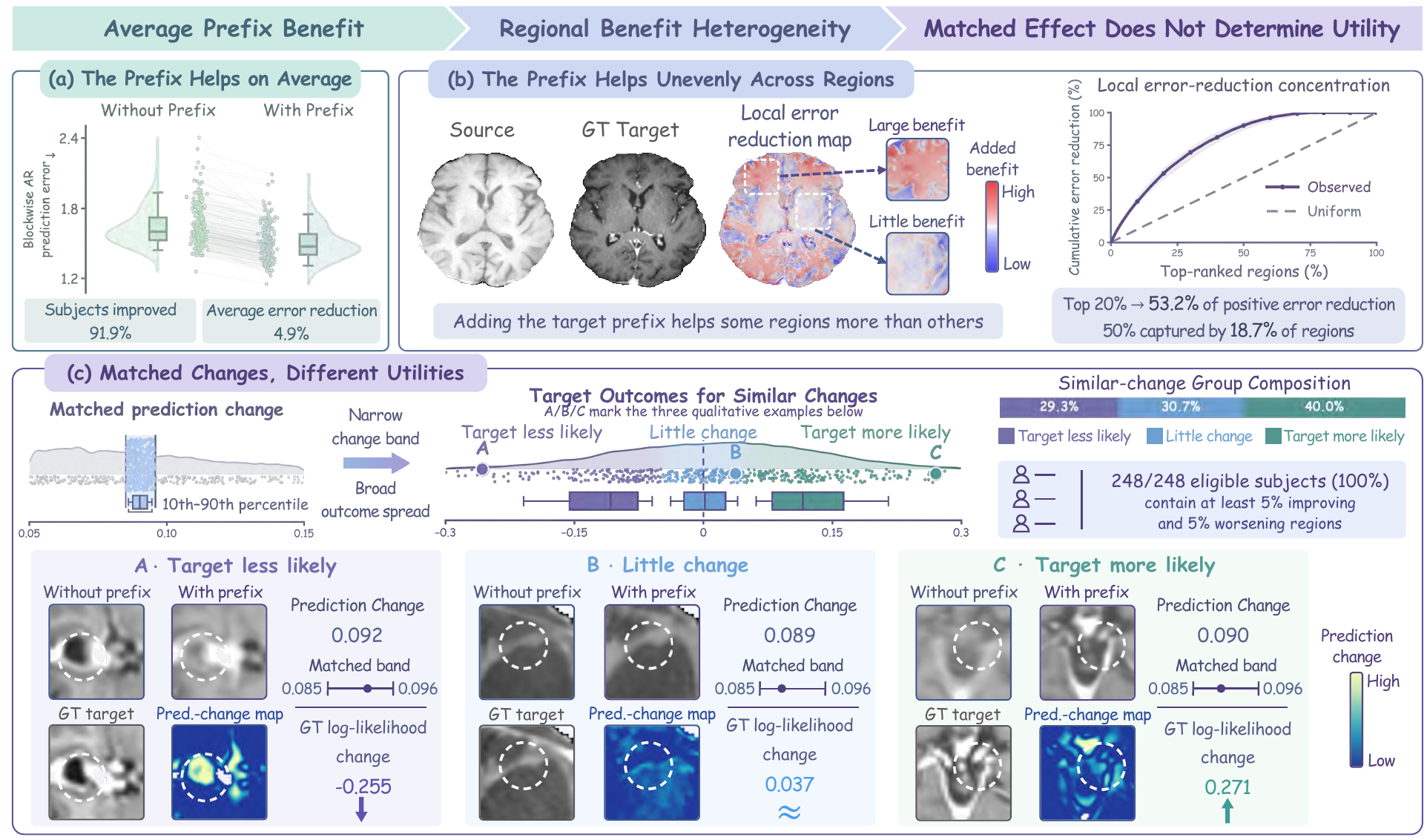}
\vspace{-8pt}
\caption{\textbf{Motivation analysis.} 
\textbf{(a)} In token space, the target prefix reduces prediction error overall. 
\textbf{(b)} In decoded image space, its benefit varies substantially across regions. 
\textbf{(c)} In predictive-distribution space, similar prediction changes can yield different true-target outcomes.}
\vspace{-16pt}
\label{fig:motivation}
\end{figure*}

We conduct the analysis on BraTS 2024~\citep{de20242024} using the T1n$\rightarrow$T1c task. Under teacher forcing, we compare predictions with and without the preceding GT target prefix to isolate clean prefix effects from accumulated generation errors. Further details are provided in Appendix~\ref{app:motivation}.

\textbf{The Prefix Helps on Average.}
As shown in Fig.~\ref{fig:motivation}(a), adding the target prefix improves 91.9\% of held-out subjects and reduces blockwise AR prediction error by 4.9\% on average, confirming that it provides useful information beyond the source.

\textbf{The Prefix Helps Unevenly Across Regions.}
We decode both predictions and measure local reconstruction-MAE reduction over fixed 3D regions. Fig.~\ref{fig:motivation}(b) shows that some regions benefit substantially while others change little. Across subjects, the top 20\% of regions account for 53.2\% of the positive error reduction, indicating strongly region-dependent prefix benefit.

\textbf{Matched Changes, Different Utilities.}
We next ask whether similarly sized prefix-induced prediction changes yield similar target outcomes. Using factor-averaged total variation distance, we select prediction changes within the \(45\)th--\(55\)th percentile range and categorize the corresponding true-target likelihood changes using a small neutral range around zero. As shown in Fig.~\ref{fig:motivation}(c), similar prediction changes still produce improving, neutral, and worsening outcomes, and all 248 eligible subjects contain both improving and worsening regions. Because the prefix is GT, these harmful outcomes arise even with correct target-side context rather than accumulated generation errors.

\begin{insightbox}{Takeaway}
The target prefix provides region-dependent information beyond the source, while similar prediction changes can have different utilities. Thus, predictive effect alone cannot determine the appropriate source--prefix balance, or equivalently how much of each prefix correction to retain.
\end{insightbox}
\section{Methodology}
\label{sec:method}

\begin{figure*}[t]
\centering
\includegraphics[width=1.00\linewidth]{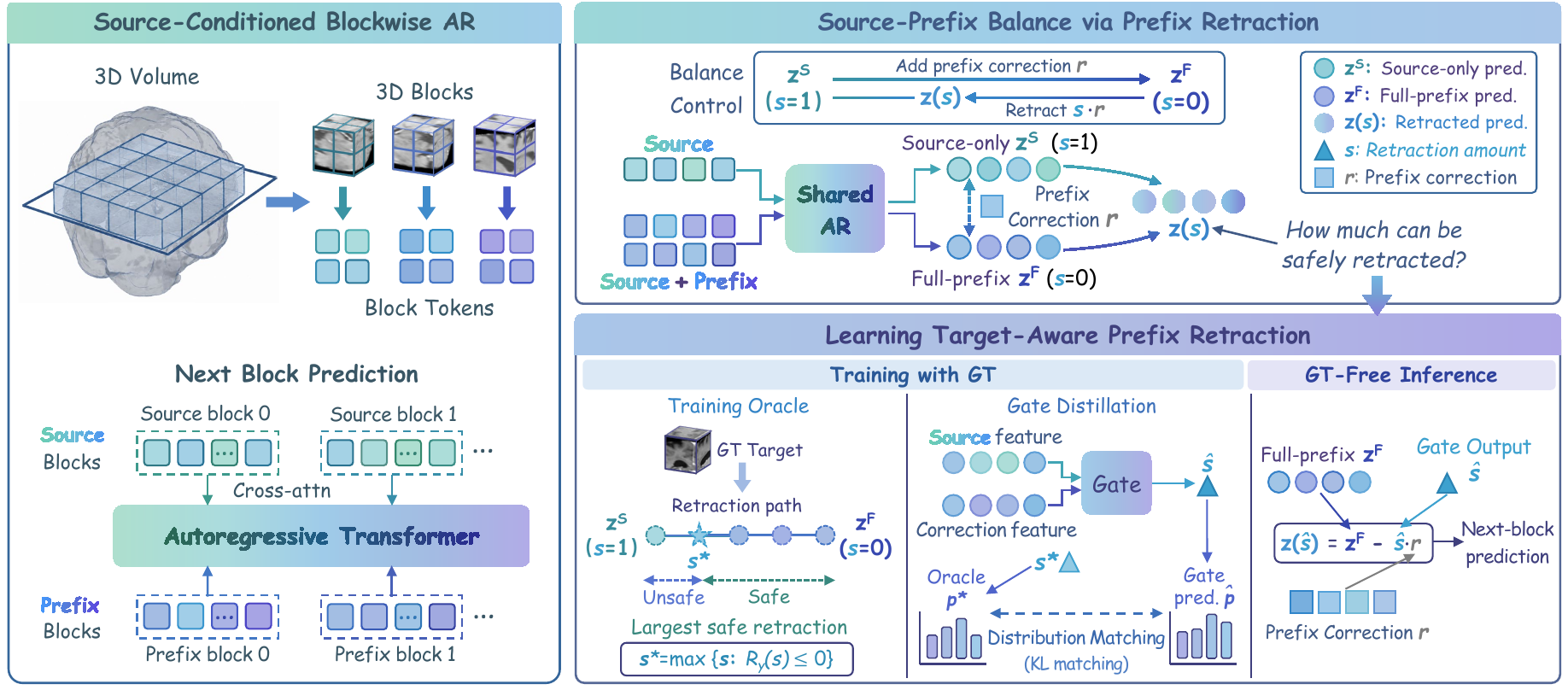}
\vspace{-16pt}
\caption{\textbf{Overview of {\spira}.}
A shared source-conditioned blockwise AR predictor produces source-only and full-prefix logits, whose difference defines a prefix correction and a continuous retraction path.
During training, a GT oracle identifies the maximal zero-regret retraction, and a correction-aware gate distills the resulting predictive distribution for GT-free generation.}
\vspace{-6pt}
\label{fig:framework}
\end{figure*}

Motivated by the above observations, we introduce {\spira}, which learns a region-wise source--prefix balance by controlling how much of the
prefix-induced update is retained at each target region. As illustrated in Fig.~\ref{fig:framework}, the same AR predictor produces predictions with and without the current prefix; their difference defines the prefix correction. {\spira} then learns to retract only the portion that can be removed without increasing target loss. We introduce the AR formulation in Sec.~\ref{sec:blockwise_AR}, characterize prefix correction and retraction in Secs.~\ref{sec:prefix_corrections}--\ref{sec:prefix_retraction}, and summarize training and inference in Sec.~\ref{sec:train_infer}.

\subsection{Source-Conditioned Blockwise AR}
\label{sec:blockwise_AR}

Given a paired 3D source and target volume, a frozen visual tokenizer maps them into spatially aligned discrete latent tokens. For simplicity, we denote the encoded source tokens by \(S\) and partition the target latent grid into an ordered sequence of non-overlapping 3D block tokens \(Y_{1:N}\). Following~\cite{ren2025next}, we factorize the conditional distribution as
\begin{equation}
    p_{\theta}(Y_{1:N}\mid S)
    =
    \prod_{i=1}^{N}
    p_{\theta}(Y_i\mid S,P_{<i}),
    \qquad
    P_{<i}=(Y_1,\ldots,Y_{i-1}),
\end{equation}
where \(P_{<i}\) denotes the target prefix available before predicting block \(i\).

The source is bidirectionally encoded and injected through inter-block cross-attention and aligned source features within each predicted block. At each step, the AR predictor produces next-block logits \(
\mathbf{z}_i=f_{\theta}(S,P_{<i},i)\), where \(f_{\theta}\) denotes the shared source-conditioned blockwise AR predictor, conditioned on the source \(S\), causal target prefix \(P_{<i}\), and current block index \(i\). Spatial tokens within each block are predicted in parallel. Further architectural details are provided in Appendix~\ref{app:blockwise_AR}.

\subsection{Source-prefix Balance via Prefix Retraction}
\label{sec:prefix_corrections}

To make the source--prefix balance explicit, we reformulate each prediction as a source-only prediction plus a controllable prefix-induced correction. To isolate the contribution of the target prefix, we evaluate the same AR predictor at the same target position with and without the prefix:
\begin{equation}
    \mathbf{z}_i^{S}=f_{\theta}(S,\varnothing,i),
    \qquad
    \mathbf{z}_i^{F}=f_{\theta}(S,P_{<i},i),
\end{equation}
where \(\mathbf{z}_i^{S}\) and \(\mathbf{z}_i^{F}\) denote the \emph{source-only} and \emph{full-prefix} predictions, respectively.
In implementation, \(\varnothing\) is realized by masking all target-prefix blocks as invalid, so that no preceding target blocks are visible to the predictor. The two paths share all model parameters and differ only in whether the prefix is visible. Their difference,
\(\mathbf{r}_i=\mathbf{z}_i^{F}-\mathbf{z}_i^{S}\), defines the \emph{prefix correction}, i.e., the logit-space update introduced by the prefix beyond the source. The correction therefore isolates the contribution of evolving target-side context beyond the fixed source condition.

We control how much of this correction is retained through
\begin{equation}
    \mathbf{z}_i(s)
    =
    \mathbf{z}_i^{F}-s\mathbf{r}_i
    =
    \mathbf{z}_i^{S}+(1-s)\mathbf{r}_i,
    \qquad s\in[0,1],
\label{eqn:retraction_path}
\end{equation}
where \(\mathbf{z}_i(s)\) is the \emph{retracted prediction}, and \(s\) is the fraction of the correction being removed, i.e., the \emph{retraction amount}. Thus, \(s=0\) retains the full prefix correction, \(s=1\) returns to the source-only prediction, and intermediate values partially retract it. Within this guidance-style interpolation, the remaining question is \emph{how to determine an appropriate retraction amount $s$ for each target region}.

\subsection{Learning Target-aware Prefix Retraction}
\label{sec:prefix_retraction}

To determine this retraction amount, we first characterize when retraction is safe, and then learn to reproduce the corresponding target-aware decision without requiring the GT target at inference.

\textbf{When is retraction safe?}
Let \(\mathbf{p}_i^{F}=\operatorname{softmax}(\mathbf{z}_i^{F})\) and \(\mathbf{p}_i(s)=\operatorname{softmax}(\mathbf{z}_i(s))\). We define the predictive \emph{effect} as \(E_i(s)=D_{\mathrm{KL}}(\mathbf{p}_i^{F}\|\mathbf{p}_i(s))\), which measures how much retraction changes the prediction without using the GT target. To determine whether a correction is useful to the target, we measure the loss change caused by retraction, which we call \emph{retraction regret}:
\begin{equation}
    R_{i,y}(s)
    =
    \ell_y(\mathbf{z}_i(s))
    -
    \ell_y(\mathbf{z}_i^{F}),
\end{equation}
where \(\ell_y(\mathbf{z})=-\log\operatorname{softmax}(\mathbf{z})_y\). A positive regret means that retraction removes useful target information, whereas \(R_{i,y}(s)\leq0\) means that the correction can be retracted without increasing target loss, i.e., safe retraction. 



For a fixed prediction state, define the normalizing constant
$Z_i(s)=\sum_k \exp(z^F_{i,k}-s r_{i,k})$.
Since $z_i(s)=z_i^F-sr_i$, expanding the target loss and
predictive effect gives 
\begin{equation}
\begin{aligned}
R_{i,y}(s)
&= \log\frac{p^F_{i,y}}{p_{i,y}(s)}
 = s r_{i,y}+\log\frac{Z_i(s)}{Z_i(0)},\\
E_i(s)
&= \sum_k p^F_{i,k}
   \log\frac{p^F_{i,k}}{p_{i,k}(s)}
 = s\mathbb{E}_{k\sim p_i^F}[r_{i,k}]
   +\log\frac{Z_i(s)}{Z_i(0)}.
\end{aligned}
\label{eq:regret_effect_expansion}
\end{equation}

Subtracting these expressions yields the exact decomposition

\begingroup
\begin{equation}
R_{i,y}(s)=E_i(s)+sA_i,
\qquad
A_i=r_{i,y}-\mathbb{E}_{k\sim p_i^F}[r_{i,k}].
\label{eq:regret_decomposition}
\end{equation}
\endgroup
Here, $A_i$ measures the correction at the true target relative to its expectation under the full-prefix prediction. Thus, whether retraction increases target loss depends on both its predictive effect and target-relative alignment.

\textbf{Training oracle.}
Using the GT target during training, we define the \emph{maximal zero-regret
retraction} as
\begin{equation}
s_i^*=\max\{s\in[0,1]\mid R_{i,y}(s)\leq 0\}.
\label{eq:oracle}
\end{equation}

This oracle removes as much correction as possible without increasing the full-prefix target loss at the current prediction state. Its construction is supported by 
\begingroup
\begin{equation}{
\begin{aligned}
R_{i,y}'(s)
&=r_{i,y}-\mathbb{E}_{k\sim p_i(s)}[r_{i,k}],\\
R_{i,y}''(s)
&=\mathbb{E}_{k\sim p_i(s)}[r_{i,k}^2]
  -\left(\mathbb{E}_{k\sim p_i(s)}[r_{i,k}]\right)^2\\
&=\operatorname{Var}_{k\sim p_i(s)}[r_{i,k}]
\geq 0.
\end{aligned}
}
\end{equation}
\endgroup
Hence, $R_{i,y}$ is convex and $R_{i,y}'(0)=A_i$. Since it is also continuous with $R_{i,y}(0)=0$, the feasible set $\{s\in[0,1]:R_{i,y}(s)\leq0\}$ is a closed interval
$[0,s_i^*]$. We therefore set $s_i^*=1$ if $R_{i,y}(1)\leq0$;
otherwise, fixed-step bisection locates the right boundary
of this interval (Appendix~\ref{app:oracle_construction}). Unlike a minimum-loss oracle, this selects the largest feasible retraction rather than the lowest-loss point along the path. The decomposition and convexity remain valid when averaging over multiple prediction factors under a shared fixed history.

\textbf{Correction-aware gate.}
Because the oracle requires the GT target, it is unavailable during generation. We therefore use a lightweight MLP gate to predict
\(\hat{s}_i=g_{\phi}(\mathbf{u}_i^{S},\mathbf{u}_i^{r},w_i,e_i)\in[0,1]\),
where \(\mathbf{u}_i^{S}\) is the normalized source-only hidden feature and \(\mathbf{u}_i^{r}\) is the normalized hidden residual between the full-prefix and source-only paths, while \(w_i\) and \(e_i\) summarize the difference between the two prediction endpoints. Unlike the AR predictor trained for next-block prediction, the gate learns how much of the current source-to-prefix correction should be retained.

\paragraph{Gate supervision.}
Rather than regressing the oracle scalar $s_i^*$ itself, we supervise the gate by matching the predictive distribution induced by the oracle retraction. Let $\mathbf{p}_i^*=\operatorname{softmax}(\mathbf{z}_i(s_i^*))$ and $\hat{\mathbf{p}}_i=\operatorname{softmax}(\mathbf{z}_i(\hat{s}_i))$. We minimize $\mathcal{L}_{\mathrm{ret}}=\mathbb{E}_{i\in\mathcal{V}}\!\left[D_{\mathrm{KL}}\!\left(\mathbf{p}_i^* \,\|\, \hat{\mathbf{p}}_i\right)\right]$, which we call predictive-action distillation, distilling the oracle distribution along the source--full-prefix retraction path. Here, the KL averages conditional divergences over token factors under a shared teacher-forced history.

\subsection{Training and Inference}
\label{sec:train_infer}

\textbf{Training.}
{\spira} is trained in a single stage with
\(\mathcal{L}=\frac{1}{2}(\mathcal{L}_{\mathrm{full}}+\mathcal{L}_{\mathrm{source}})+\mathcal{L}_{\mathrm{ret}}\), where \(\mathcal L_{\rm full}\) and \(\mathcal L_{\rm source}\) are standard next-block prediction losses for the shared AR predictor, while
\(\mathcal{L}_{\mathrm{ret}}\) updates only the gate with detached oracle and endpoint quantities.

\textbf{Inference.}
At inference, the gate predicts $\hat{s}_i$ from source and prefix-derived features without GT or oracle supervision.
Source-only pre-head contexts are precomputed and cached, while
factor-conditioned logits are evaluated on demand. Setting $\hat{s}_i=0$ throughout recovers Full-prefix AR.

\section{Experiments}

In this section, we conduct experiments to answer the following four research questions (RQs):

\begin{RQbox}
\textbf{(RQ1) Effectiveness}:  Does {\spira} improve translation performance?  \S\ref{sec:exp_performance}

\textbf{(RQ2) Design}: Are {\spira}’s key design choices necessary for its performance gains?  \S\ref{sec:exp_ablation}

\textbf{(RQ3) Utility-Effect}: What explains the utility--effect gap?  \S\ref{sec:exp_phenomenon}

\textbf{(RQ4) Actionability}: Can {\spira} translate the utility--effect gap into generation gains?  \S\ref{sec:exp_mechanism}
\end{RQbox}


\subsection{Experimental Setup}
\label{sec:exp_setup}

\textbf{Datasets}. We evaluate {\spira} on three datasets. BraTS 2024~\citep{de20242024} provides brain MRI with four modalities: T1c, T1n, T2w, and T2f. SynthRAD 2025~\citep{https://doi.org/10.1002/mp.17981} covers head-and-neck and abdominal regions with paired MRI/CT and CBCT/CT data for MRI$\rightarrow$CT and CBCT$\rightarrow$CT translation. AutoPET~\citep{dexl2026autopet3} provides paired PET/CT scans across the brain, pelvis, and abdomen. Further dataset details are provided in Appendix~\ref{app:dataset}.

\textbf{Implementation Details}. We use a fine-tuned non-causal VidTok-FSQ~\citep{tang2024vidtok} as the visual tokenizer, with a compression ratio of $(8\times8\times4)$. Our AR backbone is adapted from LlamaGen~\citep{sun2024autoregressive}, with 3D patchification and 3D rotary positional embeddings to support volumetric generation. 
All models are trained on $(256\times256\times128)$ 3D crops. During inference, we use either full-volume or sliding-window generation, with KV caching to improve efficiency. Additional training details and hyperparameter settings are presented in Appendix~\ref{app:train_infer}.

\textbf{Comparison Methods.}
We compare {\spira} with representative GAN- and diffusion-based baselines, together with two matched internal baselines from the same trained model. Full-prefix AR provides the most controlled AR comparison by sharing the same tokenizer and architecture with {\spira}. \textbf{\textit{Source-only}} removes the target prefix ($s=1$), while \textbf{\textit{Full-prefix AR}} disables retraction ($s=0$) and uses the complete generated prefix. Together, they define the two endpoints of the retraction path and isolate whether {\spira}'s gains arise from selective prefix retraction.

\begin{table}[t]
\centering
\setlength{\tabcolsep}{0.7mm}
\renewcommand\arraystretch{1.0}
\caption{\textbf{Quantitative results for 3D MRI translation on BraTS 2024}. 
}
\vspace{-8pt}
\resizebox{\linewidth}{!}{
\begin{tabular}{lcccc|cccc|cccc|cccc}
\toprule
\multirow{2}{*}{\textbf{Methods}} 
& \multicolumn{4}{c|}{\textbf{T1n$\rightarrow$T1c}} 
& \multicolumn{4}{c|}{\textbf{T1c$\rightarrow$T1n}} 
& \multicolumn{4}{c|}{\textbf{T2w$\rightarrow$T2f}} 
& \multicolumn{4}{c}{\textbf{T2f$\rightarrow$T1c}} \\
\cmidrule(lr){2-5}
\cmidrule(lr){6-9}
\cmidrule(lr){10-13}
\cmidrule(lr){14-17}

& PSNR$\uparrow$ 
& SSIM$\uparrow$ 
& LPIPS$\downarrow$ 
& FID$\downarrow$ 
& PSNR$\uparrow$ 
& SSIM$\uparrow$ 
& LPIPS$\downarrow$ 
& FID$\downarrow$ 
& PSNR$\uparrow$ 
& SSIM$\uparrow$ 
& LPIPS$\downarrow$ 
& FID$\downarrow$ 
& PSNR$\uparrow$ 
& SSIM$\uparrow$ 
& LPIPS$\downarrow$ 
& FID$\downarrow$ \\

\midrule
\rowcolor{tablegray!70}
\multicolumn{17}{c}{\textit{GAN-based Methods}} \\

3D-mADUNet~\citep{10944137}
& 21.30 & 0.680 & 0.162 & 47.6
& 24.68 & 0.822 & 0.096 & 37.0
& 20.84 & 0.646 & 0.177 & 64.3
& 20.71 & 0.653 & 0.178 & 61.0 \\

MMHVAE~\citep{11197215}
& 21.66 & 0.691 & 0.157 & 45.2
& 24.23 & 0.819 & 0.097 & 39.2
& 20.67 & 0.643 & 0.182 & 61.9
& 20.90 & 0.662 & 0.170 & 54.6 \\

\midrule
\rowcolor{tablegray!70}
\multicolumn{17}{c}{\textit{Diffusion-based Methods}} \\

M2DN~\citep{10444695}
& 21.15 & 0.673 & 0.161 & 47.8
& 24.35 & 0.816 & 0.099 & 40.7
& 20.83 & 0.648 & 0.179 & 62.8
& 20.64 & 0.651 & 0.184 & 59.7 \\

selfRDB~\citep{ARSLAN2025103747}
& 22.03 & 0.695 & 0.139 & 44.8
& 25.16 & 0.834 & 0.082 & 34.0
& 21.42 & 0.674 & 0.159 & 46.9
& 21.33 & 0.677 & 0.160 & 47.0 \\

CRSG~\citep{11559110}
& 21.85 & 0.691 & 0.148 & 46.1
& 25.31 & 0.834 & 0.085 & 33.6
& 21.19 & 0.670 & 0.166 & 54.8
& 21.02 & 0.669 & 0.171 & 53.9 \\

\midrule

\rowcolor{tablegreen!60}
Source-only
& 21.59 & 0.690 & 0.151 & 46.1
& 24.96 & 0.828 & 0.091 & 34.7
& 21.01 & 0.670 & 0.164 & 57.2
& 21.09 & 0.671 & 0.163 & 49.6 \\

\rowcolor{tableblue!70}
Full-prefix AR
& 22.10 & 0.696 & 0.135 & 41.9
& 25.49 & 0.834 & 0.085 & 32.2
& 21.30 & 0.674 & 0.155 & 48.6
& 21.55 & 0.681 & 0.154 & 46.1 \\

\rowcolor{tablepurple!70}
\shortstack[l]{
    \textbf{{\spira}}\\[0.3pt]
    {\small\textcolor{gainblue}{\textbf{$\Delta$ vs. Best External}}}
}
& \shortstack[c]{\textbf{22.72}\\[0.3pt]{\small\gainblueup{0.69}}}
& \shortstack[c]{\textbf{0.705}\\[0.3pt]{\small\gainblueup{0.010}}}
& \shortstack[c]{\textbf{0.126}\\[0.3pt]{\small\gainbluedown{0.013}}}
& \shortstack[c]{\textbf{38.3}\\[0.3pt]{\small\gainbluedown{6.5}}}

& \shortstack[c]{\textbf{25.93}\\[0.3pt]{\small\gainblueup{0.62}}}
& \shortstack[c]{\textbf{0.842}\\[0.3pt]{\small\gainblueup{0.008}}}
& \shortstack[c]{\textbf{0.077}\\[0.3pt]{\small\gainbluedown{0.005}}}
& \shortstack[c]{\textbf{28.6}\\[0.3pt]{\small\gainbluedown{5.0}}}

& \shortstack[c]{\textbf{21.77}\\[0.3pt]{\small\gainblueup{0.35}}}
& \shortstack[c]{\textbf{0.680}\\[0.3pt]{\small\gainblueup{0.006}}}
& \shortstack[c]{\textbf{0.150}\\[0.3pt]{\small\gainbluedown{0.009}}}
& \shortstack[c]{\textbf{43.8}\\[0.3pt]{\small\gainbluedown{3.1}}}

& \shortstack[c]{\textbf{22.08}\\[0.3pt]{\small\gainblueup{0.75}}}
& \shortstack[c]{\textbf{0.687}\\[0.3pt]{\small\gainblueup{0.010}}}
& \shortstack[c]{\textbf{0.149}\\[0.3pt]{\small\gainbluedown{0.011}}}
& \shortstack[c]{\textbf{41.5}\\[0.3pt]{\small\gainbluedown{5.5}}} \\

\bottomrule
\end{tabular}
}
\vspace{-8pt}
\label{tab:brats24}
\end{table}

\begin{table}[t]
\centering
\setlength{\tabcolsep}{0.7mm}
\renewcommand\arraystretch{1.0}
\caption{\textbf{Quantitative results for 3D medical image translation on SynthRAD 2025 and AutoPET}.}
\vspace{-8pt}
\resizebox{\linewidth}{!}{
\begin{tabular}{lcccc|cccc|cccc|cccc}
\toprule

\multirow{3}{*}{\textbf{Methods}} 
& \multicolumn{8}{c|}{\textbf{SynthRAD 2025}} 
& \multicolumn{8}{c}{\textbf{AutoPET}} \\

\cmidrule(lr){2-9}
\cmidrule(lr){10-17}

& \multicolumn{4}{c|}{\textbf{MRI$\rightarrow$CT}} 
& \multicolumn{4}{c|}{\textbf{CBCT$\rightarrow$CT}} 
& \multicolumn{4}{c|}{\textbf{CT$\rightarrow$PET}}
& \multicolumn{4}{c}{\textbf{PET$\rightarrow$CT}} \\

\cmidrule(lr){2-5}
\cmidrule(lr){6-9}
\cmidrule(lr){10-13}
\cmidrule(lr){14-17}

& PSNR$\uparrow$ 
& SSIM$\uparrow$ 
& LPIPS$\downarrow$ 
& FID$\downarrow$ 
& PSNR$\uparrow$ 
& SSIM$\uparrow$ 
& LPIPS$\downarrow$ 
& FID$\downarrow$ 
& PSNR$\uparrow$ 
& SSIM$\uparrow$ 
& LPIPS$\downarrow$ 
& FID$\downarrow$
& PSNR$\uparrow$ 
& SSIM$\uparrow$ 
& LPIPS$\downarrow$ 
& FID$\downarrow$ \\

\midrule
\rowcolor{tablegray!70}
\multicolumn{17}{c}{\textit{GAN-based Methods}} \\

3D-mADUNet~\citep{10944137}
& 20.78 & 0.701 & 0.238 & 95.7
& 22.94 & 0.737 & 0.128 & 43.2
& 24.73 & 0.902 & 0.130 & 56.0
& 23.49 & 0.896 & 0.105 & 91.3 \\

MMHVAE~\citep{11197215}
& 20.49 & 0.697 & 0.249 & 101.5
& 22.73 & 0.736 & 0.131 & 46.6
& 25.57 & 0.914 & 0.106 & 53.1
& 23.71 & 0.899 & 0.103 & 89.4  \\

\midrule
\rowcolor{tablegray!70}
\multicolumn{17}{c}{\textit{Diffusion-based Methods}} \\

M2DN~\citep{10444695}
& 20.55 & 0.701 & 0.242 & 97.5
& 22.66 & 0.734 & 0.133 & 43.3
& 25.69 & 0.912 & 0.104 & 52.4
& 23.81 & 0.902 & 0.095 & 86.0 \\

selfRDB~\citep{ARSLAN2025103747}
& 20.81 & 0.708 & 0.236 & 93.6
& 23.19 & 0.740 & 0.127 & 39.2
& 26.08 & 0.923 & 0.094 & 51.1
& 23.67 & 0.901 & 0.099 & 84.6 \\

CRSG~\citep{11559110}
& 20.62 & 0.694 & 0.248 & 97.3
& 22.78 & 0.732 & 0.136 & 42.4
& 26.85 & 0.937 & 0.080 & 51.7
& 23.98 & 0.903 & 0.095 & 78.2 \\

\midrule

\rowcolor{tablegreen!60}
Source-only
& 20.27 & 0.691 & 0.252 & 98.8
& 23.02 & 0.739 & 0.129 & 40.5
& 27.30 & 0.939 & 0.078 & 49.5
& 23.58 & 0.901 & 0.097 & 81.9 \\

\rowcolor{tableblue!70}
Full-prefix AR
& 20.73 & 0.705 & 0.232 & 93.2
& 23.35 & 0.744 & 0.123 & 37.6
& 27.93 & 0.947 & 0.066 & 45.7
& 24.15 & 0.905 & 0.093 & 75.7  \\

\rowcolor{tablepurple!70}
\shortstack[l]{
    \textbf{{\spira}}\\[0.3pt]
    {\small\textcolor{gainblue}{\textbf{$\Delta$ vs. Best External}}}
}
& \shortstack[c]{\textbf{21.29}\\[0.3pt]{\small\gainblueup{0.48}}}
& \shortstack[c]{\textbf{0.714}\\[0.3pt]{\small\gainblueup{0.006}}}
& \shortstack[c]{\textbf{0.225}\\[0.3pt]{\small\gainbluedown{0.011}}}
& \shortstack[c]{\textbf{87.3}\\[0.3pt]{\small\gainbluedown{6.3}}}

& \shortstack[c]{\textbf{23.59}\\[0.3pt]{\small\gainblueup{0.40}}}
& \shortstack[c]{\textbf{0.751}\\[0.3pt]{\small\gainblueup{0.011}}}
& \shortstack[c]{\textbf{0.119}\\[0.3pt]{\small\gainbluedown{0.008}}}
& \shortstack[c]{\textbf{35.6}\\[0.3pt]{\small\gainbluedown{3.6}}}

& \shortstack[c]{\textbf{28.44}\\[0.3pt]{\small\gainblueup{1.59}}}
& \shortstack[c]{\textbf{0.956}\\[0.3pt]{\small\gainblueup{0.019}}}
& \shortstack[c]{\textbf{0.058}\\[0.3pt]{\small\gainbluedown{0.022}}}
& \shortstack[c]{\textbf{44.6}\\[0.3pt]{\small\gainbluedown{6.5}}}

& \shortstack[c]{\textbf{24.76}\\[0.3pt]{\small\gainblueup{0.78}}}
& \shortstack[c]{\textbf{0.912}\\[0.3pt]{\small\gainblueup{0.009}}}
& \shortstack[c]{\textbf{0.086}\\[0.3pt]{\small\gainbluedown{0.009}}}
& \shortstack[c]{\textbf{69.5}\\[0.3pt]{\small\gainbluedown{8.7}}} \\

\bottomrule
\end{tabular}
}
\vspace{-12pt}
\label{tab:synthrad_autopet}
\end{table}

\subsection{Comparison with State-of-the-Art (RQ1)}
\label{sec:exp_performance}

\begin{figure*}[t]
\centering
\includegraphics[width=1.00\linewidth]{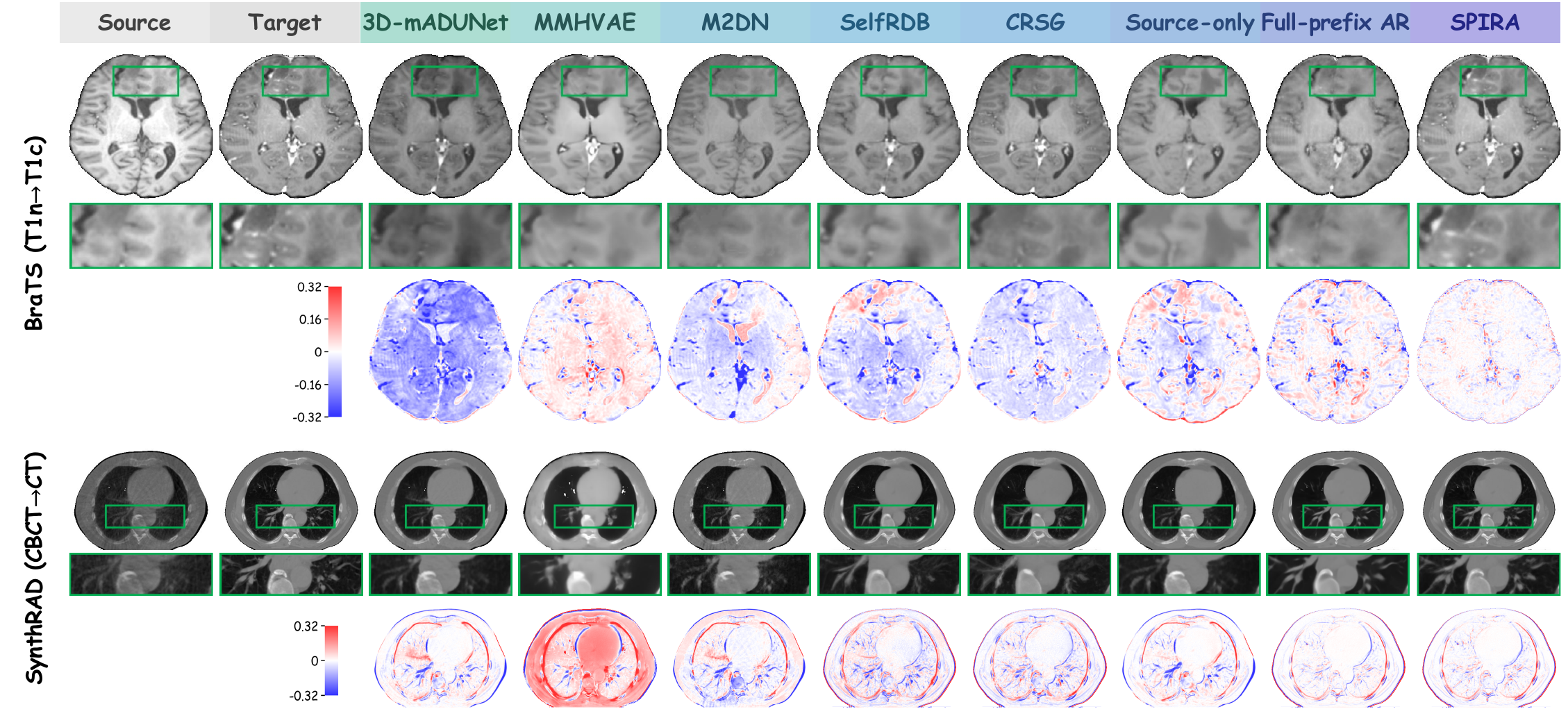}
\vspace{-16pt}
\caption{\textbf{Qualitative comparison} on BraTS T1n\(\rightarrow\)T1c and SynthRAD CBCT\(\rightarrow\)CT cases.}
\vspace{-4pt}
\label{fig:qualitative}
\end{figure*}

\begin{wraptable}{r}{0.55\textwidth}
\centering
\vspace{-14pt}
\setlength{\tabcolsep}{0.5mm}
\renewcommand{\arraystretch}{1.0}

\caption{\textbf{Ablation studies} on BraTS 2024 T1n$\rightarrow$T1c
and SynthRAD 2025 MRI$\rightarrow$CT.}
\vspace{-8pt}

\resizebox{\linewidth}{!}{
\begin{tabular}{lcccc|cccc}
\toprule

\multirow{2}{*}{\textbf{Variants}}
& \multicolumn{4}{c|}{\textbf{BraTS T1n$\rightarrow$T1c}}
& \multicolumn{4}{c}{\textbf{SynthRAD MRI$\rightarrow$CT}} \\

\cmidrule(lr){2-5}
\cmidrule(lr){6-9}

& PSNR$\uparrow$
& SSIM$\uparrow$
& LPIPS$\downarrow$
& FID$\downarrow$
& PSNR$\uparrow$
& SSIM$\uparrow$
& LPIPS$\downarrow$
& FID$\downarrow$ \\

\midrule

\rowcolor{tablegray!70}
\multicolumn{9}{c}{\textit{Reference Model}} \\

Standard AR
& 22.16 & 0.696 & 0.137 & 42.5
& 20.89 & 0.705 & 0.233 & 89.8 \\


\midrule

\rowcolor{tablegray!70}
\multicolumn{9}{c}{\textit{Retraction Strategy}} \\

Tuned global scalar $s$
& 22.28 & 0.699 & 0.133 & 42.8
& 21.02 & 0.709 & 0.230 & 91.6 \\

Position-only gate
& 22.45 & 0.698 & 0.131 & 41.0
& 21.13 & 0.706 & 0.228 & 89.1\\

Effect-only gate
& 22.31 & 0.700 & 0.131 & 40.3
& 20.82 & 0.705 & 0.230 & 90.7 \\

\midrule

\rowcolor{tablegray!70}
\multicolumn{9}{c}{\textit{Gate Design}} \\



Direct $s^\ast$ regression
& 22.60 & 0.701 & 0.129 & 39.9
& 21.04 & 0.707 & 0.230 & 89.3 \\

\midrule

\rowcolor{tablepurple!70}
\textbf{{\spira}}
& \textbf{22.72} & \textbf{0.705} & \textbf{0.126} & \textbf{38.3}
& \textbf{21.29} & \textbf{0.714} & \textbf{0.225} & \textbf{87.3} \\

\bottomrule
\end{tabular}
}
\label{tab:ablation}
\vspace{-8pt}
\end{wraptable}

\textbf{Quantitative Results}.
As reported in Tables~\ref{tab:brats24} and~\ref{tab:synthrad_autopet}, {\spira} consistently achieves the best performance across all tasks and metrics, outperforming both representative GAN- and diffusion-based methods and the matched internal baselines.
In particular, {\spira} consistently improves upon Full-prefix AR across all tasks and metrics, demonstrating the effectiveness of our prefix retraction strategy. Please refer to Appendix~\ref{app:exp} for annotation-defined tumor/lesion ROI evaluation, statistical significance, and downstream segmentation experiments.

\textbf{Qualitative Comparison}.
Fig.~\ref{fig:qualitative} provides illustrative cases in which {\spira} shows lower local residual errors while maintaining target-specific appearance relative to the baselines.

\subsection{Ablation Studies (RQ2)}
\label{sec:exp_ablation}

We compare {\spira} with standard AR, global retraction, position-only and effect-only gates, and direct \(s^\ast\) regression (Appendix~\ref{app:exp_ablation}).
Table~\ref{tab:ablation} shows that joint training or effect alone does not explain the gain, while {\spira} performs best.
Global amplification also falls below Full-prefix AR on both tasks (Table~\ref{tab:additional_ablation}), while token-wise retraction outperforms coarser sharing (Table~\ref{tab:ablation_granularity}).

\subsection{Explaining the Utility--Effect Gap in Source--Prefix Balance (RQ3)}
\label{sec:exp_phenomenon}

\begin{figure*}[t]
\centering
\includegraphics[width=1.00\linewidth]{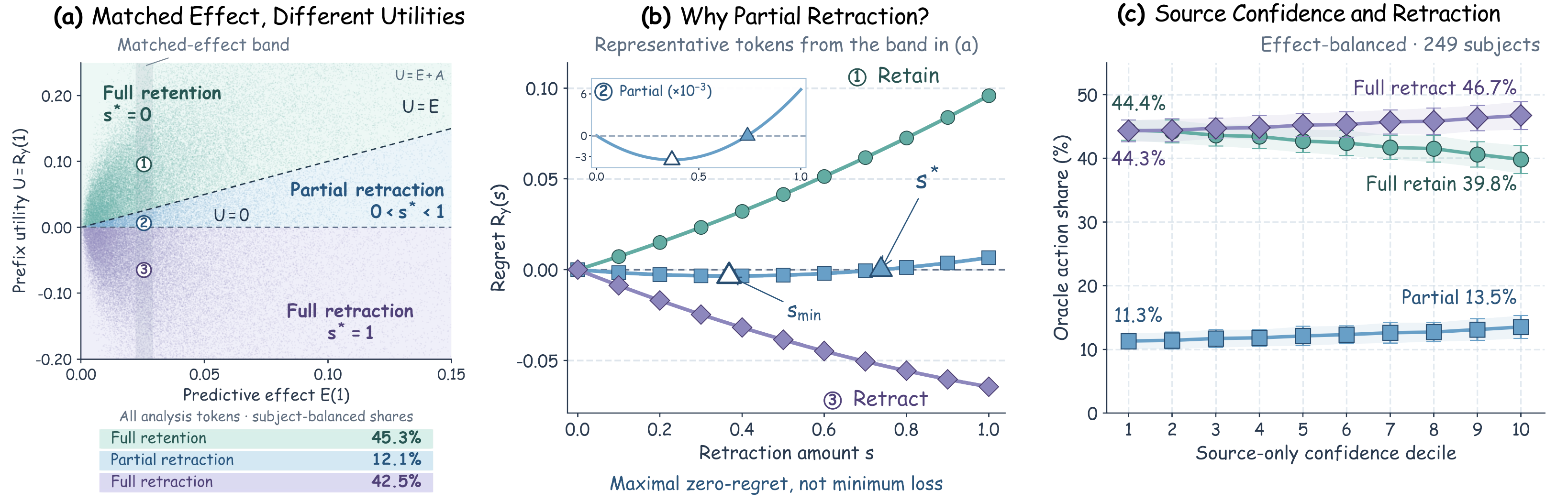}
\vspace{-12pt}
\caption{\textbf{Explaining the utility--effect gap.}
(a) Effect--utility geometry separates three oracle regimes.
(b) Matched-effect regret paths distinguish maximal zero-regret
from minimum-loss retraction.
(c) Source-confidence groups exhibit a modest shift in oracle
regime shares after effect balancing.
Shares in (c) are effect-standardized, whereas (a) reports
marginal subject-balanced shares.}
\vspace{-12pt}
\label{fig:utility_effect}
\end{figure*}

Sec.~\ref{sec:motivation} reveals that corrections with similar predictive effects can have different utilities. We now explain this gap on the frozen dual-endpoint predictor under teacher forcing, using the GT prefix and no gate, with effect measured at the token level by the endpoint KL $E_i(1)$.

\textbf{Why can matched effects have different utilities?}
Let $U_i=R_{i,y}(1)$ denote prefix utility, with $U_i>0$ indicating that the prefix lowers target loss. By Eq.~\ref{eq:regret_decomposition}, $U_i=E_i(1)+A_i$, where $A_i=r_{i,y}-\mathbb{E}_{k\sim p_i^F}[r_{i,k}]$ measures target alignment; $A_i>0$ means that the correction favors the true target relative to the full-prefix belief. For $E_i(1)>0$, convexity of $R_{i,y}$ implies that $A_i\geq0$ admits no safe positive retraction, whereas $A_i<0$ reduces target loss. This gives full retention for $U_i\geq E_i(1)$, partial retraction for $0<U_i<E_i(1)$, and full retraction for $U_i\leq0$. Fig.~\ref{fig:utility_effect}(a) reports corresponding shares of $45.3\%$, $12.1\%$, and $42.5\%$; the heterogeneity persists across effect levels in Fig.~\ref{fig:app_utility_effect}(a).

\textbf{Why partially retract a useful correction?}
Fig.~\ref{fig:utility_effect}(b) traces corrections from the matched-effect band.
In the partial regime, maximal zero-regret preserves the Full-prefix target loss while retaining less correction than minimum-loss.
At inference, generated-prefix corrections may carry errors, whose influence is also scaled by $1-s$ in Eq.~\ref{eqn:retraction_path}.
This predicts slower error growth with stronger safe retraction, a prediction we test in Sec.~\ref{sec:exp_mechanism} and Fig.~\ref{fig:app_actionability}(c).


\textbf{Is this variation associated with source evidence?}
After effect balancing, higher source-only confidence only slightly reduces the full-retention share, from 44.4\% to 39.8\% (Fig.~\ref{fig:utility_effect}(c)).
Combining the GT-free source and correction features predicts oracle choices much better than confidence or effect alone (macro AUROC 0.80--0.85; Fig.~\ref{fig:app_actionability}(a)).

\begin{insightbox}{Takeaway}
Target alignment explains why corrections with similar effects can favor
different source--prefix balances, and even useful prefix corrections do not always need to be retained in full.
\end{insightbox}

\subsection{From Utility--Effect Gap to Performance Gains (RQ4)}
\label{sec:exp_mechanism}

\begin{figure*}[t]
\centering
\includegraphics[width=1.00\linewidth]{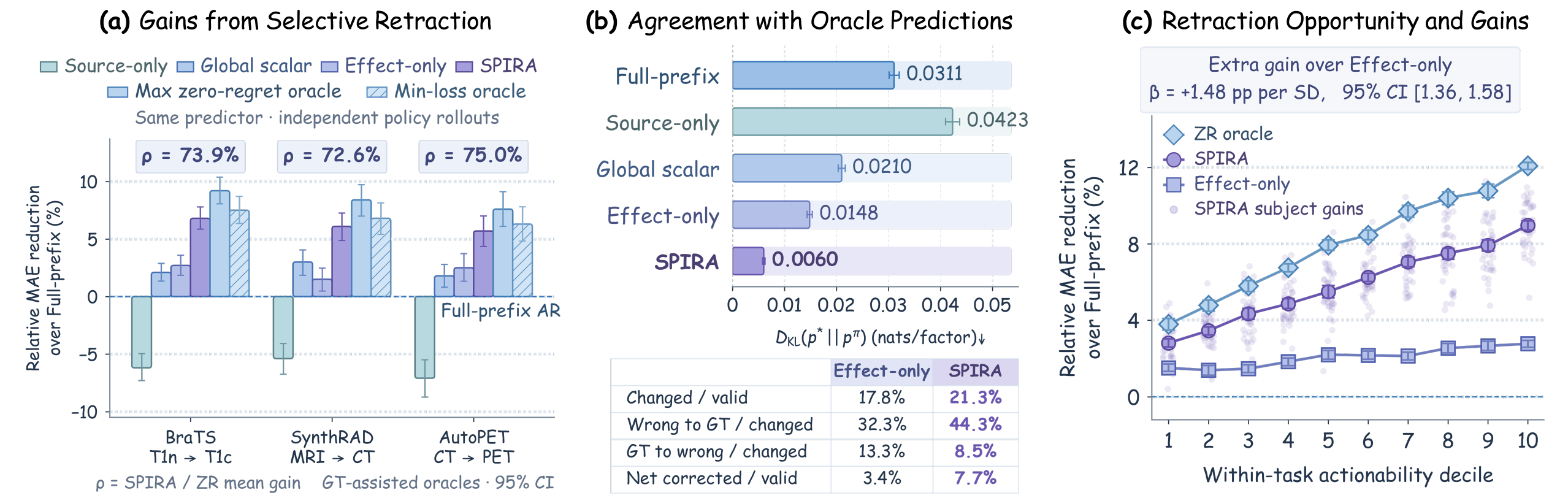}
\vspace{-16pt}
\caption{\textbf{From selective retraction to generation gains.}
(a) Free-running gains of different retraction policies, where $\rho$ denotes the zero-regret oracle gain recovered by {\spira}.
(b) Oracle-distribution KL and prediction changes on identical {\spira}-generated states.
(c) Generation gains across baseline-actionability deciles, with $\beta$
measuring {\spira}'s extra gain over effect-only.}
\vspace{-16pt}
\label{fig:gap_gain}
\end{figure*}

Following the teacher-forced analysis, we now examine whether adaptively
balancing source and prefix information through selective retraction improves free-running generation on three tasks. 


\textbf{Can Selective Retraction Improve Generation?}
In Fig.~\ref{fig:gap_gain}(a), {\spira} outperforms source-only, global scalar, and effect-only, reducing MAE over Full-prefix AR by 6.8\%, 6.1\%, and 5.7\% on BraTS, SynthRAD, and AutoPET (6.2\% on average), while recovering 72.6\%--75.0\% of the maximal zero-regret oracle gain.
We further compare this oracle with a minimum-loss oracle that chooses the lowest-loss action at each current state (Appendices~\ref{app:train_infer} and~\ref{app:exp_ablation}).
Although minimum-loss achieves lower local loss, maximal zero-regret yields better free-running generation with slower error growth (Fig.~\ref{fig:app_actionability}(b,c)); its distilled gate also performs better (Table~\ref{tab:additional_ablation}).

\textbf{Does {\spira} Recover Oracle Predictions?}
On identical rollout states, {\spira} better matches the oracle than effect-only, with a KL divergence of 0.0060 versus 0.0148 nats per factor (Fig.~\ref{fig:gap_gain}(b)). We define prediction changes as factor-wise argmax differences from the unretracted prediction at the same state. Corrected errors minus newly introduced errors account for $7.7\%$ of valid factor decisions for {\spira}, compared with $3.4\%$ for effect-only. Since the gate is trained under teacher forcing, this agreement also suggests that its learned decisions transfer to generated-prefix states.

\textbf{Do More Actionable Cases Benefit More?}
We measure baseline actionability on Full-prefix rollout states as the fraction of endpoint predictive effect that can be safely retracted by the GT-assisted oracle.
Higher actionability therefore means that the current-state oracle can
retract more of the prefix correction without increasing local target
loss.
In Fig.~\ref{fig:gap_gain}(c), from the lowest to highest actionability decile, {\spira}'s gain rises from 2.81\% to 8.97\%, while effect-only changes only from 1.51\% to 2.77\%, widening {\spira}'s extra gain from 1.30 to 6.20 percentage points.
Consistently, this extra gain increases by 1.48 percentage points per standard deviation of actionability.

\begin{figure*}[t]
\centering
\includegraphics[width=1.00\linewidth]{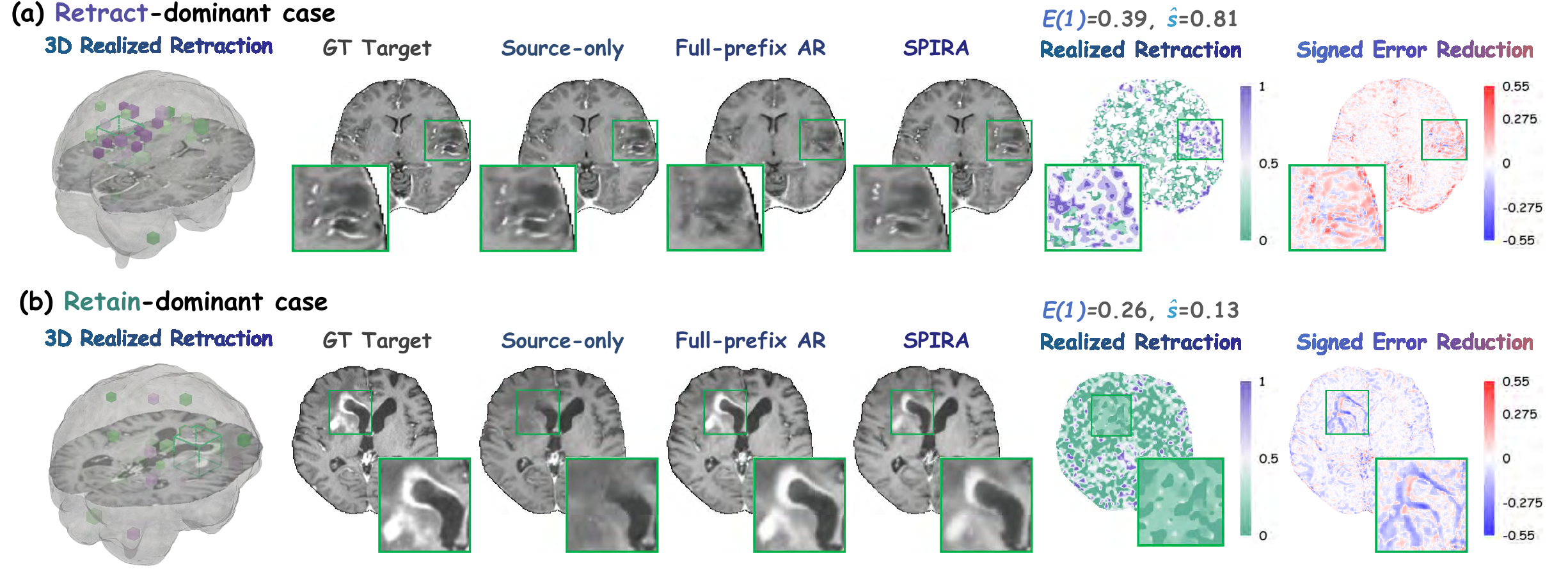}
\vspace{-6pt}
\caption{\textbf{Visualization of selective prefix retraction.}
Retract-dominant (top) and retain-dominant (bottom) cases illustrate how {\spira} \retractsem{retracts harmful corrections} while \retainsem{retaining useful ones}. $E(1)$ and $\hat{s}$ are region averages, with \retainsem{green} indicating stronger retention and \retractsem{purple} stronger retraction.}
\vspace{-16pt}
\label{fig:retraction}
\end{figure*}

Fig.~\ref{fig:retraction} illustrates two characteristic source--prefix balancing behaviors.
In the retract-dominant case (top), stronger retraction
\retractsem{removes harmful prefix corrections} and reduces local
reconstruction errors. In the retain-dominant case (bottom), weaker
retraction \retainsem{preserves target-useful prefix information}
where source-only prediction misses target-specific details. Together, the two cases illustrate that {\spira} adaptively retracts or retains prefix corrections according to local prediction needs.

\begin{insightbox}{Takeaway}
{\spira} improves generation by
\retainsem{retaining target-useful prefix information} while \retractsem{removing harmful corrections}. {\spira} better matches oracle decisions, with larger gains when more correction can be retracted.
\end{insightbox}

\section{Related Work}

\textbf{Medical Image Translation.}
Medical image translation has been widely studied with GANs~\citep{10944137, 11197215} and diffusion models~\citep{PANG2026114148, liu2026exploiting, 10.1145/3746027.3754923}. 
Most methods perform direct or iterative source-conditioned generation~\citep{Zhan_2024_CVPR, ARSLAN2025103747, zhu2025introducing, Wu_2026_CVPR}, whereas AR translation exposes an evolving target prefix~\citep{Liu_2025_ICCV, NEURIPS2025_f4823f83, Jin_2026_CVPR}. 
{\spira} studies its region-wise interaction with the fixed source condition and learns how much prefix correction to retain.

\textbf{Autoregressive Visual Generation.}
Autoregressive visual generation has progressed from pixel-level to token-based modeling~\citep{NEURIPS2024_9a24e284, Ren_2025_ICCV, wang2024continuous}, and has recently expanded to videos and structured visual data~\citep{NEURIPS2025_f4823f83, Yin_2025_CVPR, ren2025next, yang2026longlive} as well as 3D settings~\citep{he2026medvar, Feng_2026_CVPR, NEURIPS2025_e19b6f65}. 
Most existing work focuses on visual representation and prediction, while the interaction between fixed conditioning and evolving AR context is underexplored~\citep{chen2026context, NEURIPS2025_2bde8fef, gu2025long}. 
For conditional translation, where the source and target prefix provide complementary information~\citep{Mu_2025_CVPR}, {\spira} learns how much of each prefix correction to retain region-wise.

\textbf{Guidance and Context Control in Generative Models.}
Classifier-free guidance contrasts conditional and unconditional predictions~\citep{ho2022classifier, Shen_2024_CVPR, NEURIPS2025_73c90c0c}, while contrastive and context-aware decoding adjust generation through prediction contrasts~\citep{li-etal-2023-contrastive, shi-etal-2024-trusting, ICLR2024_edc36117}. 
Some recent approaches further incorporate confidence-based routing or extrapolation~\citep{tao-agrawal-2026-worse, liao2026vpg}, whereas {\spira} learns continuous token-wise prefix retraction by distilling a target-aware zero-regret oracle.

\section{Conclusion}
\label{sec:conclusion}

In this work, we study source-conditioned AR translation as a region-wise balance between fixed source evidence and evolving context. 
We propose {\spira}, which treats the target prefix as a controllable correction and learns to \retractsem{reduce harmful or unnecessary corrections} while \retainsem{preserving target-useful prefix information}.
Across diverse 3D medical image translation tasks, {\spira}
consistently improves performance. Our analyses link different retention needs at similar predictive effects to target alignment, while {\spira} better matches target-aware oracle predictions and gains more when greater retraction is possible. We hope {\spira} motivates broader consideration of how fixed conditioning and AR context should be balanced in conditional generation.

\subsubsection*{AI Use Statement}
Generative AI tools were used to aid and polish manuscript writing, support literature retrieval and discovery, including identifying relevant related work, and provide assistance during research ideation and experimental development. All AI-assisted suggestions and outputs were independently reviewed and verified by the authors. The research ideas, methodological design, experimental results, and scientific claims were developed and validated by the authors, who take full responsibility for the final content of the paper.


\bibliography{iclr2027_conference}
\bibliographystyle{iclr2027_conference}

\clearpage
\clearpage

\begin{center}
    {\LARGE\bfseries Appendix Contents}
\end{center}

{
\hypersetup{linktoc=page}
\startcontents[appendix]
\printcontents[appendix]{l}{1}{\setcounter{tocdepth}{2}}
}

\clearpage

\appendix
\section*{Appendix}

\section{Datasets}
\label{app:dataset}

\textbf{BraTS 2024.}
We use three cohorts from the BraTS 2024~\citep{de20242024} collection, including post-treatment adult glioma (GLI), brain metastases (MET), and pediatric brain tumors (PED).
Each case provides four spatially aligned MRI modalities, T1n, T1c, T2w, and T2f, together with tumor annotations.
Tumor annotations are used only for ROI and downstream evaluation.
We split the subjects into training and validation sets at a 9:1 ratio.

\textbf{SynthRAD 2025.}
SynthRAD 2025~\citep{https://doi.org/10.1002/mp.17981} is a multi-center radiotherapy dataset containing paired MRI--CT and CBCT--CT scans across multiple anatomical regions.
We use the head-and-neck and abdominal subsets.
The released image pairs are spatially aligned through dataset preprocessing, providing paired 3D source and target volumes for translation.
We follow the official training and validation splits.

\textbf{AutoPET.}
AutoPET~\citep{dexl2026autopet3} provides paired whole-body PET/CT examinations with tumor-lesion annotations and broad anatomical coverage.
We use the corresponding CT and PET volumes to evaluate CT$\rightarrow$PET and PET$\rightarrow$CT translation, covering diverse anatomical regions including the brain, pelvis, and abdomen.
Lesion annotations are reserved for ROI-based evaluation and are not provided to the translation model.
We follow the official training and validation splits.

The dataset splits and the number of training and validation cases for each translation task are summarized in Table~\ref{tab:dataset_statistics}. For model selection, we reserve a subject-disjoint subset from the training partition, while the validation subjects in Table~\ref{tab:dataset_statistics} are used only for final evaluation. All model and hyperparameter choices are fixed before evaluation.

\begin{table}[h]
\centering
\caption{\textbf{Dataset composition and statistics} used in our experiments. The held-out evaluation subjects are not used for model training or selection.}
\setlength{\tabcolsep}{2.5mm}
\resizebox{0.75\linewidth}{!}{
\begin{tabular}{lccc}
\toprule
\textbf{Dataset} & \textbf{Translation Task} & \textbf{\#Train} & \textbf{\#Val} \\
\midrule

\multirow{4}{*}{BraTS 2024~\citep{de20242024}}
& T1n$\rightarrow$T1c & 2,280 & 254 \\
& T1c$\rightarrow$T1n & 2,280 & 254 \\
& T2w$\rightarrow$T2f & 2,280 & 254 \\
& T2f$\rightarrow$T1c & 2,280 & 254 \\

\midrule

\multirow{2}{*}{SynthRAD 2025~\citep{https://doi.org/10.1002/mp.17981}}
& MRI$\rightarrow$CT & 407 & 106 \\
& CBCT$\rightarrow$CT & 622 & 152 \\

\midrule

\multirow{2}{*}{AutoPET~\citep{dexl2026autopet3}}
& CT$\rightarrow$PET & 1,279 & 323 \\
& PET$\rightarrow$CT & 1,279 & 323 \\

\bottomrule
\end{tabular}
}
\label{tab:dataset_statistics}
\end{table}

\section{Additional Implementation and Training Details}
\label{app:imple}

\subsection{Visual Tokenizer Fine-Tuning}
\label{app:vidtok_finetune}

We initialize the tokenizer from the pre-trained non-causal VidTok-FSQ-262144 checkpoint~\citep{tang2024vidtok} and fine-tune one modality-shared tokenizer for each dataset, such that all source and target modalities within the same dataset share a common latent vocabulary. 

To adapt VidTok to volumetric data, we treat each 3D volume as a video, with slices along the depth axis regarded as video frames such that the depth dimension corresponds to the temporal dimension. For tokenizer fine-tuning, we follow the same image preprocessing as in the main experiments. Grayscale medical volumes are repeated to three channels before VidTok encoding, while the three-channel decoder output is averaged back to a single-channel reconstruction. We retain the original $8\times8\times4$ compression ratio and FSQ-262144 quantization configuration, and optimize the original VidTok objective together with grayscale $\ell_1$ and 3D SSIM reconstruction losses. We first adapt the decoder while keeping the encoder and FSQ module fixed, and subsequently jointly fine-tune the encoder and decoder with a smaller learning rate for the encoder. The complete fine-tuning configurations are summarized in Table~\ref{tab:vidtok_setting}. 

After fine-tuning, the tokenizer is frozen throughout autoregressive training and inference, and Table~\ref{tab:vidtok_rec} compares the reconstruction quality of the pre-trained and fine-tuned tokenizers for each modality.

\begin{table}[t]
\centering
\caption{Fine-tuning configurations for the VidTok tokenizer.}
\setlength{\tabcolsep}{2.0mm}
\resizebox{0.55\linewidth}{!}{
\renewcommand{\arraystretch}{1.0}
\begin{tabular}{ll}
\toprule
\textbf{Setting} & \textbf{Value} \\
\midrule
Base tokenizer & VidTok-FSQ-262144, non-causal \\
Compression ratio & $8\times8\times4$ \\
Stage I & 0--5k steps, decoder only \\
Stage II & 5k--30k steps, encoder + decoder \\
Encoder learning rate & $1\times10^{-5}$ \\
Decoder learning rate & $5\times10^{-5}$ \\
Optimizer & AdamW \\
Weight decay & 0.01 \\
Batch size & 4 \\
Warm-up steps & 1,000 \\
Gradient clipping & 1.0 \\
Precision & bfloat16 \\
\bottomrule
\end{tabular}
}
\label{tab:vidtok_setting}
\end{table}

\begin{table}[t]
\centering
\caption{\textbf{Reconstruction performance} of the pre-trained and
fine-tuned VidTok tokenizers. Averages are computed over all evaluated
volumes before rounding.}
\setlength{\tabcolsep}{1.5mm}
\renewcommand{\arraystretch}{1.0}
\resizebox{0.85\linewidth}{!}{
\begin{tabular}{llccc|ccc}
\toprule
\multirow{2}{*}{\textbf{Dataset}} &
\multirow{2}{*}{\textbf{Modality}} &
\multicolumn{3}{c|}{\textbf{Pre-trained VidTok}} &
\multicolumn{3}{c}{\textbf{Fine-tuned VidTok}} \\
\cmidrule(lr){3-5}
\cmidrule(lr){6-8}
& & PSNR$\uparrow$ & SSIM$\uparrow$ & LPIPS$\downarrow$
  & PSNR$\uparrow$ & SSIM$\uparrow$ & LPIPS$\downarrow$ \\
\midrule

\multirow{5}{*}{BraTS 2024}
& T1n & 32.66 & 0.926 & 0.077
& 33.98 {\scriptsize\gainblueup{1.32}}
& 0.950 {\scriptsize\gainblueup{0.024}}
& 0.069 {\scriptsize\gainbluedown{0.008}} \\

& T1c & 32.98 & 0.938 & 0.067
& 33.61 {\scriptsize\gainblueup{0.63}}
& 0.940 {\scriptsize\gainblueup{0.002}}
& 0.060 {\scriptsize\gainbluedown{0.007}} \\

& T2w & 33.80 & 0.951 & 0.057
& 34.80 {\scriptsize\gainblueup{1.00}}
& 0.960 {\scriptsize\gainblueup{0.009}}
& 0.051 {\scriptsize\gainbluedown{0.006}} \\

& T2f & 32.51 & 0.924 & 0.074
& 33.74 {\scriptsize\gainblueup{1.23}}
& 0.941 {\scriptsize\gainblueup{0.017}}
& 0.070 {\scriptsize\gainbluedown{0.004}} \\

& Avg & 32.99 & 0.934 & 0.068
& 34.03 {\scriptsize\gainblueup{1.04}}
& 0.948 {\scriptsize\gainblueup{0.014}}
& 0.062 {\scriptsize\gainbluedown{0.006}} \\

\midrule

\multirow{4}{*}{SynthRAD 2025}
& MRI & 28.16 & 0.790 & 0.122
& 34.05 {\scriptsize\gainblueup{5.89}}
& 0.943 {\scriptsize\gainblueup{0.153}}
& 0.092 {\scriptsize\gainbluedown{0.030}} \\

& CT & 28.49 & 0.833 & 0.125
& 35.52 {\scriptsize\gainblueup{7.03}}
& 0.958 {\scriptsize\gainblueup{0.125}}
& 0.095 {\scriptsize\gainbluedown{0.030}} \\

& CBCT & 29.38 & 0.807 & 0.147
& 35.72 {\scriptsize\gainblueup{6.34}}
& 0.937 {\scriptsize\gainblueup{0.130}}
& 0.108 {\scriptsize\gainbluedown{0.039}} \\

& Avg & 28.69 & 0.816 & 0.131
& 35.28 {\scriptsize\gainblueup{6.59}}
& 0.949 {\scriptsize\gainblueup{0.133}}
& 0.098 {\scriptsize\gainbluedown{0.033}} \\

\midrule

\multirow{3}{*}{AutoPET}
& CT & 30.68 & 0.885 & 0.101
& 35.36 {\scriptsize\gainblueup{4.68}}
& 0.959 {\scriptsize\gainblueup{0.074}}
& 0.081 {\scriptsize\gainbluedown{0.020}} \\

& PET & 35.25 & 0.954 & 0.041
& 36.28 {\scriptsize\gainblueup{1.03}}
& 0.971 {\scriptsize\gainblueup{0.017}}
& 0.035 {\scriptsize\gainbluedown{0.006}} \\

& Avg & 32.97 & 0.920 & 0.071
& 35.81 {\scriptsize\gainblueup{2.84}}
& 0.965 {\scriptsize\gainblueup{0.045}}
& 0.058 {\scriptsize\gainbluedown{0.013}} \\

\bottomrule
\end{tabular}
}
\label{tab:vidtok_rec}
\end{table}

\subsection{Latent Pre-computation and Caching}
\label{app:latent_cache}

Before AR training, we use the fine-tuned and frozen VidTok tokenizer~\citep{tang2024vidtok} to pre-compute full-volume latent representations for all modalities, which are subsequently stored as local caches. Each modality is encoded and cached independently, while all modalities from the same case share identical preprocessing, spatial grids, padding, and patch coordinates to preserve voxel-wise correspondence. For variable-sized volumes, we employ overlapping tiled encoding with center-crop stitching to construct seamless full-volume latents without averaging the discrete FSQ scalar indices. Each cache stores the factorized FSQ scalar indices together with the corresponding latent valid mask, and autoregressive training directly samples fixed-size latent windows from these caches without repeatedly invoking the VidTok encoder.

\subsection{Blockwise AR Architecture}
\label{app:blockwise_AR}

We instantiate \(f_\theta\) as a source-conditioned 3D blockwise Transformer adapted from LlamaGen~\citep{sun2024autoregressive}.
The spatially aligned source and target FSQ grids are partitioned into the same non-overlapping 3D blocks and traversed in a fixed depth--height--width raster order.
The architecture is causal across blocks but bidirectional within each block, allowing all spatial tokens in the current target block to be predicted in parallel.

\textbf{Source and prefix encoding.}
At each source position, embeddings of the factorized FSQ scalars are summed to form a token representation.
A block-local bidirectional Transformer first contextualizes tokens within each source block, followed by attention pooling into one representation per block.
A global bidirectional Transformer with 3D coordinate embeddings then produces source block memory \(M^S\).
The resulting global block features are also projected back to the local tokens, yielding spatially aligned dense source features \(U^S\).

Visible target blocks are embedded and encoded using block-local bidirectional attention, and their pooled representations form the inter-block sequence.
A Llama-style causal Transformer processes this sequence using 3D rotary positional embeddings.
Each layer applies
\begin{equation}
\begin{aligned}
H &\leftarrow H+
\operatorname{CausalAttn}
\bigl(\operatorname{RMSNorm}(H);
      \operatorname{RoPE}_{3D}\bigr),\\
H &\leftarrow H+
\operatorname{CrossAttn}
\bigl(\operatorname{RMSNorm}(H),M^S\bigr),\\
H &\leftarrow H+
\operatorname{SwiGLU}
\bigl(\operatorname{RMSNorm}(H)\bigr),
\end{aligned}
\end{equation}
where source cross-attention uses a factorized relative 3D positional bias.

\textbf{Next-block prediction.}
The first target block is predicted from a virtual source-conditioned start block, which is processed through the same intra- and inter-block modules as ordinary target history.
For each subsequent block, a local bidirectional decoder combines the causal block context, the encoded preceding block, the aligned source features \(U_i^S\), intra-block positional embeddings, and the 3D transition between consecutive blocks.
This produces a context vector \(c_{i,v}\) for each spatial position \(v\) in block \(i\).
The source-only and full-prefix predictions use exactly the same architecture and differ only in whether the preceding target blocks are visible.

\textbf{Factorized FSQ prediction.}
Each latent token is represented by \(d\) low-cardinality FSQ factors
\(q_{i,v,1:d}\).
Rather than predicting their Cartesian-product code with a single large classifier, the output head factorizes the token distribution as
\begin{equation}
p_\theta(Y_{i,v}\mid S,P_{<i})
=
\prod_{j=1}^{d}
p_\theta
\bigl(q_{i,v,j}\mid c_{i,v},q_{i,v,<j}\bigr).
\end{equation}
The factors are generated sequentially within each token, whereas all spatial positions in the current block are processed in parallel.


\textbf{Correction-aware gate inputs.}
For each valid spatial position, the source-only and full-prefix passes also provide
pre-head context features \(\mathbf{h}_{i,v}^{S}\) and
\(\mathbf{h}_{i,v}^{F}\).
We construct the source and correction features as
\begin{equation}
\mathbf{u}_{i,v}^{S}
=
\operatorname{RMSNorm}\!\bigl(\operatorname{sg}(\mathbf{h}_{i,v}^{S})\bigr),
\qquad
\mathbf{u}_{i,v}^{r}
=
\operatorname{RMSNorm}\!\bigl(
\operatorname{sg}(\mathbf{h}_{i,v}^{F}-\mathbf{h}_{i,v}^{S})
\bigr),
\end{equation}
where \(\operatorname{sg}\) denotes stop-gradient.
Therefore, \(\mathbf{u}^{S}\) represents the source-only prediction state, while
\(\mathbf{u}^{r}\) explicitly captures the hidden-space change introduced by the current prefix.

We further provide two lightweight endpoint-difference statistics:
\(w_{i,v}\), the factor-averaged total-variation distance between the
source-only and full-prefix predictive distributions, and \(e_{i,v}\), the log-RMS magnitude of their logit residual. The gate concatenates
\([\mathbf{u}_{i,v}^{S},\mathbf{u}_{i,v}^{r},w_{i,v},e_{i,v}]\)
and predicts the token-wise retraction amount \(\hat{s}_{i,v}\) with a
lightweight MLP. All gate inputs are available at inference and are detached from the shared AR predictor. The gate maps the concatenated $(2H+2)$-dimensional input through a 128-dimensional linear layer, SiLU, LayerNorm, and a scalar linear output layer followed by sigmoid, where $H$ is the context feature dimension.

\textbf{Oracle construction during training.}
\label{app:oracle_construction}

Under teacher forcing, the shared predictor produces source-only and full-prefix logits
\(z^S_{i,v,j}\) and \(z^F_{i,v,j}\) for FSQ factor \(j\) at each valid position \(v\).
For a candidate retraction \(s\), we form
\begin{equation}
z_{i,v,j}(s)
=
z^F_{i,v,j}
-
s\left(z^F_{i,v,j}-z^S_{i,v,j}\right),
\end{equation}
and evaluate the factor-averaged GT token loss
\begin{equation}
\ell_{i,v}(s)
=
-\frac{1}{d}
\sum_{j=1}^{d}
\log
\left[
\operatorname{softmax}
\left(z_{i,v,j}(s)\right)
\right]_{q^{\mathrm{GT}}_{i,v,j}}.
\end{equation}
At the full-prefix endpoint, $p_i(0)=p_i^F$, giving
$E_i(0)=E_i'(0)=0$. Differentiating Eq.~\ref{eq:regret_decomposition}
then yields $R'_{i,y}(0)=A_i$. The oracle target is the maximal zero-regret retraction
\begin{equation}
s^*_{i,v}
=
\max
\left\{
s\in[0,1]
\;\middle|\;
\ell_{i,v}(s)\leq \ell_{i,v}(0)
\right\},
\end{equation}
where \(\ell_{i,v}(0)\) is the full-prefix loss.
The oracle uses the feasibility criterion $\ell_{i,v}(s)\leq\ell_{i,v}(0)+10^{-6}$. We accept $s=1$ when feasible; otherwise, we perform 10 bisection iterations on $[0,1]$ and return the feasible lower bound.
The resulting factorized distribution at \(s^*_{i,v}\) is detached and used as the supervision target for the gate rather than directly regressing \(s^*_{i,v}\).
The exact architectural dimensions and training configurations are provided in Appendix~\ref{app:train_infer}.

\textbf{Factor-wise inference and distillation.}
For the AR FSQ head, token-wise gating is implemented in two passes. We first construct a deterministic greedy factor trajectory using the full-prefix endpoint and evaluate both endpoint heads along this shared trajectory to obtain the gate statistics. The gate then selects one retraction amount per token. Factor generation restarts from an empty history, with both endpoint heads recomputed sequentially using the factors actually generated under the selected action. The action remains fixed within the token, and only the final generated block is appended to the target prefix. Source-only pre-head contexts can be precomputed across blocks, while their conditional factor logits are evaluated using the corresponding within-token history. During teacher-forced training, both endpoints use the same ground-truth factor history when computing gate statistics, oracle actions, and distillation targets. The retraction objective averages $D_{\mathrm{KL}}(p_j^{s^*}\|p_j^{\hat{s}})$ equally over FSQ factors and then over valid tokens. It therefore matches conditional factor distributions at a fixed history, rather than complete autoregressive token distributions. The regret decomposition and maximal zero-regret criterion are defined on the same fixed-history, factor-averaged quantities.

\begin{table}[t]
\centering
\caption{\textbf{Dataset- and modality-specific preprocessing} in our experiments. $P_q$ denotes the $q$th percentile computed independently for each volume and modality, whereas $P_q^{+}$ is computed over positive-valued voxels only.}
\setlength{\tabcolsep}{1.0mm}
\renewcommand{\arraystretch}{1.05}
\resizebox{\linewidth}{!}{%
\begin{tabular}{l*{4}{>{\centering\arraybackslash}p{1.0cm}}*{5}{c}}
\toprule
\textbf{Dataset}
& \multicolumn{4}{c}{\textbf{BraTS 2024}}
& \multicolumn{3}{c}{\textbf{SynthRAD 2025}}
& \multicolumn{2}{c}{\textbf{AutoPET}} \\
\cmidrule(lr){2-5}
\cmidrule(lr){6-8}
\cmidrule(lr){9-10}
\textbf{Modality}
& T1n & T1c & T2w & T2f
& CT & CBCT & MRI
& CT & PET \\
\midrule

Orientation
& \multicolumn{4}{c}{RAS}
& \multicolumn{3}{c}{RAS}
& \multicolumn{2}{c}{RAS} \\

Voxel spacing
& \multicolumn{4}{c}{Adjusted by volume resizing}
& \multicolumn{3}{c}{Native spacing}
& \multicolumn{2}{c}{Native spacing} \\

Foreground threshold
& \multicolumn{4}{c}{$x>0$}
& $x>-950$ & $x>-950$ & $x>0$
& $x>-950$ & $x>0$ \\

Intensity clipping
& \multicolumn{4}{c}{$[P_{0.5},P_{99.5}]$}
& $[-1000,1000]$ & $[-1000,1000]$ & $[P_{0.5}^{+},P_{99.5}^{+}]$
& $[-1000,1000]$ & $[P_{0.5}^{+},P_{99.5}^{+}]$ \\

Normalization
& \multicolumn{4}{c}{$[-1,1]$}
& \multicolumn{3}{c}{$[-1,1]$}
& \multicolumn{2}{c}{$[-1,1]$} \\

Volume resizing
& \multicolumn{4}{c}{$256\times256\times128$}
& \multicolumn{3}{c}{None; variable volume size}
& \multicolumn{2}{c}{None; variable volume size} \\

Augmentation
& \multicolumn{4}{c}{Random rotation and flipping}
& \multicolumn{3}{c}{Random rotation and flipping}
& \multicolumn{2}{c}{Random rotation and flipping} \\

\bottomrule
\end{tabular}
}
\label{tab:preprocessing}
\end{table}

\subsection{Training and Inference Details}
\label{app:train_infer}

\textbf{Preprocessing.}
Because the datasets contain different imaging modalities with distinct
intensity characteristics and spatial configurations, we use a separate
preprocessing pipeline for each dataset and modality, as summarized in
Table~\ref{tab:preprocessing}.
For BraTS 2024~\citep{de20242024}, all volumes are resized to
$256\times256\times128$.
For SynthRAD 2025~\citep{https://doi.org/10.1002/mp.17981} and AutoPET~\citep{dexl2026autopet3}, the native volume dimensions vary
substantially across subjects, and we therefore do not resize the full
volumes to a fixed spatial size. The resulting source and target volumes within each case are processed on the same spatial grid to preserve their correspondence.

\textbf{Tokenization and AR setup.}
We use the fine-tuned non-causal VidTok-FSQ~\citep{tang2024vidtok} described in Appendix~\ref{app:vidtok_finetune} as the visual tokenizer.
Each 3D volume is treated as a video by regarding the depth axis as the
temporal dimension. The tokenizer uses a volumetric compression ratio of
$(8,8,4)$ in $(H,W,D)$, corresponding to $(4,8,8)$ in VidTok's native
$(T,H,W)$ ordering. Before autoregressive training, we pre-compute the latent representation of each full volume and cache it locally, as detailed in Appendix~\ref{app:latent_cache}, avoiding repeated tokenizer encoding during training. The latent grids are partitioned into non-overlapping
$4\times4\times4$ blocks and traversed using a fixed
depth--height--width raster order. Dataset-specific model configurations are summarized in Table~\ref{tab:ar_configuration}.

\textbf{Optimization.}
We train the AR predictor for 100k optimization steps using
AdamW with $(\beta_1,\beta_2)=(0.9,0.95)$ and weight decay $0.05$.
The initial learning rate is $1\times10^{-4}$ and follows a cosine decay
schedule after warmup.
We clip the global gradient norm to $2.0$.
For the gate parameter group, we use a learning rate three times that
of the AR predictor, i.e., $3\times10^{-4}$, with weight decay $0.01$. All experiments are conducted on a single NVIDIA GeForce RTX 5090 GPU.
The remaining dataset-specific training configurations are provided in
Table~\ref{tab:ar_configuration}.

\textbf{Inference.}
For whole-volume synthesis, we use a $256\times256\times128$ inference window. This window covers the full resized BraTS volume, whereas SynthRAD 2025 and AutoPET are generated using overlapping sliding windows. We adopt KV caching to avoid recomputing previously generated context, and the source-only endpoint required by {\spira} is also cached and reused during generation. Generation is deterministic, with each FSQ factor decoded by greedy argmax without temperature scaling or top-$k$/top-$p$ sampling. Each predicted block is appended to the target prefix before generating the next block, following the same raster order used during training.

For all retraction policies, the source-only and full-prefix endpoints are evaluated at the same current generated state before applying the corresponding retraction. For the GT-assisted oracle analyses in Sec.~\ref{sec:exp_mechanism}, candidate retraction amounts are evaluated while keeping the source, generated prefix, and factor-conditioning history fixed; the GT token is used only to score candidate actions and is never appended to the prefix. The completed latent volume is finally decoded by the frozen VidTok decoder to obtain the synthesized 3D volume.

\begin{table}[t]
\centering
\caption{\textbf{Model, training, and inference configurations} for the datasets used in our experiments.}
\setlength{\tabcolsep}{2.5mm}
\renewcommand{\arraystretch}{1.05}
\resizebox{0.90\linewidth}{!}{
\begin{tabular}{lccc}
\toprule
\textbf{Setting} & \textbf{BraTS 2024} & \textbf{SynthRAD 2025} & \textbf{AutoPET} \\
\midrule
\rowcolor{tablegray!80}
\multicolumn{4}{c}{\textit{Model Configurations}} \\
Layers         & 12 & 6 & 8 \\
Hidden dim       & 512 & 384 & 384 \\
Attention heads       & 8 & 8 & 8 \\
Parameters   & 77.86M & 27.94M & 32.66M \\

\midrule
\rowcolor{tablegray!80}
\multicolumn{4}{c}{\textit{Training Settings}} \\
Optimizer  & AdamW & AdamW & AdamW  \\
Learning rate  & $1\mathrm{e}{-4}$ & $1\mathrm{e}{-4}$ & $1\mathrm{e}{-4}$  \\
Weight decay  & 0.05 & 0.05 & 0.05  \\
$(\beta_1,\beta_2)$  & $(0.9, 0.95)$ & $(0.9, 0.95)$ & $(0.9, 0.95)$  \\
Learning rate schedule  & Warmup cosine & Warmup cosine & Warmup cosine  \\
Batch size  & 2 & 4 & 3  \\
Training steps  & 100k & 100k & 100k \\
Gradient clipping  & 2.0 & 2.0 & 2.0  \\
Precision  & bfloat16  & bfloat16 & bfloat16  \\
Training crop size    & $256\times256\times128$ & $256\times256\times128$ & $256\times256\times128$  \\
Latent window size  & $32\times32\times32$ & $32\times32\times32$ & $32\times32\times32$  \\
Block size  & $4\times4\times4$ & $4\times4\times4$ & $4\times4\times4$  \\
\#AR blocks / window  & 512 & 512 & 512  \\

\midrule
\rowcolor{tablegray!80}
\multicolumn{4}{c}{\textit{Inference Settings}} \\
Inference protocol  & Full volume & Sliding window inference & Sliding window inference  \\
Inference window size  & $256\times256\times128$ & $256\times256\times128$ & $256\times256\times128$  \\
Inference overlap  & -- & 0.25 & 0.25  \\

\bottomrule
\end{tabular}
}
\label{tab:ar_configuration}
\end{table}

\section{Motivation Analysis Details}
\label{app:motivation}

We provide additional details for the three analyses in
Fig.~\ref{fig:motivation}. All analyses use the BraTS 2024
T1n$\rightarrow$T1c held-out set and the same frozen AR predictor. No
{\spira} gate or prefix retraction is applied.

\textbf{Source-only and Full-prefix Predictions.}
For subject $n$ and target unit $i$, let $S_n$ denote the source volume
and $P_{n,<i}$ the preceding GT target blocks. We construct
\begin{equation}
    \mathbf{z}^{S}_{n,i}
    =
    f_{\theta}(S_n,\varnothing,i),
    \qquad
    \mathbf{z}^{F}_{n,i}
    =
    f_{\theta}(S_n,P_{n,<i},i),
\end{equation}
where $\mathbf{z}^{S}_{n,i}$ and $\mathbf{z}^{F}_{n,i}$ denote the
source-only and full-prefix logits. The two paths use the same predictor
and differ only in whether the preceding target prefix is visible.

The full-prefix path uses the GT prefix under teacher forcing to isolate
prefix contributions from accumulated generation errors. We exclude
padding, the first target unit with an empty prefix, and positions outside
the valid nonzero brain support. For each FSQ token, probabilities are
computed separately over its $d$ categorical factors and then averaged
equally across factors.

\textbf{Average Prefix Benefit.}
Let $p^{X}_{n,i,j}$ be the predicted distribution for factor $j$ under
path $X\in\{S,F\}$, and let $y_{n,i,j}$ be its GT category.
The subject-level prediction error is
\begin{equation}
    \operatorname{Err}^{X}_{n}
    =
    \frac{1}{|\mathcal{V}_n|}
    \sum_{i\in\mathcal{V}_n}
    \left[
        -\frac{1}{d}
        \sum_{j=1}^{d}
        \log p^{X}_{n,i,j}(y_{n,i,j})
    \right],
\end{equation}
where $\mathcal{V}_n$ contains the valid target positions.

For each subject, we compute the relative error reduction as
$100(\operatorname{Err}^{S}_{n}-\operatorname{Err}^{F}_{n})/
\operatorname{Err}^{S}_{n}$. The reported average error reduction is
the equal-weight mean across subjects. The percentage of improved
subjects is the fraction satisfying
$\operatorname{Err}^{F}_{n}<\operatorname{Err}^{S}_{n}$.

\textbf{Regional Benefit Heterogeneity.}
We decode the source-only and full-prefix predictions into volumes
$\widehat{Y}^{S}_{n}$ and $\widehat{Y}^{F}_{n}$ using the frozen VidTok
decoder. Each volume is divided into non-overlapping 3D regions matching
the spatial footprint of one AR block, and regions with at least $50\%$
valid brain voxels are retained.

For region $r$, we define the local error reduction and positive benefit as
\begin{equation}
    d_{n,r}
    =
    \operatorname{MAE}^{S}_{n,r}
    -
    \operatorname{MAE}^{F}_{n,r},
    \qquad
    b_{n,r}
    =
    n_{n,r}\max(d_{n,r},0),
\end{equation}
where $n_{n,r}$ is the number of valid voxels. Positive $d_{n,r}$ indicates that the target prefix reduces local reconstruction error. After ranking regions by $d_{n,r}$, let $\mathcal{T}_n(q)$ denote the top fraction $q$. Their cumulative share of positive benefit is
\begin{equation}
    C_n(q)
    =
    \frac{
        \sum_{r\in\mathcal{T}_n(q)} b_{n,r}
    }{
        \sum_r b_{n,r}
    }.
\end{equation}
We report $C_n(0.20)$ and $q_{50}$, the smallest region fraction capturing $50\%$ of the positive benefit, and summarize both by the subject-level median. The heatmap visualizes the positive voxel-wise reduction in absolute reconstruction error from source-only to full-prefix prediction.

\textbf{Matched Prediction Changes and Target Outcomes.}
For each valid latent token, we measure the prefix-induced prediction
change and its effect on the GT target as
\begin{align}
    \Delta^{\mathrm{pred}}_{n,i}
    &=
    \frac{1}{d}
    \sum_{j=1}^{d}
    \frac{1}{2}
    \sum_{c}
    \left|
        p^{F}_{n,i,j}(c)
        -
        p^{S}_{n,i,j}(c)
    \right|,
    \\
    \Delta^{\mathrm{GT}}_{n,i}
    &=
    \frac{1}{d}
    \sum_{j=1}^{d}
    \left[
        \log p^{F}_{n,i,j}(y_{n,i,j})
        -
        \log p^{S}_{n,i,j}(y_{n,i,j})
    \right].
\end{align}
Here, $\Delta^{\mathrm{pred}}$ is the factor-averaged total variation
distance, while $\Delta^{\mathrm{GT}}$ measures the change in GT
log-likelihood, with positive values indicating a more likely target.

The matched-change interval is defined using $\Delta^{\mathrm{pred}}$
only. After removing the lowest $20\%$ of near-zero changes, we use the same maximum number of tokens per subject and select the 45th--55th
percentiles, yielding the fixed interval $[0.085,0.096]$.

Using $\tau=0.05$, matched tokens are categorized as \emph{target more
likely} if $\Delta^{\mathrm{GT}}>\tau$, \emph{little change} if
$|\Delta^{\mathrm{GT}}|\leq\tau$, and \emph{target less likely} if
$\Delta^{\mathrm{GT}}<-\tau$. Proportions are computed per subject and
then averaged equally across subjects.

For the coexistence analysis, subjects with at least 100 matched tokens
are included. A subject is counted as containing both improving and
worsening outcomes when each occupies at least $5\%$ of its matched
tokens, based on the sign of $\Delta^{\mathrm{GT}}$.

\section{Additional Experimental Results}
\label{app:exp}

\subsection{Tumor- and Lesion-ROI Reconstruction Evaluation}
\label{app:exp_roi}

\begin{table}[t]
\centering
\setlength{\tabcolsep}{1.5mm}
\renewcommand\arraystretch{1.0}
\caption{\textbf{Tumor-ROI reconstruction evaluation on BraTS 2024.}
We report PSNR and SSIM within annotation-defined tumor regions across four translation tasks.}
\resizebox{\linewidth}{!}{
\begin{tabular}{lcc|cc|cc|cc}
\toprule
\multirow{2}{*}{\textbf{Methods}} 
& \multicolumn{2}{c|}{\textbf{T1n$\rightarrow$T1c}} 
& \multicolumn{2}{c|}{\textbf{T1c$\rightarrow$T1n}} 
& \multicolumn{2}{c|}{\textbf{T2w$\rightarrow$T2f}} 
& \multicolumn{2}{c}{\textbf{T2f$\rightarrow$T1c}} \\

\cmidrule(lr){2-3}
\cmidrule(lr){4-5}
\cmidrule(lr){6-7}
\cmidrule(lr){8-9}

& PSNR$\uparrow$ 
& SSIM$\uparrow$ 
& PSNR$\uparrow$ 
& SSIM$\uparrow$ 
& PSNR$\uparrow$ 
& SSIM$\uparrow$ 
& PSNR$\uparrow$ 
& SSIM$\uparrow$ \\

\midrule
\rowcolor{tablegray!50}
\multicolumn{9}{c}{\textit{GAN-based Methods}} \\

3D-mADUNet~\citep{10944137}
& \roicellgray{19.47}{3.05}
& \roicellgray{0.678}{0.12}
& \roicellgray{20.95}{3.35}
& \roicellgray{0.794}{0.11}
& \roicellgray{17.75}{2.96}
& \roicellgray{0.641}{0.10}
& \roicellgray{17.97}{3.44}
& \roicellgray{0.651}{0.09} \\

MMHVAE~\citep{11197215}
& \roicellgray{19.73}{2.76}
& \roicellgray{0.685}{0.09}
& \roicellgray{21.25}{2.94}
& \roicellgray{0.791}{0.12}
& \roicellgray{17.69}{3.28}
& \roicellgray{0.641}{0.13}
& \roicellgray{18.00}{2.79}
& \roicellgray{0.662}{0.10} \\

\midrule
\rowcolor{tablegray!50}
\multicolumn{9}{c}{\textit{Diffusion-based Methods}} \\

M2DN~\citep{10444695}
& \roicellgray{19.59}{3.12}
& \roicellgray{0.682}{0.13}
& \roicellgray{21.01}{2.74}
& \roicellgray{0.792}{0.10}
& \roicellgray{17.88}{3.51}
& \roicellgray{0.645}{0.07}
& \roicellgray{17.92}{3.22}
& \roicellgray{0.649}{0.08} \\

selfRDB~\citep{ARSLAN2025103747}
& \roicellgray{19.86}{3.23}
& \roicellgray{0.692}{0.10}
& \roicellgray{21.12}{3.11}
& \roicellgray{0.795}{0.07}
& \roicellgray{18.15}{2.98}
& \roicellgray{0.662}{0.09}
& \roicellgray{18.40}{2.70}
& \roicellgray{0.670}{0.11} \\

CRSG~\citep{11559110}
& \roicellgray{20.07}{2.89}
& \roicellgray{0.692}{0.08}
& \roicellgray{21.57}{2.69}
& \roicellgray{0.801}{0.09}
& \roicellgray{18.03}{2.72}
& \roicellgray{0.665}{0.09}
& \roicellgray{18.24}{2.85}
& \roicellgray{0.667}{0.11} \\

\midrule

\rowcolor{tablegreen!60}
Source-only
& \roicellgreen{19.79}{3.20}
& \roicellgreen{0.687}{0.12}
& \roicellgreen{21.38}{3.29}
& \roicellgreen{0.799}{0.10}
& \roicellgreen{18.11}{3.14}
& \roicellgreen{0.664}{0.09}
& \roicellgreen{18.03}{3.04}
& \roicellgreen{0.668}{0.12} \\

\rowcolor{tableblue!60}
Full-prefix AR
& \roicellblue{20.03}{3.40}
& \roicellblue{0.690}{0.09}
& \roicellblue{21.50}{3.07}
& \roicellblue{0.800}{0.11}
& \textbf{\roicellblue{18.29}{2.84}}
& \textbf{\roicellblue{0.669}{0.10}}
& \roicellblue{18.21}{2.90}
& \roicellblue{0.674}{0.09} \\

\rowcolor{tablepurple!70}
\textbf{\spira}
& \textbf{\roicellpurple{20.12}{3.06}}
& \textbf{\roicellpurple{0.696}{0.10}}
& \textbf{\roicellpurple{21.71}{2.82}}
& \textbf{\roicellpurple{0.803}{0.09}}
& \roicellpurple{18.25}{3.36}
& \roicellpurple{0.667}{0.12}
& \textbf{\roicellpurple{18.64}{2.75}}
& \textbf{\roicellpurple{0.679}{0.09}} \\

\bottomrule
\end{tabular}
}
\label{tab:brats24_roi}
\end{table}

\begin{table}[t]
\centering
\setlength{\tabcolsep}{3.0mm}
\renewcommand\arraystretch{1.10}
\caption{\textbf{Lesion-ROI reconstruction evaluation on AutoPET.}
We report PSNR and SSIM within annotation-defined tumor-lesion regions.}
\resizebox{0.80\linewidth}{!}{
\begin{tabular}{lcc|cc}
\toprule

\multirow{2}{*}{\textbf{Methods}} 
& \multicolumn{2}{c|}{\textbf{CT$\rightarrow$PET}} 
& \multicolumn{2}{c}{\textbf{PET$\rightarrow$CT}} \\

\cmidrule(lr){2-3}
\cmidrule(lr){4-5}

& PSNR$\uparrow$ 
& SSIM$\uparrow$ 
& PSNR$\uparrow$ 
& SSIM$\uparrow$ \\

\midrule
\rowcolor{tablegray!50}
\multicolumn{5}{c}{\textit{GAN-based Methods}} \\

3D-mADUNet~\citep{10944137}
& \roicellgray{16.76}{1.39}
& \roicellgray{0.887}{0.04}
& \roicellgray{17.64}{1.19}
& \roicellgray{0.779}{0.03} \\

MMHVAE~\citep{11197215}
& \roicellgray{17.08}{1.21}
& \roicellgray{0.902}{0.02}
& \roicellgray{17.80}{1.33}
& \roicellgray{0.784}{0.02} \\

\midrule
\rowcolor{tablegray!50}
\multicolumn{5}{c}{\textit{Diffusion-based Methods}} \\

M2DN~\citep{10444695}
& \roicellgray{17.53}{1.52}
& \roicellgray{0.905}{0.04}
& \roicellgray{17.68}{1.40}
& \roicellgray{0.785}{0.03} \\

selfRDB~\citep{ARSLAN2025103747}
& \roicellgray{17.29}{1.36}
& \roicellgray{0.907}{0.03}
& \roicellgray{17.52}{1.60}
& \roicellgray{0.782}{0.04} \\

CRSG~\citep{11559110}
& \roicellgray{17.51}{1.34}
& \roicellgray{0.913}{0.02}
& \roicellgray{17.79}{1.27}
& \roicellgray{0.788}{0.02} \\

\midrule

\rowcolor{tablegreen!60}
Source-only
& \roicellgreen{17.26}{1.50}
& \roicellgreen{0.910}{0.02}
& \roicellgreen{17.94}{1.44}
& \roicellgreen{0.791}{0.03} \\

\rowcolor{tableblue!60}
Full-prefix AR
& \roicellblue{17.42}{1.45}
& \roicellblue{0.910}{0.03}
& \roicellblue{17.85}{1.52}
& \roicellblue{0.788}{0.03} \\

\rowcolor{tablepurple!70}
\textbf{\spira}
& \textbf{\roicellpurple{17.79}{1.28}}
& \textbf{\roicellpurple{0.917}{0.02}}
& \textbf{\roicellpurple{18.22}{1.30}}
& \textbf{\roicellpurple{0.792}{0.04}} \\

\bottomrule
\end{tabular}
}
\label{tab:synthrad_autopet_roi}
\end{table}

We further evaluate reconstruction quality within annotation-defined
tumor or lesion ROIs on BraTS 2024~\citep{de20242024} and AutoPET~\citep{dexl2026autopet3}, using the provided annotations only for evaluation.
As reported in Tables~\ref{tab:brats24_roi} and
\ref{tab:synthrad_autopet_roi}, {\spira} improves over Full-prefix AR
on five of six task-level ROI comparisons and achieves the best results
on all AutoPET ROI metrics; the small T2w$\rightarrow$T2f difference
is within the reported subject-level variability.
The smaller ROI gains are consistent with the higher full-retention
share in tumor regions (Fig.~\ref{fig:app_utility_effect}(c)), where less prefix correction can be safely retracted, in line with the actionability analysis in Sec.~\ref{sec:exp_mechanism}.

\subsection{Statistical Significance of Main Results}
\label{app:exp_significance}

\begin{table}[t]
\centering
\setlength{\tabcolsep}{3.0mm}
\renewcommand\arraystretch{1.10}
\caption{\textbf{Subject-level statistical analysis on BraTS 2024.}
We report the mean performance with subject-level standard deviations for four translation tasks.
Asterisks indicate statistically significant differences between each competing method and {\spira} based on two-sided paired Wilcoxon signed-rank tests: $^{*}p<0.05$, $^{**}p<0.01$, and $^{***}p<0.001$. Subscripts denote subject-level standard deviations.}
\resizebox{1.00\linewidth}{!}{
\begin{tabular}{lccc|ccc}
\toprule
\multirow{2}{*}{\textbf{Methods}}
& \multicolumn{3}{c|}{\textbf{T1n$\rightarrow$T1c}}
& \multicolumn{3}{c}{\textbf{T1c$\rightarrow$T1n}} \\
\cmidrule(lr){2-4}
\cmidrule(lr){5-7}
& PSNR$\uparrow$
& SSIM$\uparrow$
& LPIPS$\downarrow$
& PSNR$\uparrow$
& SSIM$\uparrow$
& LPIPS$\downarrow$ \\

\midrule
\rowcolor{tablegray!50}
\multicolumn{7}{c}{\textit{GAN-based Methods}} \\

3D-mADUNet~\citep{10944137}
& \statcellgray{21.30}{***}{2.09}
& \statcellgray{0.680}{***}{0.04}
& \statcellgray{0.162}{***}{0.01}
& \statcellgray{24.68}{***}{1.93}
& \statcellgray{0.822}{***}{0.04}
& \statcellgray{0.096}{***}{0.02} \\

MMHVAE~\citep{11197215}
& \statcellgray{21.66}{***}{2.40}
& \statcellgray{0.691}{***}{0.05}
& \statcellgray{0.157}{***}{0.02}
& \statcellgray{24.23}{***}{2.32}
& \statcellgray{0.819}{**}{0.04}
& \statcellgray{0.097}{***}{0.02} \\

\midrule
\rowcolor{tablegray!50}
\multicolumn{7}{c}{\textit{Diffusion-based Methods}} \\

M2DN~\citep{10444695}
& \statcellgray{21.15}{***}{2.27}
& \statcellgray{0.673}{***}{0.04}
& \statcellgray{0.161}{***}{0.02}
& \statcellgray{24.35}{***}{2.02}
& \statcellgray{0.816}{***}{0.03}
& \statcellgray{0.099}{***}{0.01} \\

selfRDB~\citep{ARSLAN2025103747}
& \statcellgray{22.03}{***}{2.30}
& \statcellgray{0.695}{***}{0.06}
& \statcellgray{0.139}{***}{0.02}
& \statcellgray{25.16}{***}{1.98}
& \statcellgray{0.834}{***}{0.05}
& \statcellgray{0.082}{***}{0.03} \\

CRSG~\citep{11559110}
& \statcellgray{21.85}{***}{2.55}
& \statcellgray{0.691}{***}{0.05}
& \statcellgray{0.148}{***}{0.03}
& \statcellgray{25.31}{**}{2.16}
& \statcellgray{0.834}{**}{0.04}
& \statcellgray{0.085}{**}{0.02} \\

\midrule

\rowcolor{tablegreen!60}
Source-only
& \statcellgreen{21.59}{**}{2.14}
& \statcellgreen{0.690}{*}{0.05}
& \statcellgreen{0.151}{**}{0.02}
& \statcellgreen{24.96}{***}{2.33}
& \statcellgreen{0.828}{**}{0.04}
& \statcellgreen{0.091}{***}{0.02} \\

\rowcolor{tableblue!60}
Full-prefix AR
& \statcellblue{22.10}{**}{2.12}
& \statcellblue{0.696}{**}{0.06}
& \statcellblue{0.135}{***}{0.04}
& \statcellblue{25.49}{***}{2.19}
& \statcellblue{0.834}{*}{0.04}
& \statcellblue{0.085}{**}{0.02} \\

\rowcolor{tablepurple!70}
\textbf{\spira}
& \textbf{\ourcellpurple{22.72}{2.43}}
& \textbf{\ourcellpurple{0.705}{0.06}}
& \textbf{\ourcellpurple{0.126}{0.04}}
& \textbf{\ourcellpurple{25.93}{2.07}}
& \textbf{\ourcellpurple{0.842}{0.05}}
& \textbf{\ourcellpurple{0.077}{0.03}} \\

\midrule
\midrule

\multirow{2}{*}{\textbf{Methods}}
& \multicolumn{3}{c|}{\textbf{T2w$\rightarrow$T2f}}
& \multicolumn{3}{c}{\textbf{T2f$\rightarrow$T1c}} \\
\cmidrule(lr){2-4}
\cmidrule(lr){5-7}
& PSNR$\uparrow$
& SSIM$\uparrow$
& LPIPS$\downarrow$
& PSNR$\uparrow$
& SSIM$\uparrow$
& LPIPS$\downarrow$ \\

\midrule
\rowcolor{tablegray!50}
\multicolumn{7}{c}{\textit{GAN-based Methods}} \\

3D-mADUNet~\citep{10944137}
& \statcellgray{20.84}{***}{1.91}
& \statcellgray{0.646}{***}{0.06}
& \statcellgray{0.177}{***}{0.03}
& \statcellgray{20.71}{***}{1.80}
& \statcellgray{0.653}{***}{0.04}
& \statcellgray{0.178}{***}{0.04} \\

MMHVAE~\citep{11197215}
& \statcellgray{20.67}{***}{2.20}
& \statcellgray{0.643}{***}{0.05}
& \statcellgray{0.182}{**}{0.04}
& \statcellgray{20.90}{***}{2.11}
& \statcellgray{0.662}{***}{0.05}
& \statcellgray{0.170}{***}{0.02} \\

\midrule
\rowcolor{tablegray!50}
\multicolumn{7}{c}{\textit{Diffusion-based Methods}} \\

M2DN~\citep{10444695}
& \statcellgray{20.83}{**}{2.34}
& \statcellgray{0.648}{***}{0.07}
& \statcellgray{0.179}{***}{0.05}
& \statcellgray{20.64}{***}{1.97}
& \statcellgray{0.651}{***}{0.06}
& \statcellgray{0.184}{***}{0.04} \\

selfRDB~\citep{ARSLAN2025103747}
& \statcellgray{21.42}{***}{1.77}
& \statcellgray{0.674}{***}{0.04}
& \statcellgray{0.159}{***}{0.02}
& \statcellgray{21.33}{**}{1.93}
& \statcellgray{0.677}{***}{0.05}
& \statcellgray{0.160}{**}{0.03} \\

CRSG~\citep{11559110}
& \statcellgray{21.19}{***}{2.13}
& \statcellgray{0.670}{**}{0.06}
& \statcellgray{0.166}{***}{0.03}
& \statcellgray{21.02}{***}{2.27}
& \statcellgray{0.669}{***}{0.08}
& \statcellgray{0.171}{***}{0.03} \\

\midrule

\rowcolor{tablegreen!60}
Source-only
& \statcellgreen{21.01}{**}{1.84}
& \statcellgreen{0.670}{***}{0.07}
& \statcellgreen{0.164}{***}{0.03}
& \statcellgreen{21.09}{**}{1.99}
& \statcellgreen{0.671}{***}{0.06}
& \statcellgreen{0.163}{*}{0.04} \\

\rowcolor{tableblue!60}
Full-prefix AR
& \statcellblue{21.30}{**}{2.21}
& \statcellblue{0.674}{**}{0.06}
& \statcellblue{0.155}{**}{0.04}
& \statcellblue{21.55}{***}{2.05}
& \statcellblue{0.681}{***}{0.06}
& \statcellblue{0.154}{**}{0.03} \\

\rowcolor{tablepurple!70}
\textbf{\spira}
& \textbf{\ourcellpurple{21.77}{2.09}}
& \textbf{\ourcellpurple{0.680}{0.07}}
& \textbf{\ourcellpurple{0.150}{0.04}}
& \textbf{\ourcellpurple{22.08}{1.92}}
& \textbf{\ourcellpurple{0.687}{0.05}}
& \textbf{\ourcellpurple{0.149}{0.03}} \\

\bottomrule
\end{tabular}
}
\vspace{-8pt}
\label{tab:brats24_sig}
\end{table}

\begin{table}[t]
\centering
\setlength{\tabcolsep}{3.0mm}
\renewcommand\arraystretch{1.00}
\caption{\textbf{Subject-level statistical analysis on SynthRAD 2025 and AutoPET.} We report the mean performance with subject-level standard deviations for four translation tasks.
Asterisks indicate statistically significant differences between each competing method and {\spira} based on two-sided paired Wilcoxon signed-rank tests: $^{*}p<0.05$, $^{**}p<0.01$, and $^{***}p<0.001$. Subscripts denote subject-level standard deviations.}
\resizebox{1.00\linewidth}{!}{
\begin{tabular}{lccc|ccc}
\toprule

\multirow{2}{*}{\textbf{Methods}}
& \multicolumn{3}{c|}{\textbf{MRI$\rightarrow$CT}}
& \multicolumn{3}{c}{\textbf{CBCT$\rightarrow$CT}} \\

\cmidrule(lr){2-4}
\cmidrule(lr){5-7}

& PSNR$\uparrow$
& SSIM$\uparrow$
& LPIPS$\downarrow$
& PSNR$\uparrow$
& SSIM$\uparrow$
& LPIPS$\downarrow$ \\

\midrule
\rowcolor{tablegray!50}
\multicolumn{7}{c}{\textit{GAN-based Methods}} \\

3D-mADUNet~\citep{10944137}
& \statcellgray{20.78}{***}{1.85}
& \statcellgray{0.701}{***}{0.05}
& \statcellgray{0.238}{***}{0.03}
& \statcellgray{22.94}{***}{1.81}
& \statcellgray{0.737}{***}{0.07}
& \statcellgray{0.128}{***}{0.04} \\

MMHVAE~\citep{11197215}
& \statcellgray{20.49}{***}{2.11}
& \statcellgray{0.697}{***}{0.05}
& \statcellgray{0.249}{***}{0.03}
& \statcellgray{22.73}{***}{2.08}
& \statcellgray{0.736}{***}{0.06}
& \statcellgray{0.131}{***}{0.03} \\

\midrule
\rowcolor{tablegray!50}
\multicolumn{7}{c}{\textit{Diffusion-based Methods}} \\

M2DN~\citep{10444695}
& \statcellgray{20.55}{***}{2.34}
& \statcellgray{0.701}{***}{0.05}
& \statcellgray{0.242}{***}{0.04}
& \statcellgray{22.66}{***}{1.94}
& \statcellgray{0.734}{***}{0.07}
& \statcellgray{0.133}{***}{0.04} \\

selfRDB~\citep{ARSLAN2025103747}
& \statcellgray{20.81}{***}{1.95}
& \statcellgray{0.708}{***}{0.08}
& \statcellgray{0.236}{***}{0.05}
& \statcellgray{23.19}{**}{1.98}
& \statcellgray{0.740}{**}{0.08}
& \statcellgray{0.127}{***}{0.05} \\

CRSG~\citep{11559110}
& \statcellgray{20.62}{***}{2.16}
& \statcellgray{0.694}{***}{0.07}
& \statcellgray{0.248}{***}{0.05}
& \statcellgray{22.78}{***}{1.74}
& \statcellgray{0.732}{***}{0.05}
& \statcellgray{0.136}{***}{0.03} \\

\midrule

\rowcolor{tablegreen!60}
Source-only
& \statcellgreen{20.27}{***}{2.20}
& \statcellgreen{0.691}{**}{0.05}
& \statcellgreen{0.252}{**}{0.03}
& \statcellgreen{23.02}{***}{2.04}
& \statcellgreen{0.739}{**}{0.06}
& \statcellgreen{0.129}{***}{0.05} \\

\rowcolor{tableblue!60}
Full-prefix AR
& \statcellblue{20.73}{***}{1.89}
& \statcellblue{0.705}{***}{0.06}
& \statcellblue{0.232}{**}{0.03}
& \statcellblue{23.35}{*}{1.91}
& \statcellblue{0.744}{**}{0.06}
& \statcellblue{0.123}{**}{0.04} \\

\rowcolor{tablepurple!70}
\textbf{\spira}
& \textbf{\ourcellpurple{21.29}{2.07}}
& \textbf{\ourcellpurple{0.714}{0.07}}
& \textbf{\ourcellpurple{0.225}{0.05}}
& \textbf{\ourcellpurple{23.59}{1.82}}
& \textbf{\ourcellpurple{0.751}{0.06}}
& \textbf{\ourcellpurple{0.119}{0.04}} \\

\midrule
\midrule

\multirow{2}{*}{\textbf{Methods}}
& \multicolumn{3}{c|}{\textbf{CT$\rightarrow$PET}}
& \multicolumn{3}{c}{\textbf{PET$\rightarrow$CT}} \\

\cmidrule(lr){2-4}
\cmidrule(lr){5-7}

& PSNR$\uparrow$
& SSIM$\uparrow$
& LPIPS$\downarrow$
& PSNR$\uparrow$
& SSIM$\uparrow$
& LPIPS$\downarrow$ \\

\midrule
\rowcolor{tablegray!50}
\multicolumn{7}{c}{\textit{GAN-based Methods}} \\

3D-mADUNet~\citep{10944137}
& \statcellgray{24.73}{***}{1.39}
& \statcellgray{0.902}{***}{0.03}
& \statcellgray{0.130}{***}{0.02}
& \statcellgray{23.49}{***}{0.91}
& \statcellgray{0.896}{***}{0.02}
& \statcellgray{0.105}{***}{0.01} \\

MMHVAE~\citep{11197215}
& \statcellgray{25.57}{***}{1.22}
& \statcellgray{0.914}{***}{0.02}
& \statcellgray{0.106}{***}{0.01}
& \statcellgray{23.71}{***}{1.07}
& \statcellgray{0.899}{***}{0.03}
& \statcellgray{0.103}{***}{0.01} \\

\midrule
\rowcolor{tablegray!50}
\multicolumn{7}{c}{\textit{Diffusion-based Methods}} \\

M2DN~\citep{10444695}
& \statcellgray{25.69}{***}{1.47}
& \statcellgray{0.912}{***}{0.03}
& \statcellgray{0.104}{***}{0.02}
& \statcellgray{23.81}{***}{0.96}
& \statcellgray{0.902}{***}{0.03}
& \statcellgray{0.095}{**}{0.02} \\

selfRDB~\citep{ARSLAN2025103747}
& \statcellgray{26.08}{***}{1.51}
& \statcellgray{0.923}{***}{0.03}
& \statcellgray{0.094}{***}{0.02}
& \statcellgray{23.67}{***}{1.01}
& \statcellgray{0.901}{***}{0.02}
& \statcellgray{0.099}{***}{0.01} \\

CRSG~\citep{11559110}
& \statcellgray{26.85}{***}{1.27}
& \statcellgray{0.937}{***}{0.02}
& \statcellgray{0.080}{**}{0.01}
& \statcellgray{23.98}{***}{1.12}
& \statcellgray{0.903}{***}{0.02}
& \statcellgray{0.095}{***}{0.01} \\

\midrule

\rowcolor{tablegreen!60}
Source-only
& \statcellgreen{27.30}{***}{1.58}
& \statcellgreen{0.939}{***}{0.02}
& \statcellgreen{0.078}{**}{0.01}
& \statcellgreen{23.58}{***}{1.03}
& \statcellgreen{0.901}{***}{0.02}
& \statcellgreen{0.097}{***}{0.01} \\

\rowcolor{tableblue!60}
Full-prefix AR
& \statcellblue{27.93}{**}{1.35}
& \statcellblue{0.947}{**}{0.02}
& \statcellblue{0.066}{***}{0.02}
& \statcellblue{24.15}{***}{0.99}
& \statcellblue{0.905}{***}{0.03}
& \statcellblue{0.093}{**}{0.02} \\

\rowcolor{tablepurple!70}
\textbf{\spira}
& \textbf{\ourcellpurple{28.44}{1.43}}
& \textbf{\ourcellpurple{0.956}{0.02}}
& \textbf{\ourcellpurple{0.058}{0.01}}
& \textbf{\ourcellpurple{24.76}{1.07}}
& \textbf{\ourcellpurple{0.912}{0.02}}
& \textbf{\ourcellpurple{0.086}{0.02}} \\

\bottomrule
\end{tabular}
}
\label{tab:synthrad_autopet_sig}
\end{table}

Tables~\ref{tab:brats24_sig} and~\ref{tab:synthrad_autopet_sig} report
subject-level variability and paired statistical comparisons with {\spira}. Across all eight translation tasks, {\spira} consistently outperforms Full-prefix AR on PSNR, SSIM, and LPIPS.
The paired tests, together with the improvements over external baselines, support the robustness of the main quantitative results.

\subsection{Multi-seed Robustness Analysis}
\label{app:exp_seed}

Table~\ref{tab:brats24_multiseed} reports results over three independent
training runs on all four BraTS 2024~\citep{de20242024} translation tasks.
{\spira} consistently achieves favorable three-seed mean performance
across all metrics, with modest variation across seeds.
The averaged results remain close to those in the main experiments,
supporting the stability of selective source--prefix balancing under
training randomness.

\begin{table}[t]
\centering
\setlength{\tabcolsep}{3.0mm}
\renewcommand\arraystretch{1.1}
\caption{\textbf{Multi-seed robustness analysis on BraTS 2024.}
We report the mean performance over three independent training runs.
Source-only, Full-prefix AR, and {\spira} are evaluated from the same
trained dual-endpoint predictor, with subscripts denoting
standard deviations across seeds.}
\resizebox{1.00\linewidth}{!}{
\begin{tabular}{lcccc|cccc}
\toprule
\multirow{2}{*}{\textbf{Methods}}
& \multicolumn{4}{c|}{\textbf{T1n$\rightarrow$T1c}}
& \multicolumn{4}{c}{\textbf{T1c$\rightarrow$T1n}} \\
\cmidrule(lr){2-5}
\cmidrule(lr){6-9}
& PSNR$\uparrow$
& SSIM$\uparrow$
& LPIPS$\downarrow$
& FID$\downarrow$
& PSNR$\uparrow$
& SSIM$\uparrow$
& LPIPS$\downarrow$
& FID$\downarrow$ \\

\rowcolor{tablegreen!60}
Source-only
& \seedcellgreen{21.55}{0.19}
& \seedcellgreen{0.690}{0.004}
& \seedcellgreen{0.152}{0.002}
& \seedcellgreen{47.0}{1.0}
& \seedcellgreen{24.99}{0.21}
& \seedcellgreen{0.828}{0.005}
& \seedcellgreen{0.090}{0.004}
& \seedcellgreen{34.2}{0.8} \\

\rowcolor{tableblue!60}
Full-prefix AR
& \seedcellblue{22.14}{0.22}
& \seedcellblue{0.697}{0.005}
& \seedcellblue{0.133}{0.003}
& \seedcellblue{42.8}{0.9}
& \seedcellblue{25.51}{0.15}
& \seedcellblue{0.834}{0.004}
& \seedcellblue{0.085}{0.002}
& \seedcellblue{32.5}{1.1} \\

\rowcolor{tablepurple!70}
\textbf{\spira}
& \textbf{\seedcellpurple{22.74}{0.19}}
& \textbf{\seedcellpurple{0.706}{0.005}}
& \textbf{\seedcellpurple{0.126}{0.002}}
& \textbf{\seedcellpurple{38.9}{1.3}}
& \textbf{\seedcellpurple{25.90}{0.17}}
& \textbf{\seedcellpurple{0.841}{0.004}}
& \textbf{\seedcellpurple{0.079}{0.002}}
& \textbf{\seedcellpurple{29.1}{1.2}} \\

\midrule
\midrule

\multirow{2}{*}{\textbf{Methods}}
& \multicolumn{4}{c|}{\textbf{T2w$\rightarrow$T2f}}
& \multicolumn{4}{c}{\textbf{T2f$\rightarrow$T1c}} \\
\cmidrule(lr){2-5}
\cmidrule(lr){6-9}
& PSNR$\uparrow$
& SSIM$\uparrow$
& LPIPS$\downarrow$
& FID$\downarrow$
& PSNR$\uparrow$
& SSIM$\uparrow$
& LPIPS$\downarrow$
& FID$\downarrow$ \\

\rowcolor{tablegreen!60}
Source-only
& \seedcellgreen{21.05}{0.12}
& \seedcellgreen{0.671}{0.004}
& \seedcellgreen{0.162}{0.003}
& \seedcellgreen{55.8}{1.2}
& \seedcellgreen{21.03}{0.20}
& \seedcellgreen{0.670}{0.005}
& \seedcellgreen{0.163}{0.003}
& \seedcellgreen{49.2}{1.0} \\

\rowcolor{tableblue!60}
Full-prefix AR
& \seedcellblue{21.31}{0.18}
& \seedcellblue{0.674}{0.004}
& \seedcellblue{0.154}{0.004}
& \seedcellblue{49.1}{0.8}
& \seedcellblue{21.57}{0.15}
& \seedcellblue{0.680}{0.003}
& \seedcellblue{0.152}{0.002}
& \seedcellblue{46.4}{1.1} \\

\rowcolor{tablepurple!70}
\textbf{\spira}
& \textbf{\seedcellpurple{21.75}{0.14}}
& \textbf{\seedcellpurple{0.681}{0.004}}
& \textbf{\seedcellpurple{0.152}{0.003}}
& \textbf{\seedcellpurple{43.1}{0.9}}
& \textbf{\seedcellpurple{22.03}{0.17}}
& \textbf{\seedcellpurple{0.685}{0.005}}
& \textbf{\seedcellpurple{0.151}{0.004}}
& \textbf{\seedcellpurple{42.7}{1.1}} \\

\bottomrule
\end{tabular}
}
\label{tab:brats24_multiseed}
\end{table}

\subsection{Downstream Applicability to Tumor Segmentation}
\label{app:exp_downstream}

\begin{table}[t]
\centering
\setlength{\tabcolsep}{2.0mm}
\renewcommand\arraystretch{1.1}
\caption{\textbf{Downstream applicability to tumor segmentation on BraTS 2024.} Performance is measured by Dice. Asterisks indicate statistically significant differences between each competing method and {\spira} based on paired Wilcoxon signed-rank tests: $^{*}p<0.05$, $^{**}p<0.01$, and $^{***}p<0.001$. Subscripts denote subject-level standard deviations.}
\resizebox{0.85\linewidth}{!}{
\begin{tabular}{lcccc|c}
\toprule

\textbf{Methods}
& \textbf{T1n$\rightarrow$T1c}
& \textbf{T1c$\rightarrow$T1n}
& \textbf{T2w$\rightarrow$T2f}
& \textbf{T2f$\rightarrow$T1c}
& \textbf{Avg}$\uparrow$ \\

\midrule
\rowcolor{tablegray!50}
\multicolumn{6}{c}{\textit{Reference Settings}} \\

3 Real Modalities
& \statcellgray{0.753}{}{0.09}
& \statcellgray{0.782}{}{0.11}
& \statcellgray{0.718}{}{0.13}
& \statcellgray{0.753}{}{0.09}
& \statcellgray{0.752}{}{0.11} \\

\midrule
\rowcolor{tablegray!50}
\multicolumn{6}{c}{\textit{GAN-based Methods}} \\

3D-mADUNet~\citep{10944137}
& \statcellgray{0.769}{***}{0.12}
& \statcellgray{0.808}{**}{0.09}
& \statcellgray{0.740}{***}{0.13}
& \statcellgray{0.760}{**}{0.10}
& \statcellgray{0.769}{}{0.11} \\

MMHVAE~\citep{11197215}
& \statcellgray{0.791}{**}{0.11}
& \statcellgray{0.816}{**}{0.08}
& \statcellgray{0.734}{***}{0.11}
& \statcellgray{0.762}{***}{0.13}
& \statcellgray{0.776}{}{0.11} \\

\midrule
\rowcolor{tablegray!50}
\multicolumn{6}{c}{\textit{Diffusion-based Methods}} \\

M2DN~\citep{10444695}
& \statcellgray{0.776}{**}{0.08}
& \statcellgray{0.810}{***}{0.12}
& \statcellgray{0.742}{***}{0.07}
& \statcellgray{0.762}{**}{0.08}
& \statcellgray{0.773}{}{0.09} \\

selfRDB~\citep{ARSLAN2025103747}
& \statcellgray{0.788}{***}{0.10}
& \statcellgray{0.825}{***}{0.09}
& \textbf{\statcellgray{0.761}{*}{0.11}}
& \statcellgray{0.782}{**}{0.09}
& \statcellgray{0.789}{}{0.10} \\

CRSG~\citep{11559110}
& \statcellgray{0.782}{**}{0.07}
& \statcellgray{0.837}{**}{0.11}
& \statcellgray{0.745}{***}{0.08}
& \statcellgray{0.770}{***}{0.10}
& \statcellgray{0.784}{}{0.09} \\

\midrule

\rowcolor{tablegreen!60}
Source-only
& \statcellgreen{0.793}{**}{0.10}
& \statcellgreen{0.835}{***}{0.14}
& \statcellgreen{0.748}{***}{0.12}
& \statcellgreen{0.778}{**}{0.09}
& \statcellgreen{0.789}{}{0.11} \\

\rowcolor{tableblue!60}
Full-prefix AR
& \statcellblue{0.802}{***}{0.13}
& \statcellblue{0.832}{**}{0.12}
& \statcellblue{0.750}{***}{0.09}
& \statcellblue{0.780}{***}{0.14}
& \statcellblue{0.791}{}{0.12} \\

\rowcolor{tablepurple!70}
\textbf{\spira}
& \textbf{\ourcellpurple{0.809}{0.12}}
& \textbf{\ourcellpurple{0.844}{0.10}}
& \ourcellpurple{0.758}{0.14}
& \textbf{\ourcellpurple{0.791}{0.08}}
& \textbf{\ourcellpurple{0.801}{0.11}} \\

\bottomrule
\end{tabular}
}
\label{tab:segmentation}
\end{table}

\begin{figure*}[t]
\centering
\includegraphics[width=1.00\linewidth]{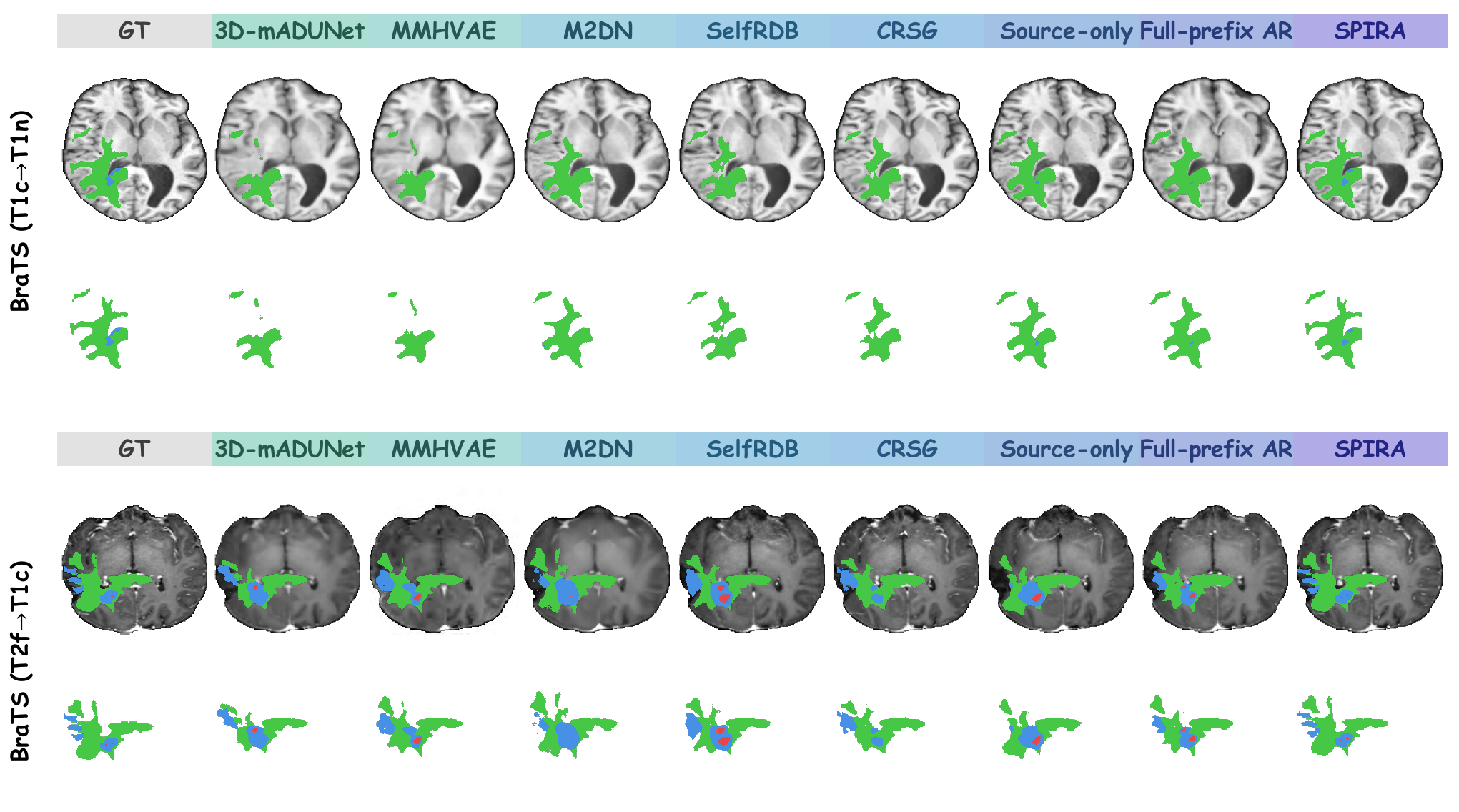}
\caption{\textbf{Qualitative comparison of downstream tumor segmentation on BraTS 2024.} Results are shown for T1c$\rightarrow$T1n (upper) and T2f$\rightarrow$T1c (lower), with segmentation overlays and the corresponding masks shown for each method.}
\label{fig:qualitative_seg}
\end{figure*}

For each BraTS 2024~\citep{de20242024} translation task, the target modality is treated as missing and synthesized by the corresponding translation model. A multimodal U-Net is trained using the three available real modalities together with the synthesized modality, and we report the subject-level mean tumor Dice. Table~\ref{tab:segmentation} shows that adding a synthesized modality consistently improves over the three-real-modality reference, while {\spira} achieves the best average Dice of 0.801 and leads on three of four tasks. Although synthesis adds no new patient information, the target-contrast representation can make existing information more useful to the segmentation network, consistent with prior missing-modality synthesis studies~\citep{10444695}. The gains over Source-only and Full-prefix AR therefore indicate improved downstream segmentation performance in this evaluated BraTS setting, rather than equivalence to an acquired target modality.
Fig.~\ref{fig:qualitative_seg} provides qualitative examples for T1c$\rightarrow$T1n and T2f$\rightarrow$T1c, which are consistent with the quantitative segmentation trends.

\subsection{Additional Qualitative Results}
\label{app:exp_vis}

\begin{figure*}[t]
\centering
\includegraphics[width=1.00\linewidth]{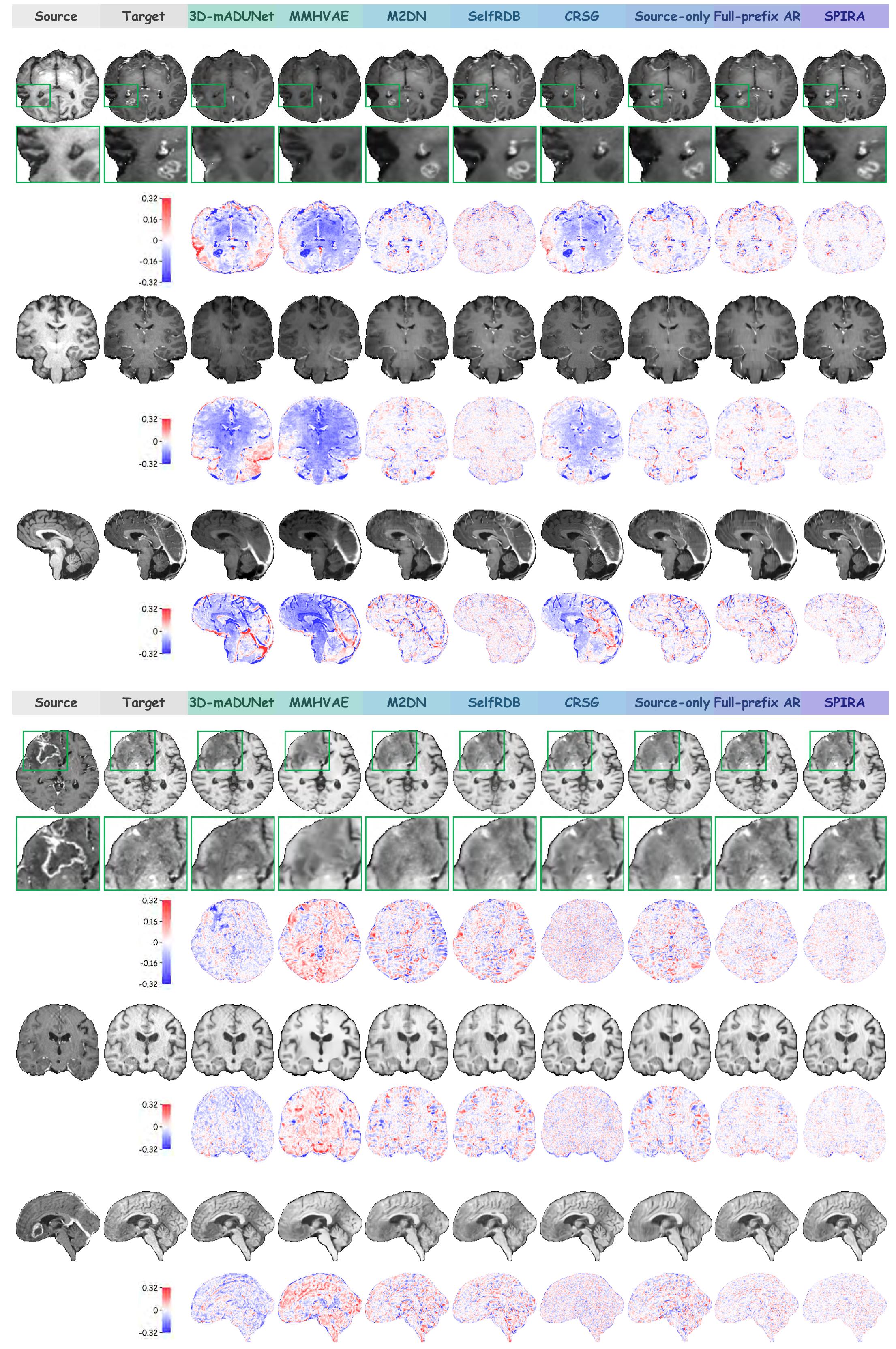}
\caption{\textbf{Additional qualitative comparisons} on BraTS 2024 T1n$\rightarrow$T1c (upper) and T1c$\rightarrow$T1n (lower), showing axial, coronal, and sagittal views from top to bottom with corresponding error maps.}
\label{fig:qualitative_brats1}
\end{figure*}

\begin{figure*}[t]
\centering
\includegraphics[width=1.00\linewidth]{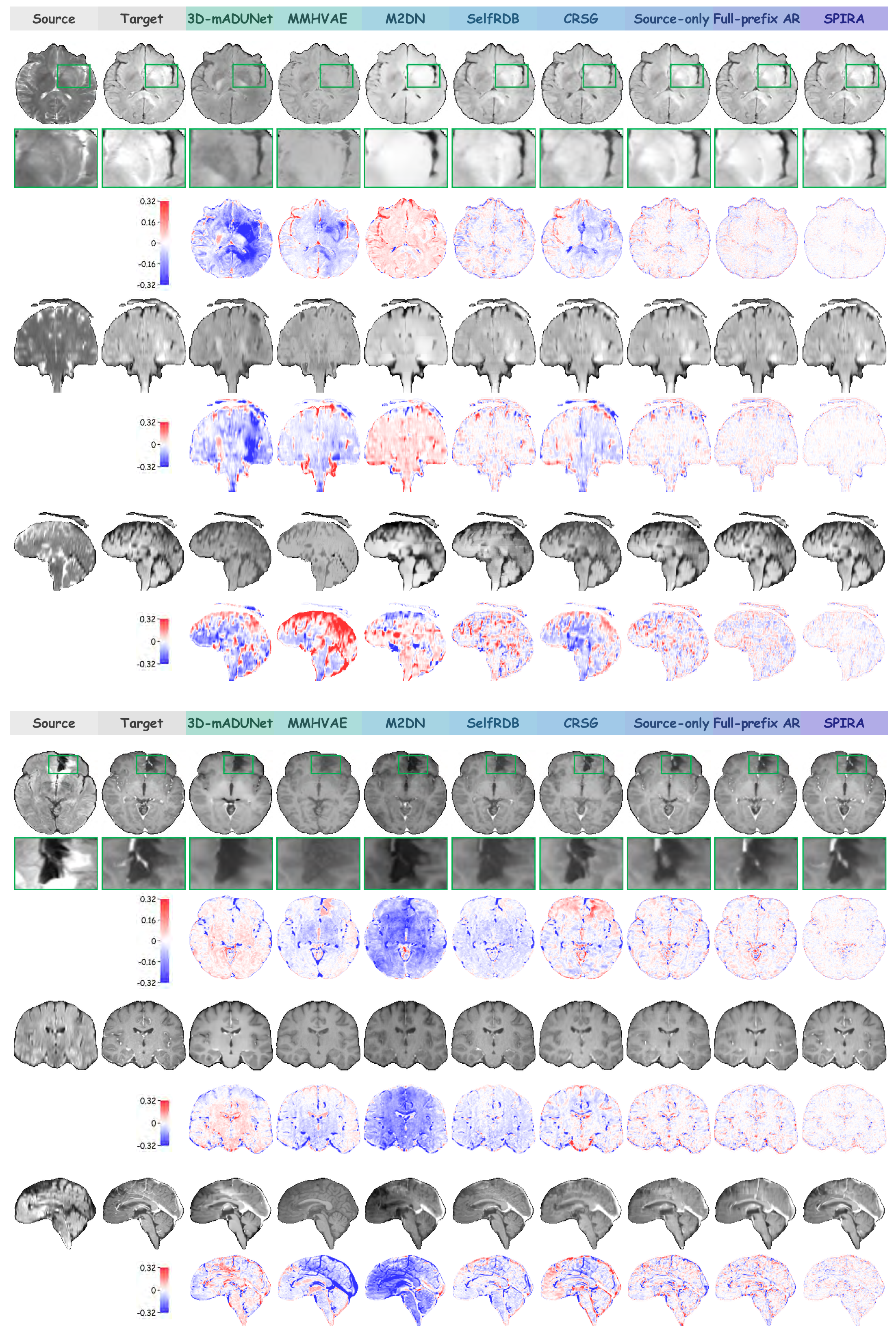}
\caption{\textbf{Additional qualitative comparisons} on BraTS 2024 T2w$\rightarrow$T2f (upper) and T2f$\rightarrow$T1c (lower), showing axial, coronal, and sagittal views from top to bottom with corresponding error maps.}
\label{fig:qualitative_brats2}
\end{figure*}

\begin{figure*}[t]
\centering
\includegraphics[width=1.00\linewidth]{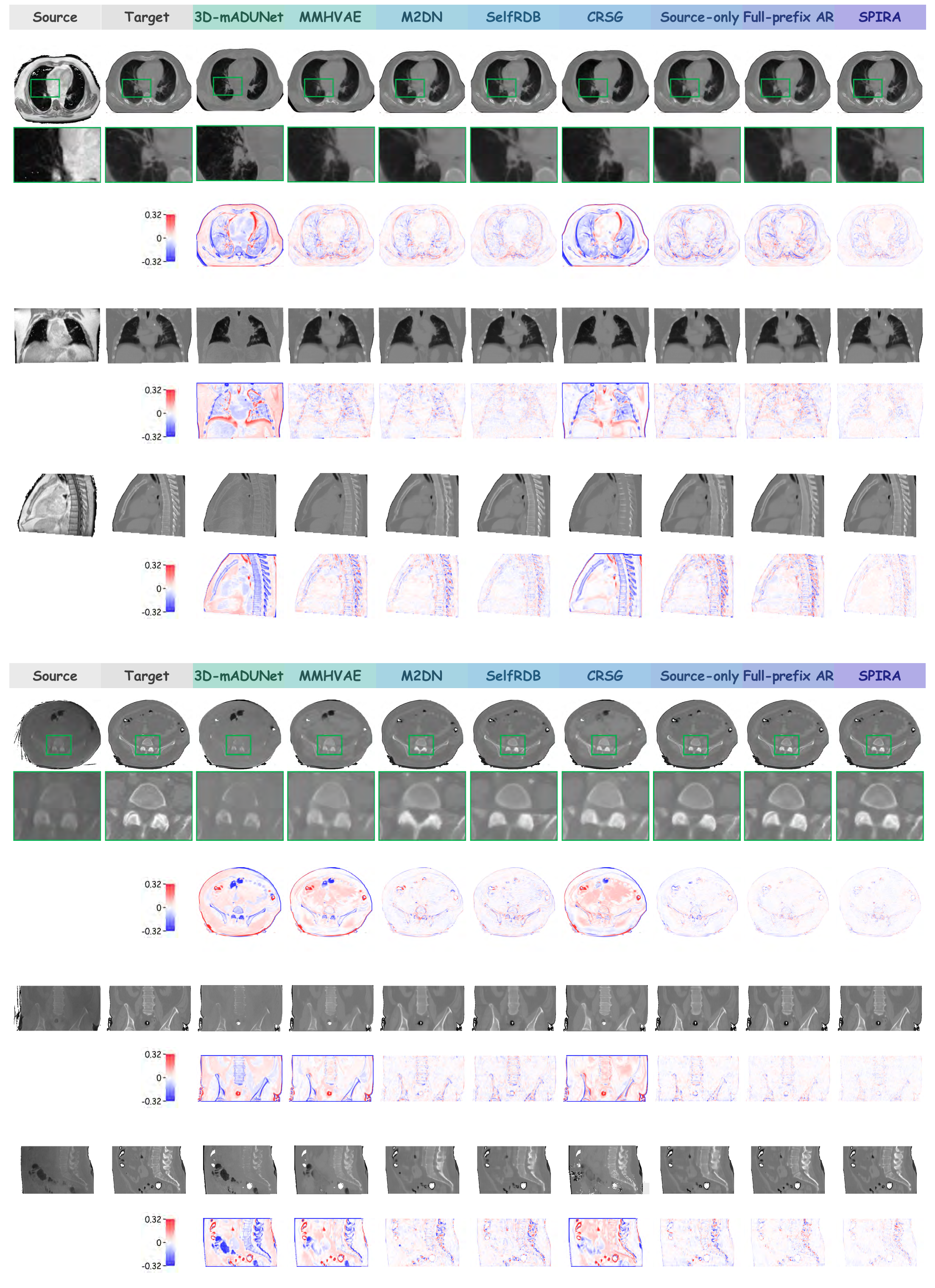}
\caption{\textbf{Additional qualitative comparisons} on SynthRAD 2025 MRI$\rightarrow$CT (upper) and CBCT$\rightarrow$CT (lower), showing axial, coronal, and sagittal views from top to bottom with corresponding error maps.}
\label{fig:qualitative_synthrad}
\end{figure*}

\begin{figure*}[t]
\centering
\includegraphics[width=1.00\linewidth]{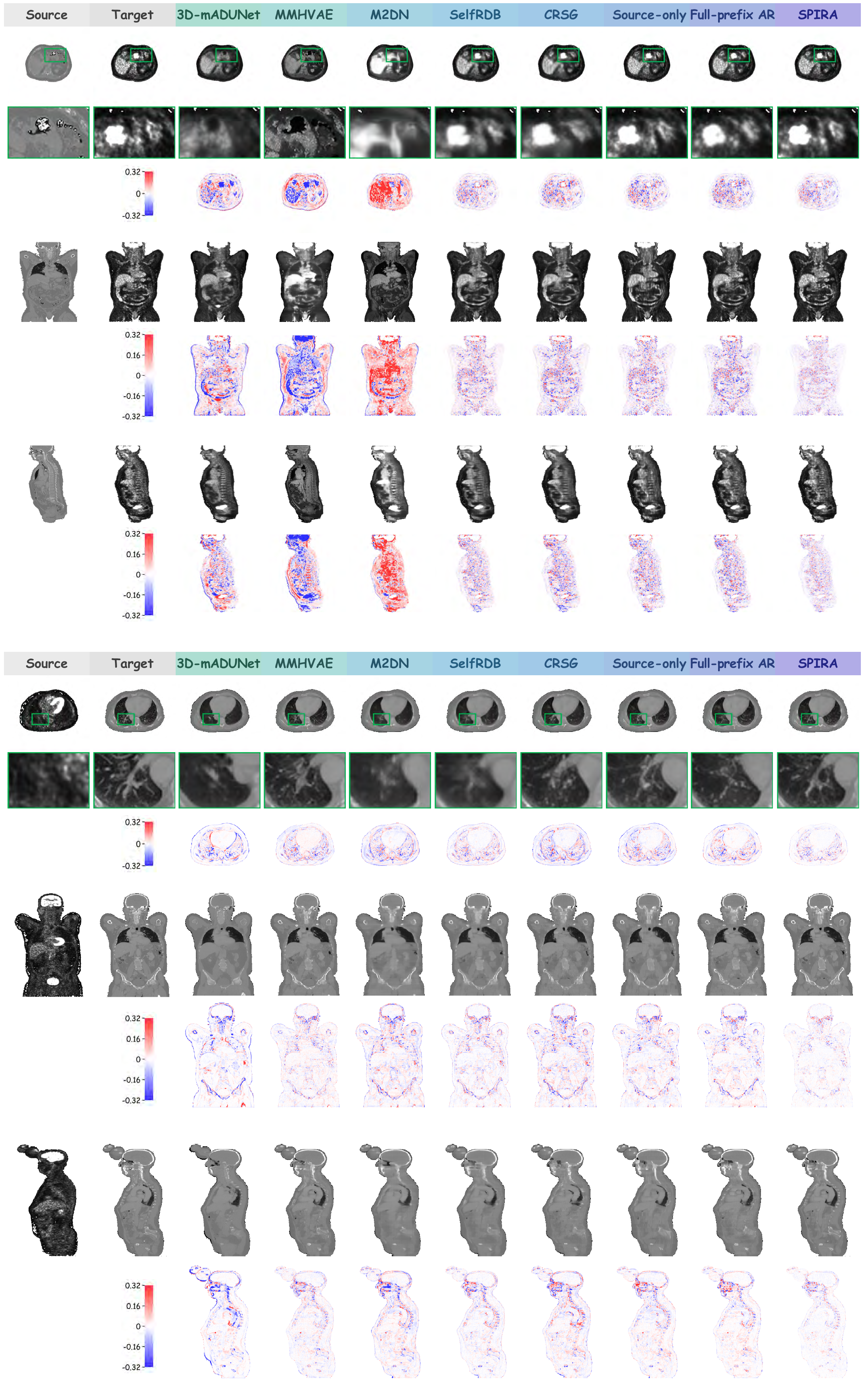}
\caption{\textbf{Additional qualitative comparisons} on AutoPET CT$\rightarrow$PET (upper) and PET$\rightarrow$CT (lower), showing axial, coronal, and sagittal views from top to bottom with corresponding error maps.}
\label{fig:qualitative_autopet}
\end{figure*}

Figs.~\ref{fig:qualitative_brats1}--\ref{fig:qualitative_autopet} provide additional illustrative qualitative comparisons
across all eight translation tasks on BraTS 2024~\citep{de20242024},
SynthRAD 2025~\citep{https://doi.org/10.1002/mp.17981}, and
AutoPET~\citep{dexl2026autopet3}.
For each task, we show axial, coronal, and sagittal views together with
the corresponding error maps. Across the selected examples, {\spira} often exhibits lower residual errors while maintaining visible target-specific structures. Together, these examples are consistent with the cross-task quantitative findings in Tables~\ref{tab:brats24}--\ref{tab:synthrad_autopet} and the ROI analyses in Tables~\ref{tab:brats24_roi}--\ref{tab:synthrad_autopet_roi}.

\subsection{More Ablation Studies}
\label{app:exp_ablation}

\textbf{Main ablation settings.}
We provide additional details for the variants evaluated in Sec.~\ref{sec:exp_ablation}.
Except for Standard AR, all variants use the same dual-endpoint AR predictor as {\spira} and modify only the retraction strategy or gate design. They therefore share the same tokenizer, AR architecture, and training configuration, allowing the comparison to isolate how prefix corrections are controlled.

\begin{itemize}
    \item \textbf{Standard AR.}
    We separately train the source-conditioned AR predictor using only \(\mathcal{L}_{\mathrm{full}}\), without the source-only endpoint, oracle, or retraction gate.
    Unlike Full-prefix AR in Tables~\ref{tab:brats24} and~\ref{tab:synthrad_autopet}, which disables retraction at \(s=0\) on the dual-endpoint predictor trained within {\spira}, this variant follows standard source-conditioned AR training.
    
    \item \textbf{Tuned global retraction.}
    We remove the adaptive gate and use a single retraction amount
    $s_i=s_{\mathrm{global}}$ for every prediction.
    The scalar is selected on the training-side model-selection subset
    from $s\in[0,1]$ and then fixed for all held-out evaluation subjects
    and spatial locations. This variant investigates whether the improvement can be obtained by globally reducing prefix influence rather than adapting the retained correction to the current region.
    
    \item \textbf{Position-only gate.}
    We retain the same oracle supervision but provide the gate only with the spatial position of the current prediction, removing all content-dependent inputs.
    This tests whether {\spira} merely learns a fixed position-dependent retraction pattern rather than adapting to the current source--prefix interaction.
    
    \item \textbf{Effect-only gate.}
    We keep the same gate architecture and oracle supervision but provide only the GT-free endpoint-difference statistics, removing the explicit source and correction representations.
    This evaluates whether predictive effect alone is sufficient to determine how much of a correction to retain.
    
    \item \textbf{Direct \(s^\ast\) regression.}
    We use the complete {\spira} gate inputs and the same maximal zero-regret oracle, but replace predictive-action distillation with direct L1 regression,
    \(\mathcal{L}_{s}=\mathbb{E}_{i\in\mathcal{V}}[|\hat{s}_i-s_i^\ast|]\).
    This comparison isolates whether matching the oracle predictive distribution is preferable to directly reproducing its scalar retraction amount.
\end{itemize}

\begin{table}[t]
\centering
\setlength{\tabcolsep}{2.0mm}
\renewcommand{\arraystretch}{1.0}

\caption{\textbf{Additional ablation studies} on BraTS 2024 T1n$\rightarrow$T1c
and SynthRAD 2025 MRI$\rightarrow$CT. We evaluate the impact of retraction strategy, oracle supervision, and gate input.}
\resizebox{1.00\linewidth}{!}{
\begin{tabular}{lcccc|cccc}
\toprule

\multirow{2}{*}{\textbf{Variants}}
& \multicolumn{4}{c|}{\textbf{BraTS T1n$\rightarrow$T1c}}
& \multicolumn{4}{c}{\textbf{SynthRAD MRI$\rightarrow$CT}} \\

\cmidrule(lr){2-5}
\cmidrule(lr){6-9}

& PSNR$\uparrow$
& SSIM$\uparrow$
& LPIPS$\downarrow$
& FID$\downarrow$
& PSNR$\uparrow$
& SSIM$\uparrow$
& LPIPS$\downarrow$
& FID$\downarrow$ \\

\midrule
\rowcolor{tablegray!70}
\multicolumn{9}{c}{\textit{Retraction Strategy}} \\

Global amplification ($s<0$)
& 21.98 & 0.696 & 0.136 & 44.1 
& 20.68 & 0.704 & 0.233 & 92.7 \\

\midrule

\rowcolor{tablegray!70}
\multicolumn{9}{c}{\textit{Supervision Strategy}} \\

Endpoint zero-regret oracle
& 22.53 & 0.699 & 0.131 & 41.7
& 21.12 & 0.708 & 0.231 & 90.1 \\

Minimum-loss oracle
& 22.60 & 0.700 & 0.130 & 40.2
& 21.13 & 0.710 & 0.229 & 89.7 \\

Direct target-NLL gate
& 22.56 & 0.700 & 0.129 & 39.6
& 21.07 & 0.710 & 0.229 & 91.3 \\

\midrule

\rowcolor{tablegray!70}
\multicolumn{9}{c}{\textit{Gate Input}} \\

w/o source feature $\mathbf{u}^{S}$
& 22.49 & 0.700 & 0.130 & 42.4
& 21.05 & 0.706 & 0.232 & 90.2 \\

w/o correction feature $\mathbf{u}^{r}$
& 22.41 & 0.697 & 0.129 & 41.5
& 21.00 & 0.704 & 0.230 & 92.4 \\

w/o endpoint statistics $(w,e)$
& 22.60 & 0.702 & 0.129 & 39.8
& 21.17 & 0.709 & 0.228 & 89.5 \\

replace $\mathbf{u}^{r}$ with full-prefix feature $\mathbf{u}^{F}$
& 22.54 & 0.701 & 0.131 & 42.2
& 21.12 & 0.710 & 0.229 & 91.0 \\

\midrule

\rowcolor{tablepurple!70}
\textbf{{\spira} (Maximal zero-regret)}
& \textbf{22.72} & \textbf{0.705} & \textbf{0.126} & \textbf{38.3}
& \textbf{21.29} & \textbf{0.714} & \textbf{0.225} & \textbf{87.3} \\

\bottomrule
\end{tabular}
}
\label{tab:additional_ablation}
\end{table}

\begin{table}[t]
\centering
\setlength{\tabcolsep}{1.5mm}
\renewcommand{\arraystretch}{1.0}
\caption{\textbf{Ablation studies of the retraction granularity} on BraTS 2024 T1n$\rightarrow$T1c and SynthRAD 2025 MRI$\rightarrow$CT.
The AR block size is fixed at $(4,4,4)$, and only the spatial granularity of the shared retraction action is varied.}
\resizebox{1.00\linewidth}{!}{
\begin{tabular}{lcc|cccc|cccc}
\toprule

\multirow{2}{*}{\textbf{Retraction Granularity}}
& \multirow{2}{*}{\makecell{\textbf{Tokens/}\\\textbf{Action}}}
& \multirow{2}{*}{\makecell{\textbf{Actions/}\\\textbf{Block}}}
& \multicolumn{4}{c|}{\textbf{BraTS T1n$\rightarrow$T1c}}
& \multicolumn{4}{c}{\textbf{SynthRAD MRI$\rightarrow$CT}} \\

\cmidrule(lr){4-7}
\cmidrule(lr){8-11}

& &
& PSNR$\uparrow$
& SSIM$\uparrow$
& LPIPS$\downarrow$
& FID$\downarrow$
& PSNR$\uparrow$
& SSIM$\uparrow$
& LPIPS$\downarrow$
& FID$\downarrow$ \\

\midrule

Block-wise $(4,4,4)$
& 64 & 1
& 22.08 & 0.693 & 0.137 & 45.2
& 20.65 & 0.695 & 0.242 & 95.9 \\

Sub-block-wise $(2,2,2)$
& 8 & 8
& 22.39 & 0.697 & 0.132 & 40.9
& 20.98 & 0.705 & 0.231 & 91.6 \\

\rowcolor{tablepurple!70}
\textbf{Token-wise $(1,1,1)$}
& \textbf{1} & \textbf{64}
& \textbf{22.72} & \textbf{0.705} & \textbf{0.126} & \textbf{38.3}
& \textbf{21.29} & \textbf{0.714} & \textbf{0.225} & \textbf{87.3} \\

\bottomrule
\end{tabular}
}
\label{tab:ablation_granularity}
\end{table}

\begin{table}[t]
\centering
\setlength{\tabcolsep}{1.3mm}
\renewcommand{\arraystretch}{1.0}
\caption{\textbf{Ablation studies of the block size} on BraTS 2024 T1n$\rightarrow$T1c and SynthRAD 2025 MRI$\rightarrow$CT. The number of AR blocks is computed on the $32^3$ latent grid.}
\resizebox{1.00\linewidth}{!}{
\begin{tabular}{lcc|cccc|cccc}
\toprule

\multirow{2}{*}{\textbf{Block Size}}
& \multirow{2}{*}{\textbf{Tokens/Block}}
& \multirow{2}{*}{\textbf{\# AR Blocks}}
& \multicolumn{4}{c|}{\textbf{BraTS T1n$\rightarrow$T1c}}
& \multicolumn{4}{c}{\textbf{SynthRAD MRI$\rightarrow$CT}} \\

\cmidrule(lr){4-7}
\cmidrule(lr){8-11}

& &
& PSNR$\uparrow$
& SSIM$\uparrow$
& LPIPS$\downarrow$
& FID$\downarrow$
& PSNR$\uparrow$
& SSIM$\uparrow$
& LPIPS$\downarrow$
& FID$\downarrow$ \\

\midrule

$2\times4\times4$
& 32 & 1024
& 22.67 & 0.705 & 0.128 & 39.2
& 21.21 & 0.712 & 0.227 & 87.9 \\

$4\times4\times2$
& 32 & 1024
& 22.69 & \textbf{0.707} & \textbf{0.125} & 38.8
& 21.17 & 0.709 & 0.229 & 88.7 \\

\rowcolor{tablepurple!70}
\textbf{$4\times4\times4$}
& \textbf{64} & \textbf{512}
& \textbf{22.72} & 0.705 & 0.126 & \textbf{38.3}
& \textbf{21.29} & \textbf{0.714} & \textbf{0.225} & \textbf{87.3} \\

$8\times4\times4$
& 128 & 256
& 22.52 & 0.700 & 0.129 & 39.5
& 20.93 & 0.706 & 0.230 & 90.0 \\

$4\times4\times8$
& 128 & 256
& 22.41 & 0.697 & 0.128 & 40.2
& 20.97 & 0.706 & 0.228 & 89.1 \\

\bottomrule
\end{tabular}
}
\label{tab:ablation_block}
\end{table}

\textbf{Impact of retraction strategy and supervision strategy.}
We compare global prefix amplification with alternative oracle targets to assess how the retraction strategy and supervision affect generation.

\begin{itemize}
    \item \textbf{Global amplification ($s<0$).}
    We tune a single global scalar over $s\in[-0.5,0)$, extending the source--prefix path beyond Full-prefix AR to amplify the prefix correction. This tests whether stronger prefix influence is beneficial.
    
    \item \textbf{Endpoint zero-regret supervision.}
    The oracle restricts the retraction action to $s\in\{0,1\}$, allowing only full retention or full retraction.
    This variant tests whether continuous partial retraction is necessary.

    \item \textbf{Minimum-loss supervision.}
    The oracle selects the retraction amount that minimizes the current target loss along the retraction path.
    This tests whether directly optimizing the local prediction loss is a better supervision target.

    \item \textbf{Direct target-NLL gate.}
    We retain the complete {\spira} gate inputs, architecture, and
    source-only/full-prefix retraction path, but remove oracle supervision. Instead, the gate directly minimizes the GT factor NLL induced by its predicted retraction:
    \[
    \mathcal{L}_{\mathrm{gate\text{-}NLL}}
    =
    \mathbb{E}_{i,v}
    \left[
    -\frac{1}{d}
    \sum_{j=1}^{d}
    \log
    \operatorname{softmax}
    \left(
    z^{F}_{i,v,j}
    -
    \hat{s}_{i,v} r_{i,v,j}
    \right)_{q^{\mathrm{GT}}_{i,v,j}}
    \right].
    \]
    As in {\spira}, all endpoint logits and gate input features are
    stop-gradient quantities, so this objective updates only the gate and does not alter the AR predictor objective. This matched baseline tests whether directly optimizing target loss
    under the gate's GT-free inputs and capacity is sufficient, without
    constructing or distilling a per-example target-aware oracle.
\end{itemize}

\textbf{Impact of gate input}.
We remove or replace individual gate inputs to assess which source--prefix cues are important for retraction decisions.

\begin{itemize}
    \item \textbf{w/o source feature $\mathbf{u}^{S}$.}
    We remove the source-only feature from the gate input.
    This tests whether source evidence is important for deciding how much prefix correction to retain.

    \item \textbf{w/o correction feature $\mathbf{u}^{r}$.}
    We remove the feature describing the hidden-space change introduced by the prefix.
    This tests whether explicitly modeling the prefix correction benefits the retraction decision.

    \item \textbf{w/o endpoint statistics $(w,e)$.}
    We remove the lightweight statistics describing the difference between the source-only and full-prefix predictions.
    This tests whether these endpoint cues provide complementary information to the hidden features.

    \item \textbf{replace $\mathbf{u}^{r}$ with full-prefix feature $\mathbf{u}^{F}$.}
    We replace the correction feature with the full-prefix feature.
    This tests whether explicitly representing the source-to-prefix change is more useful than using the full-prefix state directly.
\end{itemize}

\begin{figure*}[t]
\centering
\includegraphics[width=1.00\linewidth]{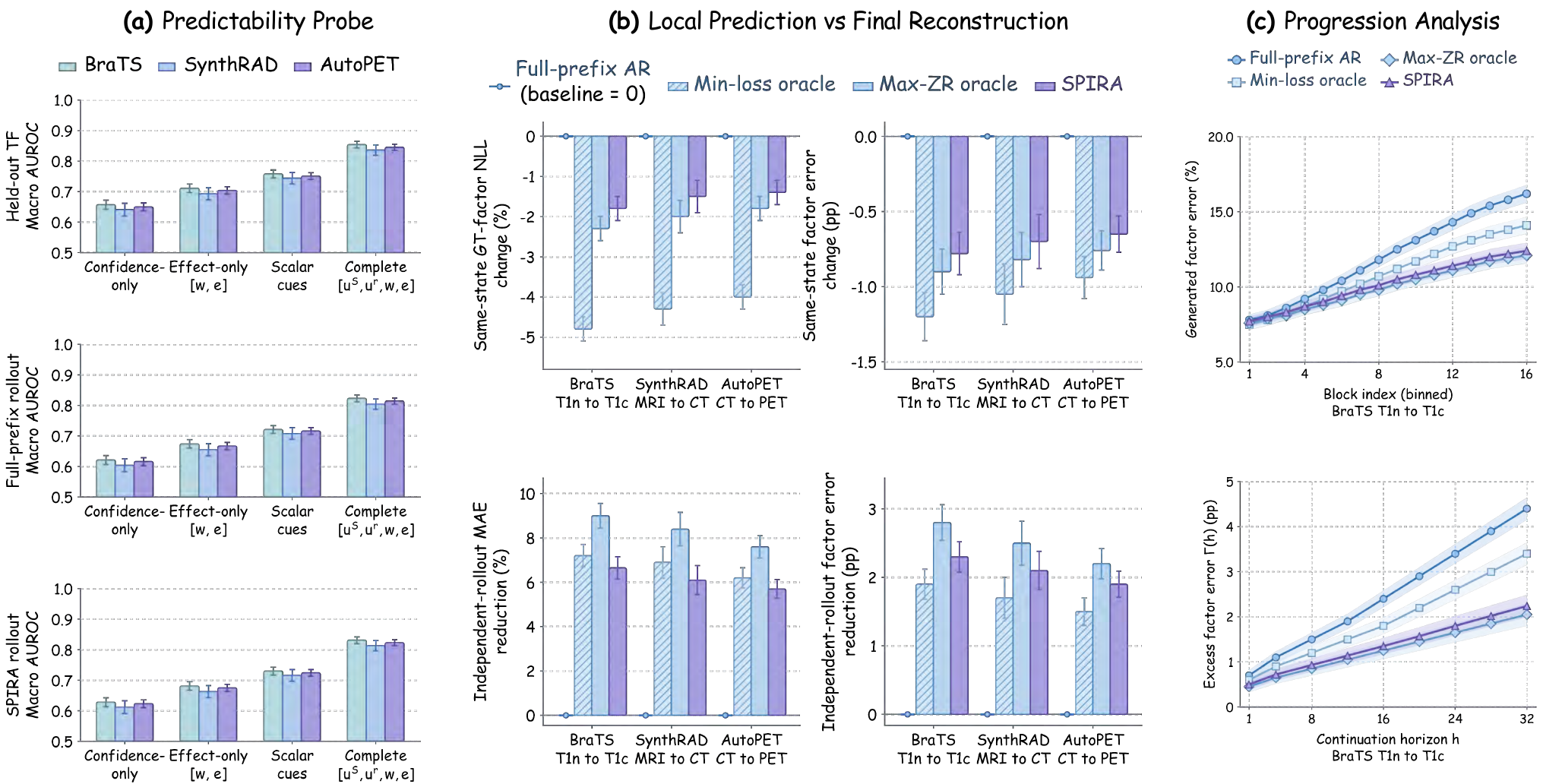}
\caption{\textbf{Additional analysis of selective retraction.}
(a) Predictability probe. Macro AUROC for predicting the three oracle retraction regimes from different GT-free feature sets on teacher-forced and generated-prefix states.
(b) Local prediction versus final reconstruction. Same-state GT-factor NLL is compared with independently rolled-out reconstruction performance.
(c) Progression and prefix sensitivity. Free-running factor error is tracked over block index, while controlled continuations from matched reference and generated prefixes measure the excess factor error $\Gamma^\pi(h)$.}
\label{fig:app_actionability}
\end{figure*}

Table~\ref{tab:additional_ablation} further examines retraction strategy, supervision strategy, and gate inputs. Global amplification falls below even Full-prefix AR on both tasks, showing that amplifying the prefix correction is detrimental rather than beneficial in these paired translation settings. Endpoint zero-regret and minimum-loss supervision also underperform {\spira} across all eight dataset--metric comparisons, supporting continuous maximal zero-regret retraction. The direct target-NLL gate, which retains the same complete gate inputs, capacity, and retraction path but directly optimizes GT target loss, also underperforms {\spira} across all eight metrics. This suggests that directly optimizing local target loss does not recover the benefit of constructing and distilling the maximal zero-regret predictive action.
This advantage is further supported by the GT-assisted policy comparisons in Figs.~\ref{fig:gap_gain}(a) and~\ref{fig:app_actionability}(b), which use the same predictor and avoid gate-training variation. Removing the source or correction features, or replacing the correction residual with the full-prefix feature, consistently degrades performance, while endpoint statistics provide smaller complementary gains. Together, these results support selective retraction, maximal zero-regret supervision, predictive-action distillation, and the proposed gate design.

\textbf{Impact of retraction granularity.}
Table~\ref{tab:ablation_granularity} compares block-wise, sub-block-wise, and token-wise retraction while keeping the AR block size fixed.
Performance consistently improves as the retraction granularity becomes finer, with token-wise control achieving the best results across both datasets and all metrics.
These findings suggest that sharing one action across multiple tokens can obscure local differences in prefix utility, while token-wise retraction better adapts the source--prefix balance to local prediction needs.

\textbf{Effect of block size.}
Table~\ref{tab:ablation_block} compares different AR block sizes, which trade off prediction granularity and the number of autoregressive steps.
Smaller blocks achieve comparable performance but require twice as many AR blocks, while larger blocks reduce the number of steps at the cost of translation quality.
Balancing translation quality and efficiency, we use $(4\times4\times4)$ as the default block size in all main experiments.

\textbf{Comparison with adapted context/prefix decoding methods.}
We compare {\spira} with adapted NWCAD–CoCoA~\citep{tao-agrawal-2026-worse} and VPG~\citep{liao2026vpg} using the same frozen dual-endpoint predictor and evaluation protocol. NWCAD–CoCoA operates on the source-only and full-prefix predictions, while VPG contrasts clean-prefix predictions with those obtained through same-block full-embedding replacement. Both baselines retain their original hyperparameter settings, keep the source condition fixed, and generate their own prefixes during free-running inference. In Table~\ref{tab:ablation_external}, {\spira} outperforms both adapted baselines across all four metrics on both tasks under this shared-predictor setting. These observations support the benefit of learned, region-wise prefix retraction over generic context/prefix decoding strategies in this setting.

\begin{table}[t]
\centering
\setlength{\tabcolsep}{1.5mm}
\renewcommand{\arraystretch}{1.0}
\caption{\textbf{Comparison with adapted context/prefix decoding methods} on BraTS T1n→T1c and SynthRAD MRI→CT. All methods share the same frozen predictor within each task.}
\resizebox{1.00\linewidth}{!}{
\begin{tabular}{lcccc|cccc}
\toprule

\multirow{2}{*}{\textbf{Methods}}
& \multicolumn{4}{c|}{\textbf{BraTS T1n$\rightarrow$T1c}}
& \multicolumn{4}{c}{\textbf{SynthRAD MRI$\rightarrow$CT}} \\

\cmidrule(lr){2-5}
\cmidrule(lr){6-9}

& PSNR$\uparrow$
& SSIM$\uparrow$
& LPIPS$\downarrow$
& FID$\downarrow$
& PSNR$\uparrow$
& SSIM$\uparrow$
& LPIPS$\downarrow$
& FID$\downarrow$ \\

\midrule

NWCAD~\citep{tao-agrawal-2026-worse}
& 21.70 & 0.692 & 0.157 & 47.0
& 20.67 & 0.698 & 0.245 & 98.5 \\

VPG~\citep{liao2026vpg}
& 21.98 & 0.696 & 0.142 & 45.2
& 20.81 & 0.700 & 0.237 & 96.4 \\

\rowcolor{tablepurple!70}
\textbf{\textbf{\spira}}
& \textbf{22.72} & \textbf{0.705} & \textbf{0.126} & \textbf{38.3}
& \textbf{21.29} & \textbf{0.714} & \textbf{0.225} & \textbf{87.3} \\

\bottomrule
\end{tabular}
}
\label{tab:ablation_external}
\end{table}

\textbf{Gate-only adaptation to generated-prefix states.}
To investigate the training--inference prefix distribution
shift, we conduct a small-scale gate-only adaptation study
while keeping the tokenizer and AR predictor frozen. We compare the original {\spira} with additional gate adaptation using either teacher-forced states alone ({\spira} + TF-only Adaptation) or an equal mixture of teacher-forced and generated-prefix states ({\spira} + Mixed-state Adaptation), under identical update budgets and supervision. Table~\ref{tab:gate_adaptation} indicates that mixed-state adaptation consistently improves all metrics over both original {\spira} and TF-only adaptation across the two tasks, whereas TF-only adaptation yields smaller and less consistent gains. These findings suggest that exposure to generated-prefix states can further improve free-running generation by mitigating the training--inference mismatch.

\begin{table}[t]
\centering
\setlength{\tabcolsep}{1.5mm}
\renewcommand{\arraystretch}{1.0}
\caption{\textbf{Gate-only adaptation to generated-prefix states}. TF-only and mixed-state adaptation use identical update budgets, with the latter combining equal proportions of teacher-forced and generated-prefix states.}
\resizebox{1.00\linewidth}{!}{
\begin{tabular}{l|cccc|cccc}
\toprule

\multirow{2}{*}{\textbf{Gate Adaptation}}
& \multicolumn{4}{c|}{\textbf{BraTS T1n$\rightarrow$T1c}}
& \multicolumn{4}{c}{\textbf{SynthRAD MRI$\rightarrow$CT}} \\

\cmidrule(lr){2-5}
\cmidrule(lr){6-9}

& PSNR$\uparrow$
& SSIM$\uparrow$
& LPIPS$\downarrow$
& FID$\downarrow$
& PSNR$\uparrow$
& SSIM$\uparrow$
& LPIPS$\downarrow$
& FID$\downarrow$ \\

\midrule

\rowcolor{tablepurple!70}
Original {\spira}
& 22.72 & 0.705 & 0.126 & 38.3
& 21.29 & 0.714 & 0.225 & 87.3 \\

{\spira} + TF-only Adaptation
& 22.78 & 0.705 & 0.124 & 37.5
& 21.38 & 0.717 & 0.224 & 88.9 \\

{\spira} + Mixed-state Adaptation
& \textbf{22.82} & \textbf{0.708}
& \textbf{0.122} & \textbf{37.2}
& \textbf{21.54} & \textbf{0.718}
& \textbf{0.221} & \textbf{85.2} \\

\bottomrule
\end{tabular}
}
\label{tab:gate_adaptation}
\end{table}

\subsection{Compute-matched Training Budget Comparison}
\label{app:exp_comupted_matched}

To disentangle the contributions of additional source-only supervision and selective retraction, we conduct a compute-matched comparison on BraTS 2024 T1n$\rightarrow$T1c. Specifically, we extend Standard AR (Sec.~\ref{sec:exp_ablation}) training from 100k to approximately 194k steps to match the training budget of {\spira} (26.32 GPU hours, Sec.~\ref{app:exp_efficiency}). We further compare Full-prefix AR ($s=0$) (Sec.~\ref{sec:exp_setup}), a validation-tuned global retraction scalar (Sec.~\ref{sec:exp_ablation}), and {\spira} using the same dual-endpoint predictor. All variants share identical tokenizers, AR architectures, data splits, and inference setup.

Table~\ref{tab:compute_matched} illustrates that extending Standard AR training improves its performance, but it still falls behind {\spira} under the same compute budget. Moreover, {\spira} consistently outperforms both Full-prefix AR and the tuned global scalar using the same dual-endpoint predictor. These results demonstrate that the gains cannot be explained by additional training compute or source-only supervision alone, highlighting the effectiveness of selective token-wise retraction.

\begin{table}[t]
\centering
\setlength{\tabcolsep}{2.5mm}
\renewcommand{\arraystretch}{1.0}
\caption{\textbf{Compute-matched training budget comparison on BraTS 2024 T1n$\rightarrow$T1c}. Standard AR is trained with an extended budget, while the remaining retraction variants share the same dual-endpoint predictor.}
\resizebox{1.00\linewidth}{!}{
\begin{tabular}{l|cc|cccc}
\toprule

\multirow{2}{*}{\textbf{Methods}}
& \multicolumn{2}{c|}{\textbf{Training Budget}}
& \multicolumn{4}{c}{\textbf{BraTS T1n$\rightarrow$T1c}} \\

\cmidrule(lr){2-3}
\cmidrule(lr){4-7}

& Steps
& GPU h
& PSNR$\uparrow$
& SSIM$\uparrow$
& LPIPS$\downarrow$
& FID$\downarrow$ \\

\midrule

Standard AR (Original)
& 100k & 13.57
& 22.16 & 0.696 & 0.137 & 42.5 \\

Standard AR (Compute-matched)
& 194k & $\approx$26.32
& 22.40 & 0.697 & 0.133 & 41.4 \\

\midrule

Full-prefix AR ($s=0$)
& 100k & 26.32
& 22.10 & 0.696 & 0.135 & 41.9 \\

Full-prefix AR + Tuned global scalar
& 100k & 25.57
& 22.28 & 0.699 & 0.133 & 42.8 \\

\midrule

\rowcolor{tablepurple!70}
\textbf{{\spira}}
& 100k & 26.32
& \textbf{22.72} & \textbf{0.705}
& \textbf{0.126} & \textbf{38.3} \\

\bottomrule
\end{tabular}
}
\label{tab:compute_matched}
\end{table}

\section{Further Analysis}
\label{app:analysis}

\begin{figure*}[t]
\centering
\includegraphics[width=1.00\linewidth]{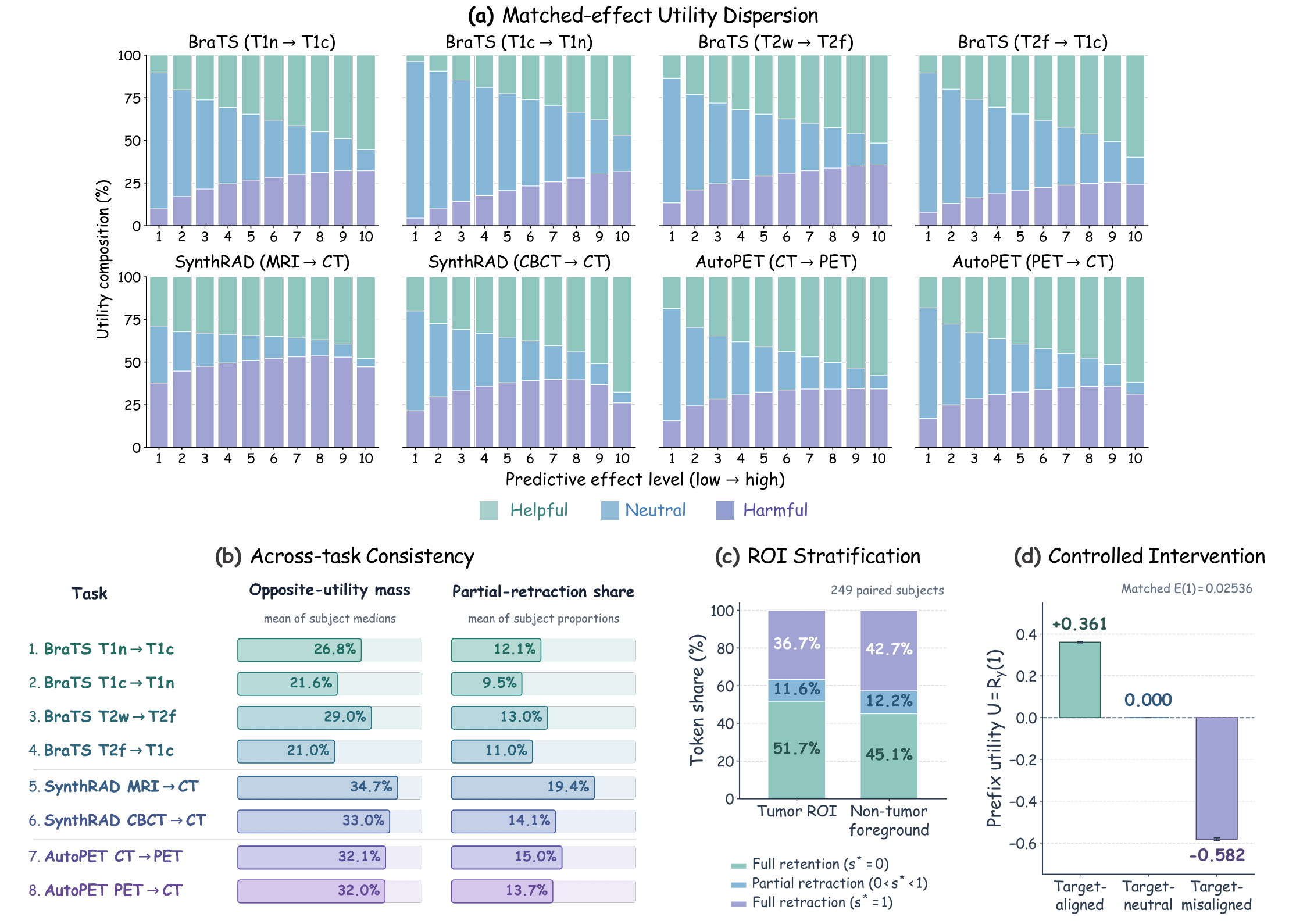}
\caption{\textbf{Additional analysis of the utility--effect gap.}
(a) Utility composition within task-specific predictive-effect deciles
across eight translation tasks.
(b) Utility mixing and partial retraction persist across tasks.
(c) Oracle retraction regimes differ between tumor ROIs and non-tumor
foreground on BraTS T1n$\rightarrow$T1c.
(d) At matched predictive effect, controlled changes in target alignment
produce positive, neutral, and negative prefix utilities.}
\label{fig:app_utility_effect}
\end{figure*}

\subsection{Additional Utility--Effect Analysis}
\label{app:utility_effect}

Building on Sec.~\ref{sec:exp_phenomenon}, we provide additional analyses of the utility--effect gap across effect levels, tasks, anatomical regions, and controlled matched-effect interventions.

\textbf{Effect measures.}
We distinguish the empirical TV-based prediction change used in Sec.~\ref{sec:motivation} from the theoretical KL effect $E_i(1)$ used in the regret decomposition. Both compare the same source-only and full-prefix prediction endpoints, with TV providing a bounded descriptive measure for the empirical analysis, whereas KL is used in Sec.~\ref{sec:exp_phenomenon} and here because it appears exactly in Eq.~\ref{eq:regret_decomposition}.

\textbf{Do matched effects imply matched utilities?}
Fig.~\ref{fig:app_utility_effect}(a) extends the utility analysis across
eight translation tasks and the full effect spectrum.
Helpful and harmful corrections coexist within the same effect levels
across all tasks, indicating that effect magnitude alone does not
determine correction utility.

\textbf{Does the pattern hold across tasks?}
Fig.~\ref{fig:app_utility_effect}(b) summarizes utility mixing and
partial retraction across tasks.
Opposite-utility mass remains 21.0--34.7\%, while partial retraction
accounts for 9.5--19.4\% of predictions, indicating that both patterns
persist across datasets and translation directions.

\textbf{Does retraction vary across regions?}
Fig.~\ref{fig:app_utility_effect}(c) compares tumor ROIs with non-tumor
foreground on BraTS T1n$\rightarrow$T1c.
Tumor regions have a higher full-retention share
(51.7\% vs. 45.1\%) and a lower full-retraction share
(36.7\% vs. 42.7\%), indicating spatial variation in the preferred
source--prefix balance.

\textbf{Does target alignment matter at fixed effect?}
For Fig.~\ref{fig:app_utility_effect}(d), we use real source-only
predictions from BraTS T1n$\rightarrow$T1c as anchors and fix the
endpoint effect to $E(1)=0.0254$.
Target-aligned, neutral, and misaligned endpoints are constructed while
numerically matching this effect.
Their utilities are $+0.361$, $0.000$, and $-0.582$, respectively,
directly illustrating that matched effects can have different utilities
when target alignment changes.

All three interventions are feasible for 81.6\% of anchors, yielding
3.54M matched endpoints across 249 subjects.
The maximum effect-matching error is $3.4\times10^{-9}$.
Bars report subject-equal means with 95\% confidence intervals from
1,000 subject-level bootstrap samples.

\subsection{Additional Analysis of Selective Retraction}
\label{app:mechanism}

\textbf{Actionability and gain definition.}
For subject $n$ in task $t$, we measure baseline actionability on
Full-prefix rollout states as
\begin{equation}
\mathcal{A}_{n,t}
=
\frac{
\sum_{i\in\mathcal{V}_{n,t}} E_{n,t,i}(s^{*,F}_{n,t,i})
}{
\sum_{i\in\mathcal{V}_{n,t}} E_{n,t,i}(1)+\epsilon
},
\end{equation}
where $s^{*,F}_{n,t,i}$ is the GT-assisted maximal zero-regret
retraction computed on the Full-prefix rollout state, and
$\mathcal{V}_{n,t}$ contains valid target tokens. Thus,
$\mathcal{A}_{n,t}$ measures the fraction of endpoint predictive
effect that can be safely retracted.

For a policy $\pi$, we define its subject-level generation gain over
Full-prefix AR as
\begin{equation}
g^\pi_{n,t}
=
100\,
\frac{
\mathrm{MAE}^{F}_{n,t}-\mathrm{MAE}^{\pi}_{n,t}
}{
\mathrm{MAE}^{F}_{n,t}
}.
\end{equation}
We quantify how actionability relates to {\spira}'s additional gain over
effect-only using
\begin{equation}
g^{\mathrm{SPIRA}}_{n,t}-g^{\mathrm{effect}}_{n,t}
=
\alpha_t+\beta\,\widetilde{\mathcal{A}}_{n,t}+\varepsilon_{n,t},
\end{equation}
where $\widetilde{\mathcal{A}}_{n,t}$ is actionability standardized
within each task and $\alpha_t$ is a task-specific intercept.
Accordingly, $\beta$ denotes the extra percentage-point gain of {\spira}
over effect-only per one standard deviation increase in actionability.

\textbf{Can oracle retraction be predicted without the target?}
We train lightweight probes on teacher-forced training states to predict the three oracle regimes, namely full retention, partial retraction, and full retraction.
We utilize a one-hidden-layer 64-d MLP probe with subject-disjoint training and selection splits.
We compare source confidence $c^S$, effect-only statistics $[w,e]$, scalar cues $[c^S,c^F,w,e,E(1)]$, and the complete GT-free gate features $[u^S,u^r,w,e]$.
The same probe is evaluated without refitting on held-out teacher-forced, Full-prefix rollout, and {\spira} rollout states, using macro one-vs-rest AUROC.
Complete features perform best across all settings (macro AUROC 0.804--0.854), exceeding confidence or effect alone and indicating that source--correction features contain additional oracle-relevant information.

\textbf{How often does the learned gate violate the local zero-regret constraint?}
We further evaluate the learned policies on identical {\spira}-generated
prefix states and compute the current-state regret
$R_y(\hat{s})=\ell_y(z(\hat{s}))-\ell_y(z^F)$.
Within each token, previously generated {\spira} factors are kept fixed
when evaluating regret, while GT factors are used only for scoring.
We report two complementary metrics. The positive-regret rate (PRR) measures how often the learned action violates the local zero-regret condition, i.e., the fraction of actions with $R_y(\hat{s})>0$. Positive-regret magnitude measures how severe these violations are, computed as the mean factor-averaged NLL regret over violating actions. As shown in Table~\ref{tab:regret_audit}, {\spira} consistently
reduces both quantities across all three tasks. Compared with the effect-only gate, PRR decreases from 26.4--31.0\% to 15.1--17.4\%, while positive-regret magnitude decreases from 0.020--0.025 to 0.011--0.014 nats. Accordingly, {\spira} better approximates the oracle's local zero-regret behavior on generated-prefix states, while the remaining violations emphasize that this property does not extend as a guarantee to the learned gate.

We further divide each 512-block generation window into early (blocks 1--171), middle (172--342), and late (343--512) stages to assess gate reliability as generation progresses. Within each stage, PRR is aggregated over all valid tokens, while positive-regret magnitude is averaged over violating tokens across subjects and windows. As reported in Table~\ref{tab:stagewise_regret}, both metrics generally increase in later stages. Nevertheless, {\spira} consistently exhibits fewer and smaller violations than the effect-only gate across all stages and tasks, indicating that its relative advantage persists throughout free-running generation.

\textbf{Does lower current token loss imply better final reconstruction?}
We first compare Full-prefix, minimum-loss, maximal zero-regret, and {\spira} on identical Full-prefix rollout states, fixing the source, generated prefix, factor history, and endpoint predictions.
Minimum-loss achieves the lowest same-state GT-factor NLL (a 4.0--4.8\% reduction relative to Full-prefix versus 1.8--2.3\% for maximal zero-regret), as expected from its objective.
We then roll out each policy independently from the beginning.
Despite its lower local NLL, minimum-loss yields a worse final decoded MAE than maximal zero-regret (6.2--7.2\% versus 7.6--9.2\% MAE reduction), showing that lower current token loss does not necessarily translate into better free-running reconstruction, while GT-free {\spira} achieves comparable gains of 5.7--6.8\%.

\textbf{How do policies respond to generated-prefix errors?}
We first report free-running factor error over 16 block-index bins within each 512-block latent window.
To isolate prefix sensitivity, we additionally start controlled continuations at fixed 25\%, 50\%, and 75\% positions from either the GT reference prefix $P_{\rm ref}$ or the matched Full-prefix generated prefix $P_{\rm gen}$, and let each policy continue for $H=32$ blocks.
We define prefix sensitivity as
$\Gamma^\pi(h)=\mathbb{E}[e^\pi(P_{\rm gen},h)-e^\pi(P_{\rm ref},h)]$,
the excess factor error caused by replacing the reference prefix with the generated prefix at the same starting block.
Maximal zero-regret and {\spira} show slower free-running error growth and lower $\Gamma^\pi$, consistent with reduced sensitivity to generated-prefix perturbations.

\begin{table}[t]
\centering
\caption{\textbf{Local regret audit on generated-prefix states.}
We evaluate learned retraction policies on identical {\spira}-generated states, using the GT target only to measure current-state regret. Positive-regret rate reports the fraction of actions with
$R_y(\hat{s})>0$, while positive-regret magnitude reports the
mean regret over these violating actions.}
\setlength{\tabcolsep}{2.0mm}
\resizebox{0.90\linewidth}{!}{
\begin{tabular}{lcccccc}
\toprule
& \multicolumn{2}{c}{\textbf{BraTS T1n$\rightarrow$T1c}}
& \multicolumn{2}{c}{\textbf{SynthRAD MRI$\rightarrow$CT}}
& \multicolumn{2}{c}{\textbf{AutoPET CT$\rightarrow$PET}} \\
\cmidrule(lr){2-3}
\cmidrule(lr){4-5}
\cmidrule(lr){6-7}

\textbf{Methods}
& \textbf{PRR (\%)$\downarrow$}
& \textbf{Regret$\downarrow$}
& \textbf{PRR (\%)$\downarrow$}
& \textbf{Regret$\downarrow$}
& \textbf{PRR (\%)$\downarrow$}
& \textbf{Regret$\downarrow$} \\
\midrule
Effect-only gate
& 27.1 & 0.022
& 31.0 & 0.025
& 26.4 & 0.020 \\

\rowcolor{tablepurple!70}
\textbf{{\spira}}
& \textbf{15.9} & \textbf{0.013}
& \textbf{17.4} & \textbf{0.014}
& \textbf{15.1} & \textbf{0.011} \\
\bottomrule
\end{tabular}
}
\label{tab:regret_audit}
\end{table}

\begin{table}[t]
\centering
\caption{\textbf{Stage-wise local regret on generated-prefix states.}
We report positive-regret rate (PRR, \%) and magnitude (nats)
across early, middle, and late AR blocks on identical
{\spira}-generated states.}
\setlength{\tabcolsep}{2.0mm}
\resizebox{0.95\linewidth}{!}{
\begin{tabular}{llcccccc}
\toprule
\multirow{2}{*}{\textbf{Task}}
& \multirow{2}{*}{\textbf{Methods}}
& \multicolumn{2}{c}{\textbf{Early}}
& \multicolumn{2}{c}{\textbf{Middle}}
& \multicolumn{2}{c}{\textbf{Late}} \\
\cmidrule(lr){3-4}
\cmidrule(lr){5-6}
\cmidrule(lr){7-8}
& & \textbf{PRR (\%)$\downarrow$}
& \textbf{Regret$\downarrow$}
& \textbf{PRR (\%)$\downarrow$}
& \textbf{Regret$\downarrow$}
& \textbf{PRR (\%)$\downarrow$}
& \textbf{Regret$\downarrow$} \\
\midrule

\multirow{2}{*}{BraTS T1n$\rightarrow$T1c}
& Effect-only
& 25.7 & 0.020
& 27.2 & 0.022
& 28.4 & 0.024 \\

& \cellcolor{tablepurple!70}\textbf{{\spira}}
& \cellcolor{tablepurple!70}\textbf{14.1}
& \cellcolor{tablepurple!70}\textbf{0.011}
& \cellcolor{tablepurple!70}\textbf{15.7}
& \cellcolor{tablepurple!70}\textbf{0.012}
& \cellcolor{tablepurple!70}\textbf{17.9}
& \cellcolor{tablepurple!70}\textbf{0.016} \\

\midrule

\multirow{2}{*}{SynthRAD MRI$\rightarrow$CT}
& Effect-only
& 28.6 & 0.021
& 31.3 & 0.025
& 33.1 & 0.029 \\

& \cellcolor{tablepurple!70}\textbf{{\spira}}
& \cellcolor{tablepurple!70}\textbf{15.4}
& \cellcolor{tablepurple!70}\textbf{0.012}
& \cellcolor{tablepurple!70}\textbf{17.1}
& \cellcolor{tablepurple!70}\textbf{0.014}
& \cellcolor{tablepurple!70}\textbf{19.7}
& \cellcolor{tablepurple!70}\textbf{0.016} \\

\midrule

\multirow{2}{*}{AutoPET CT$\rightarrow$PET}
& Effect-only
& 23.8 & 0.015
& 26.4 & 0.021
& 29.0 & 0.024 \\

& \cellcolor{tablepurple!70}\textbf{{\spira}}
& \cellcolor{tablepurple!70}\textbf{13.2}
& \cellcolor{tablepurple!70}\textbf{0.008}
& \cellcolor{tablepurple!70}\textbf{15.0}
& \cellcolor{tablepurple!70}\textbf{0.010}
& \cellcolor{tablepurple!70}\textbf{17.1}
& \cellcolor{tablepurple!70}\textbf{0.015} \\

\bottomrule
\end{tabular}
}
\label{tab:stagewise_regret}
\end{table}


\subsection{Computational Efficiency Analysis}
\label{app:exp_efficiency}

\begin{table}[t]
\centering
\caption{\textbf{Training efficiency versus translation quality on BraTS 2024.}
We compare the computational cost of representative internal variants
under the same training setting, together with translation quality on
the T1n$\rightarrow$T1c task.}
\setlength{\tabcolsep}{1.5mm}
\resizebox{0.95\linewidth}{!}{
\begin{tabular}{lccccccccc}
\toprule
& \multicolumn{5}{c}{\textbf{Training Cost}}
& \multicolumn{4}{c}{\textbf{BraTS 2024 T1n$\rightarrow$T1c}} \\
\cmidrule(lr){2-6}
\cmidrule(lr){7-10}

\multirow{2}{*}{\textbf{Variant}}
& \textbf{Params}
& \textbf{FLOPs}
& \textbf{Time/Step}
& \textbf{Peak VRAM}
& \textbf{GPU h}
& \multirow{2}{*}{\textbf{PSNR$\uparrow$}}
& \multirow{2}{*}{\textbf{SSIM$\uparrow$}}
& \multirow{2}{*}{\textbf{LPIPS$\downarrow$}}
& \multirow{2}{*}{\textbf{FID$\downarrow$}} \\

& (M)$\downarrow$
& (G)$\downarrow$
& (s)$\downarrow$
& (GB)$\downarrow$
& $\downarrow$
& & & & \\

\midrule
Standard AR
& 77.73 & 921.1 & 0.489 & {12.14} & 13.57
& 22.16 & 0.696 & 0.137 & 42.5 \\

Tuned global scalar
& 77.73 & 1802.5 & 0.920 & {21.86} & 25.57
& 22.28 & 0.699 & 0.133 & 42.8 \\

Effect-only gate
& 77.80 & 1815.1 & 0.930 & {21.86} & 25.83
& 22.31 & 0.700 & 0.131 & 40.3 \\

Direct $s^*$ regression
& 77.86 & 1826.1 & 0.948 & {21.86} & 26.32
& 22.60 & 0.701 & 0.129 & 39.9 \\

\midrule
\rowcolor{tablepurple!70}
\textbf{\spira}
& 77.86 & 1827.3 & 0.948 & {21.86} & 26.32
& \textbf{22.72} & \textbf{0.705}
& \textbf{0.126} & \textbf{38.3} \\

\bottomrule
\end{tabular}
}
\label{tab:training_efficiency}
\end{table}

\begin{table}[t]
\centering
\caption{\textbf{Inference efficiency comparison.}
We report end-to-end inference latency and peak GPU memory for
representative internal variants on BraTS 2024 T1n$\rightarrow$T1c
and SynthRAD 2025 MRI$\rightarrow$CT. All variants are evaluated
under the same hardware, precision, and batch-size settings.}
\setlength{\tabcolsep}{2.0mm}
\resizebox{0.70\linewidth}{!}{
\begin{tabular}{lcccc}
\toprule
& \multicolumn{2}{c}{\textbf{BraTS 2024}}
& \multicolumn{2}{c}{\textbf{SynthRAD 2025}} \\
\cmidrule(lr){2-3}
\cmidrule(lr){4-5}

\textbf{Variant}
& \textbf{Latency}
& \textbf{Peak VRAM}
& \textbf{Latency}
& \textbf{Peak VRAM} \\

& (s/vol)$\downarrow$
& (GB)$\downarrow$
& (s/vol)$\downarrow$
& (GB)$\downarrow$ \\

\midrule
Full-prefix AR
& 16.03 & 22.34
& 89.06 & 22.34 \\

Tuned global scalar
& 16.49 & 22.34
& 92.94 & 22.34 \\

Effect-only gate
& 16.67 & 22.34
& 93.59 & 22.34 \\

Direct $s^*$ regression
& 16.85 & 22.34
& 94.27 & 22.34 \\

\midrule
\rowcolor{tablepurple!70}
\textbf{\spira}
& 16.85 & 22.34 
& 94.27 & 22.34 \\

\bottomrule
\end{tabular}
}
\label{tab:inference_efficiency}
\end{table}

To assess the practical efficiency of {\spira}, we analyze its
computational cost during both AR training and inference.
All efficiency measurements are conducted on a single NVIDIA RTX 5090
GPU under the same precision and implementation settings.

\textbf{Wall-clock Cost of AR Training.}
Table~\ref{tab:training_efficiency} shows that SPIRA has a non-negligible training overhead relative to Standard AR, increasing total training cost from 13.57 to 26.32 GPU hours ($1.94\times$). 
This increase mainly results from dual-endpoint prediction, which raises
FLOPs from 921.1G to 1827.3G and time per step from 0.489 to
0.948 seconds, while the retraction module itself adds only 0.13M
parameters. Crucially, the matched dual-endpoint variants have nearly identical training costs but different translation performance.
For example, Direct $s^\ast$ regression has the same parameter count,
time per step, and total GPU hours as {\spira}, yet performs worse across all four metrics. The tuned global scalar and effect-only gate likewise operate at similar computational cost but achieve lower translation quality. Thus, while dual-endpoint training is substantially more expensive than Standard AR, the gains of {\spira} over compute-matched dual-endpoint variants cannot be explained by additional training compute alone.

\textbf{Inference Cost.}
As shown in Table~\ref{tab:inference_efficiency}, {\spira} introduces
modest inference overhead over Full-prefix AR, increasing latency from
16.03 to 16.85 seconds on BraTS 2024~\citep{de20242024} and from
89.06 to 94.27 seconds on SynthRAD 2025~\citep{https://doi.org/10.1002/mp.17981}. Using the tuned global scalar to isolate the endpoint cost, the additional source-only prediction accounts for 0.46 and 3.88 seconds on BraTS and SynthRAD, respectively, while adaptive gating adds only 0.36 and 1.33 seconds. Caching avoids repeated source-only evaluation but not its initial computation, indicating that most of the overhead comes from the second endpoint rather than the gate itself.
The higher SynthRAD latency reflects sliding-window whole-volume
generation, while peak VRAM remains unchanged at 22.34 GB because the
memory peak occurs during VidTok decoding. Thus, despite the higher training cost of dual-endpoint learning, {\spira} adds only modest inference latency without increasing peak memory usage.

\section{Discussion}
\label{app:discussion}

{\spira} studies source-conditioned autoregressive translation as a region-wise balance between fixed source evidence and evolving target context. By treating the prefix contribution as a controllable correction, it learns to retain target-useful information while retracting unnecessary or harmful corrections. Across eight 3D medical translation tasks, the results support both the effectiveness of this formulation and its connection to the utility--effect gap. Despite these encouraging results, several limitations remain and motivate future work.

\textbf{Multiple-source translation.}
Our experiments focus on translation from a single source modality to a single target modality. Extending {\spira} to multiple-source settings requires jointly controlling the contributions of several source modalities and the generated prefix, which is an interesting direction for future work.

\textbf{Teacher-forced gate supervision.}
The retraction oracle and main gate supervision use GT prefixes, whereas inference operates on generated prefixes. Generated-context training methods address this mismatch by exposing models to generated histories~\citep{NEURIPS2025_f4823f83, zhu2026causal, chen2026context}. While {\spira} regulates prefix contributions without changing the predictor's context training, our gate-only mixed-state adaptation (Appendix~\ref{app:exp_ablation}, Table~\ref{tab:gate_adaptation}) further improves free-running generation over teacher-forced-only adaptation under matched update budgets. More comprehensive integration with generated-context training remains future work.

\textbf{Token-space and image-space objectives.}
Our retraction regret is defined by token-level target loss, while final translation quality is evaluated after decoding in image space. Although our analyses connect token-level decisions with reconstruction behavior, directly incorporating image-space or downstream utility into retraction learning remains an open direction.

\textbf{Pairing quality and unpaired settings.}
{\spira} relies on spatially aligned source–target pairs for dual-endpoint AR training and GT-assisted oracle supervision. Imperfect pairing may introduce noise into target-loss evaluation and oracle decisions, potentially affecting gate reliability. Extending {\spira} to weakly paired or unpaired settings through registration-aware supervision, pseudo-pairing, or consistency-based objectives is an interesting direction for future work.

\textbf{Rationale for predictive-distribution distillation.}
The gate aims to reproduce the oracle’s predictive behavior rather than its scalar retraction amount \(s^*\). The same error in \(s\) may cause different prediction changes depending on the source-only and full-prefix distributions. Direct regression penalizes such scalar errors equally, whereas KL-based distribution matching accounts for their actual predictive impact. This provides a more prediction-aligned supervision signal, consistent with the superior performance observed in Sec.~\ref{sec:exp_ablation} (Table~\ref{tab:ablation}).

\textbf{Modest Gains and Potential Bottlenecks.}
Although {\spira}'s improvements over strong baselines are modest, they are consistent across diverse translation tasks
(Tables~\ref{tab:brats24}--\ref{tab:synthrad_autopet})
and extend to downstream segmentation (Appendix~\ref{app:exp_downstream}). The analyses in Secs.~\ref{sec:exp_phenomenon}--\ref{sec:exp_mechanism} further support the effectiveness of selective prefix retraction.

Several factors may constrain these gains. First, the source modality may lack target-specific information. Second, tokenization may lose fine details, although its contribution remains unclear (Appendix~\ref{app:vidtok_finetune}). Third, {\spira} is limited to adjusting predictions between the source-only and full-prefix endpoints, while the learned gate still has room for improvement (Fig.~\ref{fig:gap_gain}(a)). Finally, token-level supervision may not fully reflect image-space or downstream quality. Understanding these limitations and improving the corresponding components remain directions for future work.

\end{document}